\documentclass[letterpaper,12pt]{article}
\usepackage{amsmath} 
\usepackage[american]{babel}
\usepackage{tabularx}
\usepackage{booktabs}
\pdfoutput=1 
\usepackage{graphicx}
\usepackage{array}
\usepackage{tabu}
\usepackage{mathrsfs}

\newcommand\T{\rule{0pt}{2.6ex}}       
\newcommand\B{\rule[-1.2ex]{0pt}{0pt}} 

\usepackage{xfrac}

\usepackage{framed, color}
\usepackage{mdframed, color}

    \usepackage[utf8]{inputenc}
    \usepackage[T1]{fontenc}
    \usepackage{lmodern}
    \usepackage[x11names]{xcolor}
    \usepackage{framed}
    \colorlet{shadecolor-pink}{LavenderBlush2}
    \colorlet{framecolor}{Red1}
    \usepackage{lipsum}

    {\endMakeFramed}

    \newenvironment{frshaded*}{%
    \MakeFramed {\advance\hsize-\width \FrameRestore}}%
    {\endMakeFramed}

\usepackage{tcolorbox}

\usepackage{amsthm,color}
\theoremstyle{definition}

\mdfdefinestyle{example1}{%
    linecolor=blue,
    outerlinewidth=2pt,
    bottomline=false,
    leftline=false,rightline=false,
    skipabove=\baselineskip,
    skipbelow=\baselineskip,
    frametitle=\mbox{},
}

\mdfdefinestyle{exampledefault}{%
rightline=true,
leftline = true,
innerleftmargin=10,innerrightmargin=10,
frametitlerule=true,frametitlerulecolor=green,
frametitlebackgroundcolor=yellow,
frametitlerulewidth=2pt}

\usepackage{url}            
\usepackage{booktabs}       
\usepackage{amsfonts}       
\usepackage{nicefrac}       
\usepackage{microtype}      

\definecolor {shadecolor}{rgb}{1, 0.8, 0.3}
\definecolor {eqn-descrpt-aqua}{RGB}{210, 240, 235}
\definecolor {eqn-descrpt-frame-aqua}{RGB}{83, 111, 121}
\definecolor {med-aqua}{RGB}{165, 210, 200}
\definecolor {eqn-celadon}{RGB}{215, 240, 221}
\definecolor {eqn-frame-celadon}{RGB}{33, 130, 125}

\title{\textbf{2-D Cluster Variation Method Free Energy: Fundamentals and Pragmatics} \\
\vspace{10 mm} } 
\date{Revision Date: 2019-09-17\\
  Version 1}

\author{Alianna J. Maren \\
  Northwestern University School of Professional Studies\\
  Master of Science in Data Science Program\\  
  and\\
  Themasis  \\
  Themasis Technical Report TR-2019-02v1 (ajm)\\ 
  {\tt alianna.maren@northwestern.edu}\\
  {\tt alianna@aliannajmaren.com} 
  }

\begin{document}

\maketitle

\newpage


\abstract{Despite being invented in 1951 by R. Kikuchi, the 2-D Cluster Variation Method (CVM), has not yet received attention. Nevertheless, this method can usefully characterize 2-D topograpies using just two parameters;  the activation enthalpy and the interaction enthalpy. This Technical Report presents 2-D CVM details, including the dependence of the various configuration variables on the enthalpy parameters, as well as illustrations of various topographies (ranging from ``scale-free-like'' to ``rich club-like'') that result from different parameter selection. The complete derivation for the analytic solution, originally presented simply as a result in Kikuchi and Brush (1967) is given here, along with careful comparison of the analytically-predicted configuration variables versus those obtained when performing computational free energy minimization on a 2-D grid. The 2-D CVM can potentially function as a secondary free energy minimization within the hidden layer of a neural network, providing a basis for extending node activations over time and allowing temporal correlation of patterns.}

\vspace{10pt}


\pagebreak

%
\section{Introduction and Overview}
\label{sec:intro-and-overview}
%

This Technical Report serves as both a tutorial and as essential reference material for anyone who wishes to work with the 2-D cluster variation method (CVM), originally devised by Kikuchi in 1951 \cite{Kikuchi_1951_Theory-coop-phenomena}, and then further advanced by Kikuchi and Brush (1967) \cite{Kikuchi-Brush_1967_Improv-CVM}. 

The essential notion of the CVM is that we work with a more complex entropy expression within the free energy formalism for a system. 

In a simple Ising model, the entropy $S$ can be computed based on only the relative fraction of active units in a bistate system. That is, there are only two kinds of units; \textit{active} ones in state \textbf{A}, where the fraction of these units is denoted $x_1$, and \textit{inactive} ones in state \textbf{B}, where the fraction of these units is denoted $x_2$. (Of course, $x_1 + x_2 = 1.0$.)

In contrast to the simple entropy used in the basic Ising model, in the CVM approach, we expand the entropy term. The CVM entropy term considers not only the relative fractions of units in states \textbf{A} and \textbf{A}, but also a set of \textit{configuration variables}. The configuration variables for a  2-D CVM system include, in addition to the usual activation of single units, also the fractional values of different kinds of nearest neighbor and next-nearest-neighbor pairs, along with six different kinds of triplets. (Section~\ref{sec:config-variables} describes the configuration variables in depth.)

Our entropy term now involves what are essentially \textit{topographic} variables. That is, the location of one unit in conjunction with another now makes a substantial difference. 

As a result, when we do free energy minimization to find an equilibrium configuration for a given system, we arrive at a 2-D CVM with certain characteristic topographic properties. While the exact activation of a specific unit (whether it is in state  \textbf{A} or state \textbf{B}) can vary, each time a given system is brought to equilibrium, the overall values for the configuration variables should be consistent for a given set of enthalpy parameters. 

\textbf{\textit{These topographic patterns are worth our attention and interest.}} 

\begin{figure}[ht]
  \centering
  \fbox{
  \rule[-.5cm]{0cm}{4cm}\rule[-.5cm]{0cm}{0cm}	
  \includegraphics [trim=0.0cm 0cm 0.0cm 0cm, clip=true,   width=0.95\linewidth]{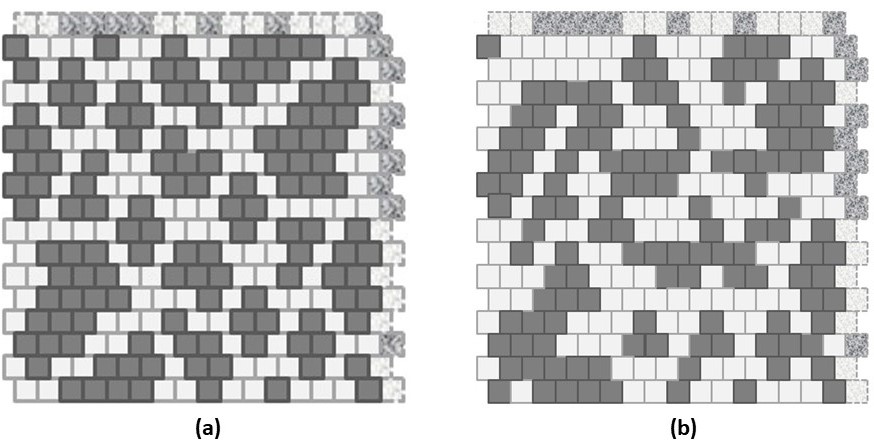}} 
  \vspace{3mm} 
  \caption{Illustration of (a) a manually-designed 2-D CVM grid that is (b) brought to a free energy equilibrium configuration.}   
\label{fig:2D-CVM-init-and-FEMin-scale-free_h-eq-1pt165_crppd_2019-06-26}
\end{figure}

Figure~\ref{fig:2D-CVM-init-and-FEMin-scale-free_h-eq-1pt165_crppd_2019-06-26} illustrates an example, for where a manually-designed initial system on the LHS (Left-Hand-Side) has been brought to free energy equilibrium on the RHS (Right-Hand-Side).

\vspace{3mm} 

The remaining sections will investigate this figure in more depth; this is presented here as an illustration of the kind of results obtained via free energy minimization in a 2-D CVM grid.  

Up until now, the CVM approach has not received a great deal of attention. However, as Figure~\ref{fig:2D-CVM-init-and-FEMin-scale-free_h-eq-1pt165_crppd_2019-06-26} illustrates, this approach is potentially useful for characterizing 2-D systems that would naturally gravitate to a free energy-minimized state. Examples of potential applications include: 

\begin{itemize}
\setlength{\itemsep}{1pt}
\item \textbf{\textit{Modeling systems where the topography is an essential component of system description}}; this can range from urban/rural topographies to brain images, 
\item \textbf{\textit{Providing an essential component of a new computational engine}}, where \textit{memories} of prior system states can persist over time, and 
\item \textbf{\textit{Providing the modeling component of a variational Bayes approach}} that uses a \textit{representational} system to model an \textit{external} system, where the two are separated by a Markov blanket, and both systems are presumed to come to separate free energy equilibrium states. 
\end{itemize}

The following Table~\ref{tbl:glossary-thermodynamic} presents a glossary of the thermodynamic terms used in this Report.

%
\begin{table}[ht!]\footnotesize
    \caption{Thermodynamic Variable Definitions}  
    \label{tbl:glossary-thermodynamic}
    \centering 
    \vspace{3mm}
    \begin{tabular}{|p{3cm}|p{10cm}|}
    \hline
	 \multicolumn{1}{|>{\centering\arraybackslash}m{3cm}|}	{\textbf{Variable}} 
    & \multicolumn{1}{>{\centering\arraybackslash}m{10cm}|}{\textbf{Meaning}} \T\B \\ 
    \hline                        
	 \multicolumn{1}{|>{\centering\arraybackslash}m{3cm}|}	{Activation enthalpy} 
    &Enthalpy  $\varepsilon_0$ associated with a single unit (node) in the ``on'' or ``active'' state (\textbf{A}); influences configuration variables and is set to 0 in order to achieve an analytic solution for the free energy equilibrium  \\ [3pt] 		

	 \multicolumn{1}{|>{\centering\arraybackslash}m{3cm}|}	{Configuration variable(s)} 
    &Nearest neighbor, next-nearest neighbor, and triplet patterns  \\ [8pt] 

	 \multicolumn{1}{|>{\centering\arraybackslash}m{3cm}|}	{Degeneracy} 
    &Number of ways in which a configuration variable can appear  \\ [5pt] 

	 \multicolumn{1}{|>{\centering\arraybackslash}m{3cm}|}	{Enthalpy} 
    &Internal energy \textit{H} results from both per unit and pairwise interactions; often denoted $H$ in thermodynamic treatments    \\ [5pt]  
    
	 \multicolumn{1}{|>{\centering\arraybackslash}m{3cm}|}	{Entropy} 
    &The entropy \textit{S} is the distribution over all possible states; often denoted $S$ in thermodynamic treatments and $H$ in information theory \\ [5pt]       

	 \multicolumn{1}{|>{\centering\arraybackslash}m{3cm}|}	{Equilibrium point} 
    &By definition, the free energy minimum for a closed system  \\ [5pt] 
 
	 \multicolumn{1}{|>{\centering\arraybackslash}m{3cm}|}	{Equilibrium distribution} 
    & Configuration variable values when free energy minimized for given \textit{h}  \\ [5pt]   

	 \multicolumn{1}{|>{\centering\arraybackslash}m{3cm}|}	{Ergodic distribution} 
    & Achieved when a system is allowed to evolve over a long period of time   \\ [5pt]  
 
	 \multicolumn{1}{|>{\centering\arraybackslash}m{3cm}|}	{Free Energy} 
    &The thermodynamic state function \textit{F}; where \textit{F = H-TS}; sometimes \textit{G} is used instead of \textit{F}; referring to (thermodynamic) Gibbs free energy   \\ [5pt]       
    
	 \multicolumn{1}{|>{\centering\arraybackslash}m{3cm}|}	{\textit{h-value}} 
    & A more useful expression for the interaction enthalpy parameter $\varepsilon_1$; $h = e^{2\beta\varepsilon_1}$, where $\beta = 1/{k_{\beta}T}$,  and where $k_{\beta}$ is Boltzmann's constant and $T$ is temperature; $\beta$ can be set to 1 for our purposes   \\ [5pt]       

	 \multicolumn{1}{|>{\centering\arraybackslash}m{3cm}|}	{Interaction enthalpy} 
    & Between two unlike units, $\varepsilon_1$; influences configuration variables  \\ [7pt] 
 
	 \multicolumn{1}{|>{\centering\arraybackslash}m{3cm}|}	{Interaction enthalpy parameter} 
    & Another term for the \textit{h-value} where $h=e^{2\varepsilon_1}$  \\ [9pt]   
    
	 \multicolumn{1}{|>{\centering\arraybackslash}m{3cm}|}	{Temperature} 
    &Temperature \textit{T} times Boltzmann's constant $k_{\beta}$ is set equal to one \\ [10pt]    
     \hline	  
  \end{tabular}
\end{table}
%

%
\section{Background}
\label{sec:background}
%

The CVM approach is a useful and insightful way for describing systems in which the entropy is considered to be more than the relative proportion of ``on'' and ``off'' units. In short, the CVM approach allows us to address the entropy of \textit{patterns}, and not just the simple entropy of unit identification. Thus, it is somewhat surprising that the CVM is not yet well understood, and is not yet used more widely. 

This section presents a brief historical overview, and identifies what may have been certain barriers to entry for researchers who might have otherwise considered this approach. 

%
\subsection{Evolution of the CVM applications}
\label{subsec:evoution-CVM-applications}
%

The primary applications of the CVM have, up until recently, predominantly been to computations of alloy phase diagrams as well as to studies of phase transitions in alloys. This application area was particularly dominant in the 1980's and 90's. For example, Sanchez et al. (1984)  addressed the CVM's role in describing configurational thermodynamics (and specifically phase stability) of alloys \cite{Sanchez-et-al_1984_Gen-Cluster-Descrp-Multicomp-Systems}, with a focus on the phenomenological and first principles theories of phase equilibrium. Maren et al. (1984) used a linear collection of 1-D CVMs to model phase transitions and hysteresis in a solid-state oxide \cite{Maren-et-al_1984_Theoretical-model-hysteresis-solid-state-phase-trans}.

In 1994, Sanchez and Becker reviewed this approach more comprehensively, focusing on how the 3-D CVM method could be used to construct the phase boundaries of solids (particularly alloys) \cite{Sanchez-and-Becker_1994_First-principles-CVM-phase-diagrams}. Similarly, Pelizzola described the CVM for modeling Pad\'{e} approximants and critical behaviour (1994) \cite{Pelizzola_1994_CVM-Pade-approx-crit-behavior}. Within the same timeframe, Finel described how the CVM approach could generate reliable approximate results, comparing favorably with Monte Carlo methods \cite{Finel_1994_CVM-applications}, and Cirillo et al. (1996) used the 3-D CVM to describe the phase structure of the 3-D gonihedric Ising system \cite{Cirillo_1996_phase-diagram-gonihedric}.

This attention to how the CVM method could be used to describe 3-D systems continued into the 2000's. Kikuchi and Masuda-Jindoi (2002) extended the original method to include continuous atomic displacements within binary alloys \cite{Kikuchi-Masuda-Jindoi_2002_CVM}.  This attention to how a 3-D CVM could describe the thermodynamic properties of alloys continued with work by Mohri (2013) \cite{Mohri_2013_CVM}.

Although the primary line of work from the 1980's through the early 2010's continued to be on applications of 3-D CVM models to alloys, some researchers began looking at the CVM in a more general context.

Beginning in the early 2000's, a few authors addressing general methods for machine learning and artificial intelligence included the CVM in their comprehensive reviews of methods. For example, Pelizzola broadened his earlier (1994) work to include the CVM in a more general treatment of probabilistic graph models (2005) \cite{Pelizzola_2005_CVM-stat-phys-prob-graph-models}. Similarly, Yedidia et al. described the role of the CVM as one method for belief propagation (2002) \cite{Yedidia-Freeman-Weiss_2002_Understanding-belief-prop}. Wainwright and Jordan included the CVM in their extensive monograph on graphical models, exponential families, and variational inference (2008) \cite{Wainwright-and-Jordan_2008_Graph-models-exp-fam-var-inf}. However, in all of these treatments, the CVM approach was included largely for completeness, and not as a primary method.
 
Practical applications, other than to metallurgy, have remained few up until now. Albers et al. used the CVM to study efficient linkage analysis on extended pedigrees (2006) \cite{Albers-et-al_2006_CVM-efficient-linkage-analysis}, and Barton and Cocco (2013) used the CVM method (which they described as ``selective cluster expansion'') to characterize neural structural and coding properties \cite{Barton-Cocco_2013_Ising-models-neural-activity}.

%
\subsection{The CVM, neural networks, and brain science}
\label{subsec:CVM-NN-brain-science}
%

While CVM applications to modeling 3-D systems such as alloys was dominant throughout the late 1990's and early 2000's, another (entirely different) application began to quietly take shape. 

In the early 1990's, Maren began evolving a new kind of CVM application. At that time, Maren was working in association with the Brain Research Center founded by Karl Pribram, who was then at Radford University, following a distinguished tenure at Stanford University \cite{Pribram_1991_Brain-and-Perception}. Also, neural networks had recently emerged as a new computational force, and there was the potential to use neurophysiologically-based insights to further advance neural networks.

Within the intellectual fervor inspired by Pribram and colleagues, several community members realized that the statistical physics methods that already were the underpinnings for some neural networks (e.g., the Hopfield neural network and the Boltzmann machine) could play an even stronger role. 

One of the challenging problems for neural networks, even then, was the difficulty making temporal associations and next-state predictions. The most well-known methods at that time were backpropagation-through-time and recurrent neural networks, both of which suffered deficiencies. A core factor underlying these deficiencies was that there was no way for a neural network to learn temporal associations that might be made over a longer time interval; both backpropagation-through-time and recurrent neural networks favored the influence of most-recent pattern presentations. 

One source of inspiration came from Fukushima's multi-layered neural network for visual (Japanese kanji) character recognition \cite{Fukushima_1980_Neocognitron}. In this early work, Fukushima anticipated the now-common convolutional neural network. In Fukushima's approach, activated nodes in the intermediate layers could form lateral connections with each other; connections that were not possible with the discriminative (e.g., supervised learning for a Multilayer Perceptron) or the generative (e.g., stochastic resonance, which was the precursor for contrastive divergence in the Boltzmann machine, and later the restricted Boltzmann machine). 

The work by Steve Grossberg and Ennio Mingolla in developing the Boundary Contour System (BCS), to emulate low-level mammalian visual processes, also influenced notions of how a CVM could play a role in a computational system \cite{Grossberg-and-Mingolla_1985a-Neural-dynamics-perceptual-grouping, Grossberg-and-Mingolla_1985b-Neural-dynamics-form-perception}. Grossberg and Mingolla used the gestalt laws of perceptual grouping to guide certain low-level visual processes, such as connecting short line segments if they occurred in the same line with each other and were sufficiently close to each other. This overcame significant difficulties in early computer vision systems. In particular, Grossberg and Mingolla emphasized how certain factors could cause connections between low-level processing units in a system. This would activate nodes that previously had not been active, to create at least contiguous line segments (if not clusters).  

Thus, one of the key inspirations for using the CVM in a neural network-based computational system came not from the now-classic discriminative and generative architectures, but from researchers who considered how lateral connections between active nodes might be useful.

%
\subsection{Potential role for a CVM system}
\label{subsec:potential-role}
%

Maren conceived the notion that if an entire CVM grid was used to create a set of responses to a given input stimulus, then the activation of certain nodes might decay gradually over time, and that lateral connections (created via Hebbian learning) between active nodes might stabilize responses to not only a currently-presented pattern, but also responses to short pattern sequences. 

The unique insight that came from interactions with Pribram and colleagues was that a 1-D or 2-D CVM could model interacting domains of neurons. As a step beyond the then-current thinking of neural networks as ``neural modeling systems,'' the CVM-based approach recognized the crucial role of creating clusters or physically-connected groups of neurons (neural domains) as the means by which neural systems responded to stimuli. 

One component of this insight was the realization that neural networks had hidden (latent) node activations that occurred in isolation from each other; each activation was strictly in response to learned connection weights each modulating inputs from the stimulus patterns. This meant that a given node's activation was dependent on the stimulus immediately feeding into it from the current pattern presentation. Even the backpropgation-through-time and recurrently-connected neural networks had sharp decays on the influence of prior patterns on a given node's activation. 

If a CVM approach were used, then a cluster of active nodes might potentially be more self-stabilizing than a single activated node. This self-stabilization could be enhanced by allowing lateral connections to form within an activated node cluster. Multiple factors could then impact how long the cluster's activation would persist. In fact, a cluster of activated nodes could potentially degrade node-by-node, where the nodes in the cluster core maintained longer temporal activation. This would potentially compare favorably with the single-node activation methods of existing neural networks. It could also potentially make a cluster more robust against temporal perturbations. 

The key was that free energy minimization, which was already used to guide connection weight development between network ``layers,'' could also be used \textit{within} a given layer as a secondary process. This CVM-based within-layer free energy minimization would not only produce a stable (if redundant) set of responses to pattern stimulus, it could also be a stabilizing force against pertubations. CVM free energy minimization (together with adjustable slow decay of activated nodes) could also be used to help a network learn temporal pattern association. 

With modest grant support from the Jeffries Foundation, Maren and colleagues developed an initial, 1-D proof-of-concept testbed for the CVM within a computational framework. Maren christened this the CORTECON, for \textit{COntent-Retentive, TEmporally CONnected} neural network. This led to three brief conference paper presentations between 1992 and 1994 \cite{Maren-Schwartz-Seyfried_1992_Config-entropy-stabilizes, Maren_1993_Free-energy-as-driving-function, Schwartz-Maren_1994_Domains-interacting-neurons}.

For various reasons (mostly having to do with severe intellectual property assignment agreements), the ideas behind the CORTECON languished from the mid-1990's until 2016, when Maren re-introduced this approach \cite{Maren_2016_CVM-primer-neurosci}. This work largely addressed the 1-D CVM, and did not include computational results for minimizing free energy across a 2-D CVM system. 

Maren's 2016 work was presaged by a small set of privately-produced technical reports, reviewing the CVM theory and presenting detailed derivations that were not included in the original Kikuchi (1951) and Kikuchi and Brush (1964) papers. One of these works specifically addressed the 2-D CVM (Maren, 2014) \cite{AJMaren-TR2014-003}.

%
\subsection{Further role for the CVM}
\label{subsec:Further-role}
%

There is a philsophical connection between the notion of using a 2-D CVM to model a physical system and the ideas advanced by Karl Friston regarding the role of free energy minimization as a key function in brain processes (2010, 2013, 2015) \cite{Friston_2010_Free-energy-principle-unified-brain-theory, Friston_2013_Life-as-we-know-it, Friston-et-al_2015_Knowing-ones-place-free-energy-pattern-recognition}.

Friston's equations extend the variational Bayes approximation (for example as explained by Beal (2003)\cite{Beal_2003_Variational-algorithm-approx-Bayes-inference}) into a framework in which an external system is represented by a system that is separated from the external system via a Markov blanket. The representational system is then modeled directly. Maren has produced a detailed description of the Friston's approach to using variational Bayes as compared with the classic approach \cite{AJMaren-TR2019-001v4-Deriv-Var-Bayes}.

One component of variational Bayes is that the modeling system should be brought as close as possible to the representational system. The notion of the variational Bayes approximation uses a formal resemblance between the approximation equations and statistical mechanics. Maren \cite{AJMaren-TR2019-001v4-Deriv-Var-Bayes} suggested that the process of modeling need not be simply adjusting parameters to reduce a Kullbeck-Leibler divergence, but actually carrying out a real free energy minimization. For some systems, the 2-D CVM would be an appropriate means for conducting this real free energy minimization.

%
\subsection{CVM applications summary and barriers to entry}
\label{subsec:barriers-to-entry}
%

For a method that was invented back in the 1950's, the CVM approach has been relatively under-utilized. The overwhelming applications have been to the study of phase transitions and phase diagrams in 3-D systems. 

While some leading researchers have identified the role of the CVM within the larger scope of graph theory and machine learning methods, it has not yet been practically applied to artificial intelligence or machine learning. In fact, there are no known instances of using a 2-D CVM to model any topographic system (other than the previously-cited work by Barton and Cocco \cite{Barton-Cocco_2013_Ising-models-neural-activity}). This is surprising, considering that the CVM seems to be an ideal means for modeling 2-D topographies and even dynamics. 

There are several difficulties that may have been barriers to entry. One is that Kikuchi and Brush (1967) presented the derivation of the analytic solution for the 2-D CVM configuration variables as a result, but without any of the intervening steps. The derivation of their result is a somewhat complex (and tedious and delicate) task. It has not previously been published in its entirety, except as a privately-published technical report by Maren (2014) \cite{AJMaren-TR2014-003}. 

Further, the code necessary to do even the most elementary CVM computations is likewise both substantive and delicate. Such code is not yet in the public domain, so that any researcher contemplating work in this area would be faced with extensive \textit{ab initio} code development. (See note at the end of this document for information on this author's code release.) 

This paper attempts to make the 2-D CVM approach more accessible to a diverse range of researchers. It presents, in Appendix A, the complete derivation of the configuration variables at equilibrium (for the singular case where such an analytic solution is possible, at $x_1 = x_2 = 0.5.$). It also presents a baseline study of the thermodynamic properties of the 2-D CVM. 

The cluster variation method uses a set of topographic variables, called the \textit{configuration variables}, as identified in the following Table~ \ref{tbl:config-variables-table}.

\begin{table}[t]
    \caption{Configuration Variables for the Cluster Variation Method}
    \label{tbl:config-variables-table}
    \centering
    \vspace{3mm}
    \begin{tabular}{|p{5cm}|p{2cm}|p{2cm}|}
    \hline
	 \multicolumn{1}{|>{\centering\arraybackslash}m{5cm}|}	{\textbf{Name}} 
    & 	 \multicolumn{1}{|>{\centering\arraybackslash}m{2cm}|}	{\textbf{Variable}}  
    &   \multicolumn{1}{|>{\centering\arraybackslash}m{2cm}|}   {\textbf{Instances}} \T\B \\ 
    \hline    	    

	 \multicolumn{1}{|>{\centering\arraybackslash}m{5cm}|}	{Unit} 
    &   \multicolumn{1}{|>{\centering\arraybackslash}m{2cm}|}	{$x_i$} 	
    &   \multicolumn{1}{|>{\centering\arraybackslash}m{2cm}|}	{$2$} \\ [3pt] 			  
  
	 \multicolumn{1}{|>{\centering\arraybackslash}m{5cm}|}	{Nearest-neighbor} 
    &   \multicolumn{1}{|>{\centering\arraybackslash}m{2cm}|}	{$y_i$} 	
    &   \multicolumn{1}{|>{\centering\arraybackslash}m{2cm}|}	{$3$} \\ [3pt] 	    

	 \multicolumn{1}{|>{\centering\arraybackslash}m{5cm}|}	{Next-nearest-neighbor} 
    &   \multicolumn{1}{|>{\centering\arraybackslash}m{2cm}|}	{$w_i$} 	
    &   \multicolumn{1}{|>{\centering\arraybackslash}m{2cm}|}	{$3$} \\ [3pt] 	
    
	 \multicolumn{1}{|>{\centering\arraybackslash}m{5cm}|}	{Triplet } 
    &   \multicolumn{1}{|>{\centering\arraybackslash}m{2cm}|}	{$z_i$} 	
    &   \multicolumn{1}{|>{\centering\arraybackslash}m{2cm}|}	{$6$} \\ [5pt] 	    
 
    \hline
  \end{tabular}
\end{table}

The following Section~\ref{sec:config-variables} discusses these configuration variables in depth.

%
\section{The Configuration Variables}
\label{sec:config-variables}
%

The key requirement to working with the cluster variation method is that we need to understand and use a set of \textit{configuration variables}. In a simple Ising system (the simplest statistical mechanics model for a system in which only two states are possible for each unit), we only need to think about whether a given unit is ``on'' or ``off,'' or in states \textbf{A} or \textbf{B}. 

We really only need one variable to describe the distribution of units in the system, because if wel let $x_1$ be the fraction of nodes in state \textbf{A}, and $x_2$ be the fraction of nodes in state \textbf{B}, then we also have that $x_2 = 1 - x_1$. This gives us a system that is relatively easy to model and visualize. 

In contrast, in the CVM, we expand the set of variables under consideration. We use a set of variables that are collectively referred to as the \textit{configuration variables}. 

In this section, we address three key factors that are essential for working with the \textit{configuration variables}:

\begin{enumerate} \itemsep0pt 
\item \textbf{\textit{Configuration variable definitions}} -- how they show up in the 2-D CVM grid, 
\item \textbf{\textit{Counting the onfiguration variables}} -- how each configuration variable is counted, and the V\&V thereof (verification and validation), and
\item \textbf{\textit{A brief interpretation}} -- how to interpret configuration variable values.
\end{enumerate} 

Later in this paper, we will examine how configuration variables change with the enthalpy parameters. However, even before we connect the grid topographies (expressed via configuration variables) with the free energy equation, we can establish a foundation for understanding what the configuration variables mean (in practical terms), and how they interact with each other.

%
\subsection{Introducing the Configuration Variables}
\label{subsec:Introducing-config-vars}
%

A 2-D CVM is characterized by a set of \textit{configuration variables}, which collectively represent single unit, pairwise combination, and triplet values. The configuration variables are denoted as: 

\begin{itemize}
\setlength{\itemsep}{1pt}
\item $x_i$ - Single units, 
\item $y_i$ - Nearest-neighbor pairs, 
\item $w_i$ - Next-nearest-neighbor pairs, and 
\item $z_i$ - Triplets. 
\end{itemize}

\begin{figure} [ht]
    \centering
        \includegraphics[trim=4cm 7.7cm 5.5cm 7.7cm, clip=true,  width=0.9\textwidth]{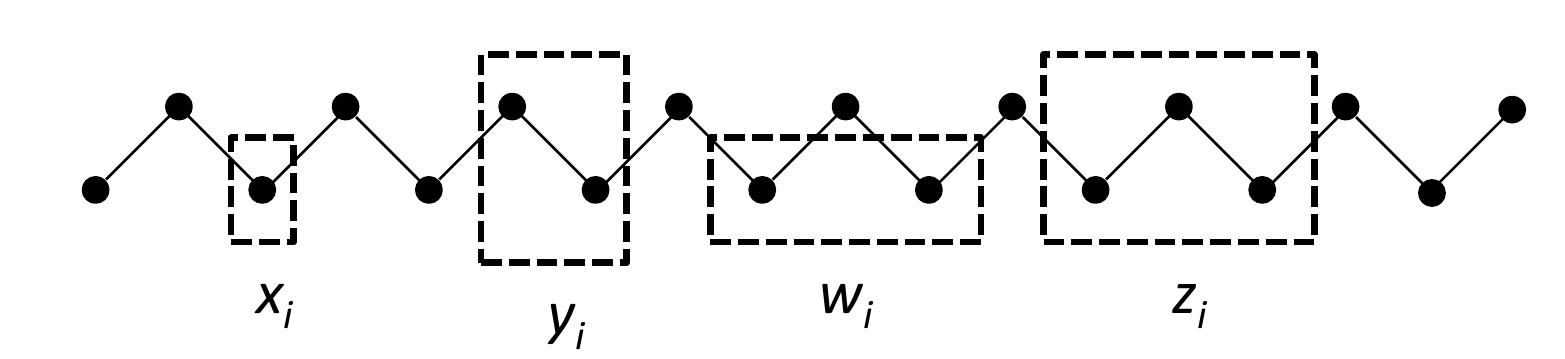}
    \caption{The 1-D single zigzag chain is created by arranging two staggered sets of $M$ units each. The configuration variables shown are $x_i$ (single units), $y_i$ (nearest-neighbors), $w_i$ (next-nearest-neighbors), and $z_i$ (triplets).}
    \label{fig:single_zigzag_chain_lbld}
\end{figure}

These configuration variables are illustrated for a single zigzag chain in Figure~\ref{fig:single_zigzag_chain_lbld}.
 
\begin{figure}[ht]
    \centering
        \includegraphics[trim=0cm 0cm 0cm 0cm, clip=true,  width=0.6\textheight]{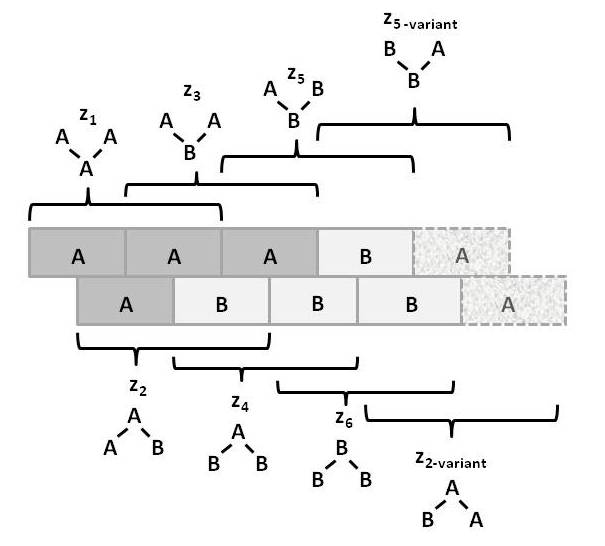}
    \caption{The six ways in which the configurations $z_i$ can be constructed.}
    \label{fig:CVM-1-D_base-graph2}
\end{figure}

For a bistate system (one in which the units can be in either state \textbf{A} or state \textbf{B}), there are six different ways in which the triplet configuration variables ($z_i$) can be constructed, as shown in Figure~\ref{fig:CVM-1-D_base-graph2}.

Notice that within Figure~\ref{fig:CVM-1-D_base-graph2}, the triplets $z_2$ and $z_5$ have two possible configurations each: \textbf{A}-\textbf{A}-\textbf{B} and \textbf{B}-\textbf{A}-\textbf{A} for $z_2$, and \textbf{B}-\textbf{B}-\textbf{A} and \textbf{A}-\textbf{B}-\textbf{B} for $z_5$. This means that \textbf{\textit{there is a degeneracy factor of 2}} for each of the $z_2$ and $z_5$ triplets. This is shown in Figure~\ref{fig:CVM-1-D_base-graph2} and also identified earlier in Table~\ref{tbl:config-variables-table}.

\begin{figure}[ht]
  \centering
  \fbox{
  \rule[-.5cm]{0cm}{4cm}\rule[-.5cm]{0cm}{0cm}	
  \includegraphics [trim=0.0cm 0cm 0.0cm 0cm, clip=true,   width=0.95\linewidth]{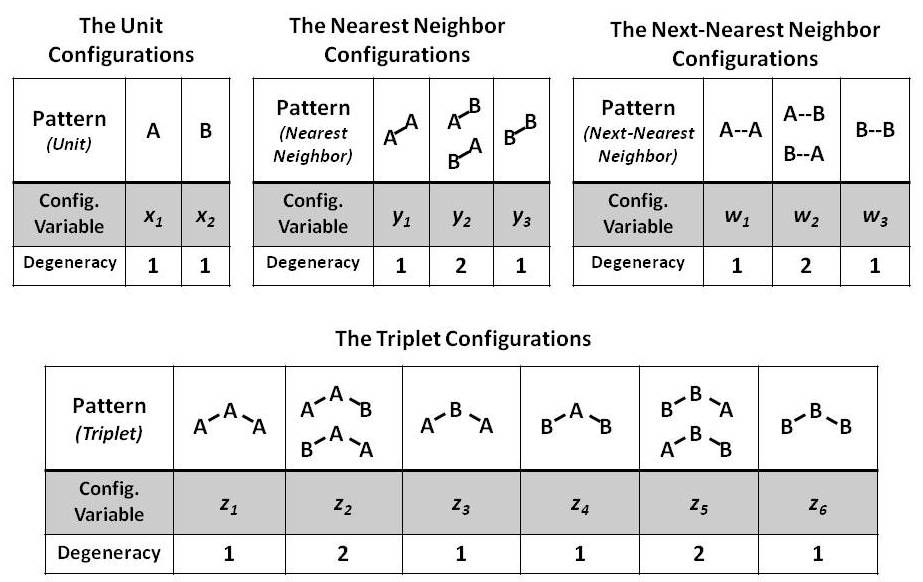}} 
  \vspace{3mm} 
  \caption{Illustration of the \textit{configuration variables} for the cluster variation method, showing the ways in which the configuration variables $y_i$, $w_i$, and $z_i$ can be constructed, together with their degeneracy factors $\beta_i$ and $\gamma_i$.}   
\label{fig:Config-var-weights_v3_crppd_2017-05-17}
\end{figure}

The degeneracy factors $\beta_i$ and $\gamma_i$ (number of ways of constructing a given configuration variable) are shown in Figure~\ref{fig:Config-var-weights_v3_crppd_2017-05-17}. For the pairwise combinations $y_2$ and $w_2$, $\beta_2 = 2$,  as $y_2$ and $w_2$ can each be constructed as either \textbf{A}-\textbf{B} or as \textbf{B}-\textbf{A} for $y_2$, or as \textbf{B}-~-\textbf{A} or as \textbf{A}-~-\textbf{B} for $w_2$. Similarly, $\gamma_2 = \gamma_5 = 2$ (for the triplets), as there are two ways each for constructing the triplets $z_2$ and $z_5$. All other degeneracy factors are set to 1. See the illustrations in Figs.~\ref{fig:CVM-1-D_base-graph2} and \ref{fig:Config-var-weights_v3_crppd_2017-05-17}.

\vspace{3mm} 

%
\subsection{Counting the Configuration Variables}
\label{subsec:counting-config-vars}
%

The first step in working with a 2-D CVM was to construct a computer program that would count the various neighboring pairs and triplets, and produce the resulting configuration variables. All computer code was devised in Python 3.6, using straightforward programming design. (The next major code iteration will require an object-oriented approach.)  

One of the most important verification and validation (V\&V) steps with these programs was to ensure that they accurately computed the various configuration variables, as described in Maren (2018) \cite{AJMaren-TR2018-001v2-V-and-V}.

\begin{figure}[ht]
  \centering
  \fbox{
  \rule[-.5cm]{0cm}{4cm}\rule[-.5cm]{0cm}{0cm}	
  \includegraphics [trim=0.0cm 0cm 0.0cm 0cm, clip=true,   width=0.95\linewidth]{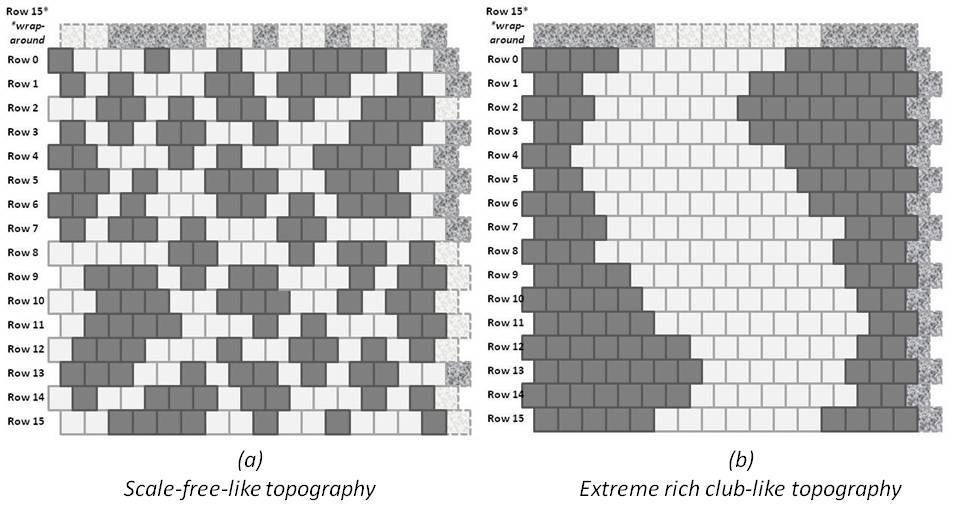}} 
  \vspace{3mm} 
  \caption{Illustration of the two different grids for experiments with the 2-D CVM system.}   
\label{fig:CVM-2-D_rich-club_and_scale-free_256-nodes}
\end{figure}

The early experiments with a 2-D CVM system (forming the bulk for the experimental work discussed here) were all done using various grids of 256 (16 x 16) units each. Two of these grids are illustrated in Figure~\ref{fig:CVM-2-D_rich-club_and_scale-free_256-nodes}. 

Keeping initial experiments confined to this relatively small 16x16 (256-unit) grid allowed three things: 

\begin{enumerate} \itemsep0pt 
\item \textbf{\textit{Sufficient variety in local patterns}} -- this grid size was large enough to illustrate several distinct kinds of topographies (each corresponding to different \textit{h-values} or enthalpy parameters; \textit{h-values} will be discussed in a following sectioni), 
\item \textbf{\textit{Sufficient nodes}} -- there were enough nodes so that triplet-configuration extrema could be explored in some detail (e.g., for relatively small numbers of nodes in state \textbf{A}, and 
\item \textbf{\textit{Countability}} --the V\&V effort required that several early versions of the grid be \textit{manually counted} for all the configuration values for a given 2-D grid configuration, and matched against the results from the program.
\end{enumerate} 

One final advantage of the 16 x 16 grid layout was that the different grid configurations were both large enough to show diversity, but small enough so that it was possible to manually create figures illustrating the activation states (\textbf{A} or \textbf{B}) of each node, thus illustrating the detailed particulars of each configuration design. (The code at this point included only rudimentary print-outs of grid patterns.)

The experiments began with manually-designed grid configurations, such as the two shown in Figure~\ref{fig:CVM-2-D_rich-club_and_scale-free_256-nodes}. These two configurations correspond (somewhat) to the notions of ``scale-free'' and ``rich club'' topographies, as observed in various neural communities. (For references to the scale-free and rich club literature on brain topographies, please consult Maren (2016) \cite{Maren_2016_CVM-primer-neurosci}.)

\vspace{3mm} 

These two different grid configurations are early attempts to characterize how the \textit{h}-values can be identified for grids with different total configuration variable values. The following sections  will discuss \textit{h}-values in the context of the free energy equation. 

Both of these systems were created with the constraint of equiprobable occurrence of units in states (\textbf{A} or \textbf{B}; that is, $x_1 = x_2= 0.5$. This was done to facilitate the necessary code V\&V step, which will be discussed in later in this section. Thus, for the configurations shown in Figure~\ref{fig:CVM-2-D_rich-club_and_scale-free_256-nodes}, both the (\textbf{\textit{a}}) and (\textbf{\textit{b}}) grids have 128 nodes each of units in state \textbf{A} and in state \textbf{B}.

The configuration on the left of Figure~\ref{fig:CVM-2-D_rich-club_and_scale-free_256-nodes} is an effort to build a ``scale-free-like'' system. The notion of a ``scale-free'' system is that the same kind of pattern replicates itself throughout various scales of observation in a system. Thus, for the ``scale-free-like'' configuration shown in Figure~\ref{fig:CVM-2-D_rich-club_and_scale-free_256-nodes} (\textbf{\textit{a}}), the design was originally intended to be 180-degree symmetrical around a central axis (dihedral group-2 symmetry). Specifically, the left and right sides were to be identical in a rotated-180-degree sense. 

For ease in design of the ``scale-free-like'' system, there was a single larger pattern on one side and duplicated on the other. The basis for this pattern was a paisley-like shape of \textbf{A} units. This shape was designed in order to create greater dispersion of values across the $z_i$ triplets; that is, to minimize having tightly-clustered islands that would yield little in the way of \textbf{A}-\textbf{B}-\textbf{A} and \textbf{B}-\textbf{A}-\textbf{B} triplets ($z_2$ and $z_5$, respectively). 

The practical limitation of attempting to fit various ``islands'' of \textbf{A} nodes (black) into a surrounding ``sea'' of \textbf{B} nodes (white) meant that there were not quite enough \textbf{B} nodes to act as borders around the more compact sets of \textbf{A} nodes. Thus, the pattern in the right half of grid (\textbf{\textit{a}}) is a bit more compressed than originally planned. 

\begin{figure}[ht]
  \centering
  \fbox{
  \rule[-.5cm]{0cm}{4cm}\rule[-.5cm]{0cm}{0cm}	
  \includegraphics [trim=0.0cm 0cm 0.0cm 0cm, clip=true,   width=0.95\linewidth]{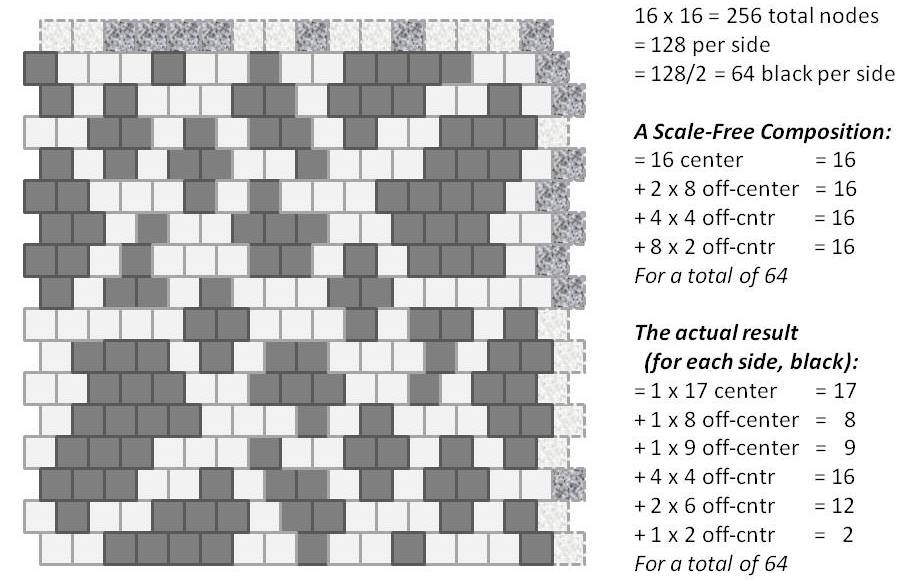}} 
  \vspace{3mm} 
  \caption{A 2-D CVM ``scale-free-like'' system with an equal number state \textbf{A} and state \textbf{B} nodes (128 nodes each).}   
\label{fig:CVM-2D_Scale-free_128-nodes_pg0_2016-10-26}
\end{figure}

The original plan was that out of 256 nodes in the grid, half (of the designed pattern) would be on the right, and half on the left; 128 nodes on each side. Of these, for each side, 64 nodes were to be in state \textbf{A} (black). Of these nodes (per side), sixteen (16 nodes) would be used to create a large, paisley-shaped island. The remaining 64 - 16 = 48 nodes would be used for smaller-sized islands; two islands of eight nodes each, etc. The plan is shown in Figure~\ref{fig:CVM-2D_Scale-free_128-nodes_pg0_2016-10-26}. The notation of ``center'' and ``off-center`` refers to the placement of the various islands; the largest (16-node) islands were to be placed more-or-less in the center of each of their respective (left or right) sides of the grid, and the remaining islands were to be ``off-center''; situated around their primary respective large islands.

The resulting patterns (shown in Figure~\ref{fig:CVM-2D_Scale-free_128-nodes_pg0_2016-10-26}) were close to the original plan, although not exactly the same.

\vspace{3mm} 

Even though some changes had to be made to the original design plan, the original constraint, that the number of units in states \textbf{A} and \textbf{B} would be identical (128 nodes in each), was kept. The details are shown in Figure~\ref{fig:CVM-2D_Scale-free_128-nodes_pg0_2016-10-26}.

\begin{figure}[ht]
  \centering
  \fbox{
  \rule[-.5cm]{0cm}{4cm}\rule[-.5cm]{0cm}{0cm}	
  \includegraphics [trim=0.0cm 0cm 0.0cm 0cm, clip=true,   width=0.95\linewidth]{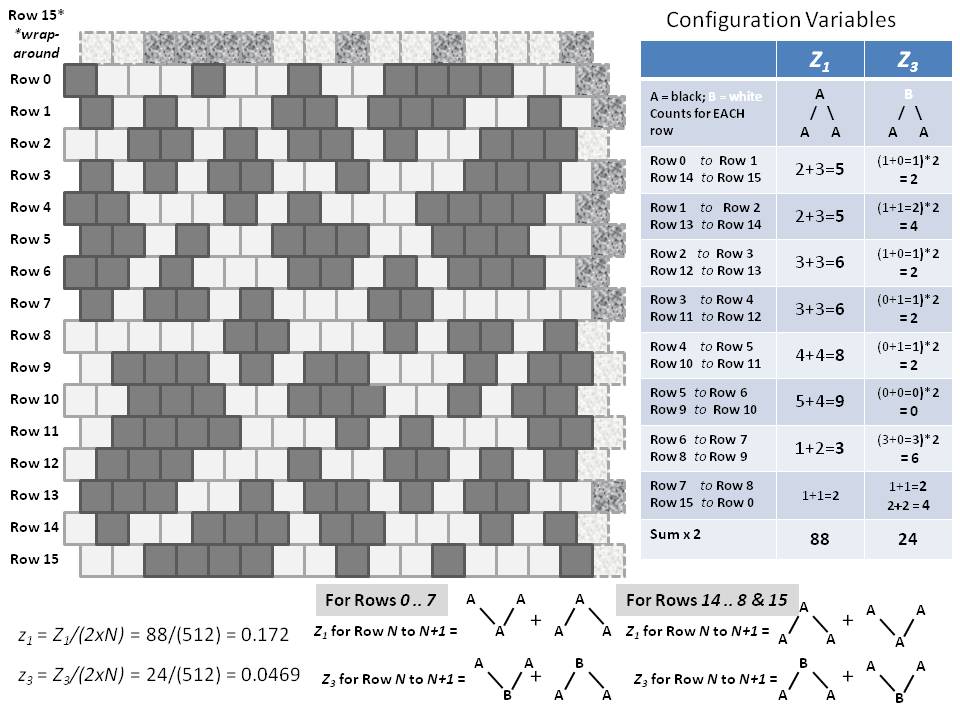}} 
  \vspace{3mm} 
  \caption{A 2-D CVM ``scale-free-like'' system with an equal number state \textbf{A} and state \textbf{B} nodes (128 nodes each).}   
\label{fig:CVM-2D_scale-free_2017-12-12}
\end{figure}
\vspace{3mm} 

The validation step for this stage of code development was to manually count all the configuration variables for several different configuration grids, such as the ones shown in Figure~\ref{fig:CVM-2-D_rich-club_and_scale-free_256-nodes}.

The counts for the ``scale-free-like'' grid shown in Figure~\ref{fig:CVM-2D_Scale-free_128-nodes_pg0_2016-10-26} are shown in Figure~\ref{fig:CVM-2D_scale-free_2017-12-12}. It suffices to say that the results from the manual counting (of all configuration variables) and those created by the computer code were identical. These held true across several different grids with different node configurations. 

\textbf{\textit{Note:}} To achieve the fractional variables shown in Figure~\ref{fig:CVM-1-D_base-graph2}, and also in  Table~\ref{tbl:config-variables-table}, the following relations are used: 

\begin{itemize}
\setlength{\itemsep}{1pt}
\item $x_i = X_i$, 
\item $y_i = Y_i/2$, for $i = 1, 3$ and $y_2 = Y_2/4$, accounting for the degeneracy with which $y_2$ occurs,
\item $w_i = W_i/2$, for $i = 1, 3$ and $w_2 = W_2/4$, accounting for the degeneracy with which $w_2$ occurs, and 
\item $z_i = Z_i/2$, for $i = 1, 3, 4, 6$ and $z_2 = Z_2/4$, $z_5 = Z_5/4$, accounting for the degeneracy with which $z_2$ and $z_5$ occur. 
\end{itemize}
 
\begin{figure}[ht]
  \centering
  \fbox{
  \rule[-.5cm]{0cm}{4cm}\rule[-.5cm]{0cm}{0cm}	
  \includegraphics [trim=0.0cm 0cm 0.0cm 0cm, clip=true,   width=0.95\linewidth]{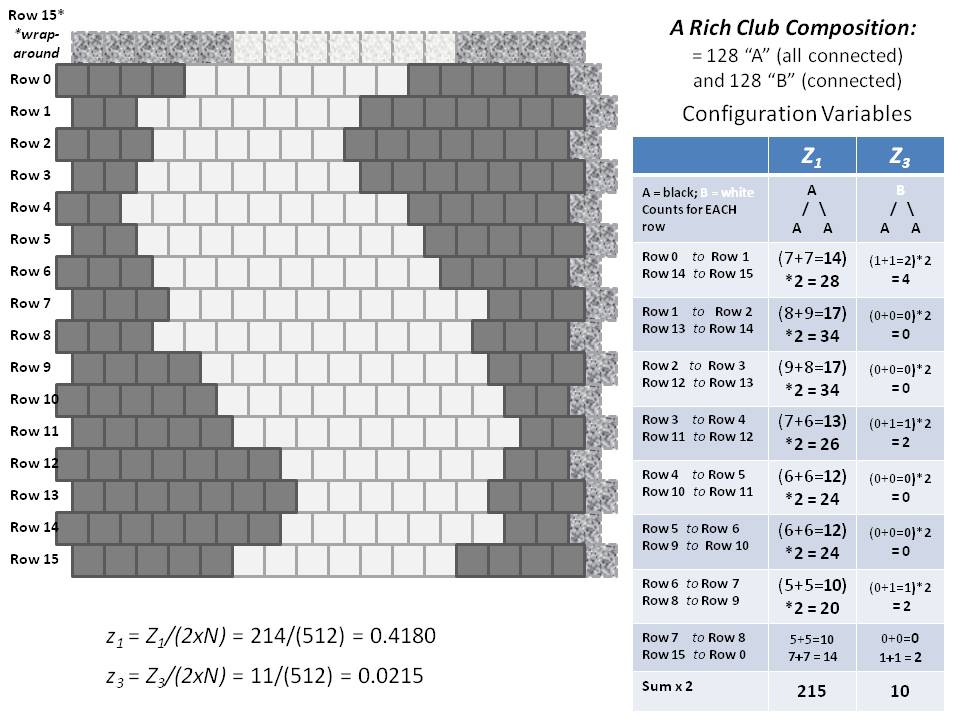}} 
  \vspace{3mm} 
  \caption{A 2-D CVM ``extreme-rich-club-like'' system with an equal number state \textbf{A} and state \textbf{B} nodes (128 nodes each).}   
\label{fig:CVM-2D_rich-club_cleaned-up_v2_2017-12-12}
\end{figure}

\textbf{\textit{Note:}} The exact details of the row counts are difficult to read in Figures~\ref{fig:CVM-2D_scale-free_2017-12-12} and~\ref{fig:CVM-2D_rich-club_cleaned-up_v2_2017-12-12}; the original diagrams are in a corresponding slidedeck that is available in the associated GitHub repository \cite{AJMaren_2019_Expt-Results_Two-epsilon-params}. 

\textbf{\textit{Note:}} The count for the $z_i$ variables is approximate; not exact. A follow-on code analysis revealed that while the counting steps for the $x_i$, $y_i$, and $w_i$ configuration variables was precise, the counting for the $z_i$ configuration variables was done only across the horizontally-expressed variables, and did not include the vertical versions. This was true for both the computer code and the manual counting. Because the size and diversity of patterns within any of the testing grids was sufficient to give a reasonably accurate result for the $z_i$, the code was kept as-is. The decision to use the approximate values for the $z_i$ was supported via manual counts on some very small-scale 2-D grids. Another reason to stay with the current code (for approximate results) is that the next step will be a transiston to a full object-oriented approach, and the time spent on code revision would be best served by moving on to the next stage. 

The second configuration, for an ``extreme-rich-club-like'' configuration, is shown in Figure~\ref{fig:CVM-2D_rich-club_cleaned-up_v2_2017-12-12}.

\vspace{3mm} 

As a contrast to the ``scale-free-like'' grid configuration used in Figures~\ref{fig:CVM-2D_Scale-free_128-nodes_pg0_2016-10-26} and~\ref{fig:CVM-2D_scale-free_2017-12-12}, the contrasting experiment used a second configuration that had only one large compact region of nodes in state \textbf{A}, which was wrapped-around the grid envelope, as shown in Figure~\ref{fig:CVM-2D_rich-club_cleaned-up_v2_2017-12-12}. This configuration was designed to maximize the number of pairwise and triplet configurations that put ``like-near-like.'' The previous configuration, shown in Figure~\ref{fig:CVM-2D_Scale-free_128-nodes_pg0_2016-10-26}, was more in the direction of ``like-near-unlike.'' 

The purpose of having patterns with such different dispersions among the configuration variable values was that they would putatively correspond to different \textit{h}-values. This means that they would (ultimately) converge to different points on an equilibrium curve for the free energy equation (in the case of equiprobable units in states \textbf{A} and \textbf{B}). 

We have the analytic results for the free energy minimum curve; specifically the equilibrium point for the free energy at different \textit{h}-values, or interaction enthalpy values. (The details for the free energy will be presented in the next section.) Thus, knowing what the anticipated \textit{h-value} would be for a given pattern, it would be interesting to find the \textit{actual} set of configuration variable values that result when the initial 2-D CVM grid is brought to free energy equilibrium, for the identified \textit{target h-value}. 

These experiments would also serve the verification and validation (V \& V) process in vetting the code. See Maren (2018) \cite{AJMaren-TR2018-001v2-V-and-V} for the V \& V Technical Report, and Maren (2019) for a slidedeck (available on GitHub) capturing details of the experimental results \cite{AJMaren_2019_Expt-Results_Two-epsilon-params}. 

By far, the most complex element of the ``configuration variable counting'' code was in counting the triplets. The V\&V step ensured that the counts wrapping around from right to left, and from top to bottom (creating a completely-wrapped envelope of the initial 2-D grid) performed as desired and expected. (See the \textit{Note} earlier in this section; the counts for the $z_i$ variables are done in the horizontal direction only, for both the code and manual verification.)

%
\section{Topographies: Interpreting configuration variables}
\label{sec:topographies-config-vars-interpretation}
%

When we work with a typical Ising system model, we can easily interpret the results in terms of simple dependence of various functions on a single variable, $x_1$. The topographic organization of the system, in this simple case, is not important. 

In contrast, when we work with a 2-D CVM, \textbf{\textit{the topographic organization is all-important}}. Further, we have fourteen different configuration variables that collectively describe this topography. This collection of variable values is a bit too much to keep in mind!

Thus, we need some guidelines for interpreting the system, given the set of configuration variables that are achieved at a free energy equilibrium point. Also, we need to reduce the number of variables that we think about to an absolute minimum, so that we can mentally capture the essentials of a given free energy solution. 

Also, we need to be able to mentally correlate this very small set of ``think-worthy'' configuration variables with some notion of what the topography would be like. We understand, of course, that we won't be envisioning a \textit{specific} topology; rather, we would be envisioning a \textit{kind of} topology. 

What we will move towards is a reduced set of just \textbf{\textit{three configuration variables}} that we need to consider, if we want a fairly good mental approximation of the 2-D CVM grid topography. 

With this in mind, we break this process down to two manageable steps: 

\begin{enumerate}
\item \textbf{\textit{Reduce the number of ``think-worthy'' configuration variables}} to an absolute minimum, and
\item \textbf{\textit{Interpret the meaning}} of each of these few configuration variables. 
\end{enumerate}

As noted above, there are a total of fourteen different configuration variables; two $x_i$, three each for the $y_i$ and $w_i$, and six of the $z_i$. 

Fortunately, we don't need to think about all fourteen configuration variables. In fact, we can get a very good sense of the 2-D CVM grid compostion by just considering three configuration variables; $z_1$, $z_3$, and $y_2$. These are selected because, taken together, they are reasonably descriptive of the grid topography: 

\begin{itemize}
\item $z_1$ - \textbf{A}-\textbf{A}-\textbf{A} triplets; indicates the relative fraction of \textbf{A} units that are included in ``islands'' or ``land masses''; this also (indirectly) indicates the compactness of these masses,
\item $z_3$ - \textbf{A}-\textbf{B}-\textbf{A} triplets; indicates the relative fraction of \textbf{A} units that are involved in a ``jagged'' border (one that involves irregular protrusions of \textbf{A} into a \textbf{B} space), or the presence of one or more ``rivers'' of \textbf{B} units extending into landmass(es) of \textbf{A} units, and
\item $y_2$ - \textbf{A}-\textbf{B} nearest-neighbor pairs; indicates the relative extent to which the \textbf{A} units are distributed among the surrounding \textbf{B} units. A higher $y_2$ value indicates lots of boundary areas between \textbf{A} and \textbf{B}, and a smaller value indicates large ``landmasses`` of \textbf{A} units.
\end{itemize}

%
\subsection{First simplification: consider the equiprobable case }
\label{subsec:first-simplification}
%

We can greatly simplify the complexity of our task by dealing with the \textit{equiprobable distribution} case, where  $x_1 = x_2 = 0.5$. We are doing this, not only because it makes it easier to think about the system, but because at this equiprobable distribution point, \textbf{\textit{we can achieve an analytic solution for the configuration variables}}. 

Further, at this equiprobable distribution point, we can also \textbf{\textit{invoke certain equalities among the configuration variables, due to symmetry considerations}}. Thus, at this point, we would have:

\begin{itemize}
\item $x_1 = x_2$ - units in state \textbf{A} equal those in state \textbf{B},
\item $y_1 = y_3$ - nearest-neighbor pairs \textbf{A}-\textbf{A} equal \textbf{B}-\textbf{B},
\item $w_1 = w_3$ - next-nearest-neighbor pairs \textbf{A}- -\textbf{A} equal \textbf{B}- -\textbf{B},
\item $z_1 = z_6$ - triplets \textbf{A}-\textbf{A}-\textbf{A} equal \textbf{B}-\textbf{B}-\textbf{B},
\item $z_2 = z_5$ - triplets \textbf{A}-\textbf{B}-\textbf{A} equal \textbf{B}-\textbf{A}-\textbf{B}, and
\item $z_3 = z_4$ - triplets \textbf{A}-\textbf{A}-\textbf{B} equal \textbf{A}-\textbf{B}-\textbf{B}. 
\end{itemize}

Thus, if we have the values for $x_1$, $y_1$, $w_1$, $z_1$, $z_2$, and $z_3$, then we also have the values for  $x_2$, $y_3$, $w_3$, $z_6$, $z_5$, and $z_4$.

Further, our normalization constraints also mean that if we know $y_1$ (and thus also $y_3$), we also know $y_2$, because we have $y_2 = 0.5(1.0 - y_1 -y_3)$. Similarly, we also would have $w_2$. 

Also, we can rewrite these expressions; if we have $y_2$, then we have $y_1 = y_3 = 0.5 - y_2$, and similarly for $w_1$ and $w_3$ if we have $w_2$. 

%
\subsection{Simplifying the numbers of $z_i$ }
\label{subsec:simplifying-z}
%

Similarly, we have the normalization for the $z_i$ configuration variables given as

\begin{equation}
  1=x_1+x_2 =\displaystyle\sum\limits_{i=1}^6 \gamma_i z_i.
\end{equation}\\ 

Thus, in the equiprobable case, we have 

\begin{equation}
  0.5 =\displaystyle\sum\limits_{i=1}^3 \gamma_i z_i = z_1 + 2z_2 + z_3.
\end{equation}\\ 

Thus, if we know $z_1$ and $z_3$, we also know $z_2$, given as $z_2 = 0.25 - (z_1 + z_3)/2$. 

In short, at the equiprobable distribution point (where $x_1 = 0.5$), if we know $y_1$, $w_1$, $z_1$, and $z_3$, then we also know the remaining configuration variables. 

%
\subsection{Relationship between $y_i$ and $w_i$}
\label{subsec:relating-y-and-w}
%

As one more step, we will note that, most of the time, $y_i$ and the corresponding $w_i$ will be approximately close to each other. In particular, the nearest-neighbor configuration variable $y_2$ indicates the extent of like-near-unlike boundaries. In the very extreme case where the units are arranged in a ``checkerboard''-like configuration, then $y_2$ will approach its maximum of $0.5$ (and $2y_2 \rightarrow 1.0$), while the values for $y_1$ and $y_3$, indicating like-near-like nearest neighbors, will both approach $0$. 

(As a side note, it is worthwhile to make sure that we don't push the system so far that either $y_1$ or $y_3$ approaches $0$, as the entropy term involves multiple instances of computing $v \ln(v) - v$, where $v$ refers to a configuration variable, and we don't want to go to the extreme of taking the logarithm of $0$.)

At the other extreme, a very small value for $y_2$ indicates that the system is composed almost exclusively of a very large mass of \textbf{A} units, existing as a single ``continent'' in a sea of \textbf{B} units. There is no single minimum value for $y_2$, as it really becomes a function of the overall number of units in the system; as we create a progressively larger and larger grid, the fraction of units on the boundary of a large \textbf{A} continent becomes smaller in comparision to the total number of units actually in the interior of this large mass (with progressively more \textbf{A}-\textbf{A} nearest neighbors).

The $w_i$ values, representing next-nearest-neighbor pairs, will not exactly be the same as the $y_i$. However, in the case of having large ``landmasses`` of \textbf{A} in a sea of \textbf{B}, then (similar to the $y_1$ and $y_3$), the values for $w_1$ and $w_3$ will each approach their limits of $0.5$, and the value for $w_2$ will shrink (getting close to but not reaching $0$). 

As visual interpretations, we can think that $y_2$ reflects the relative proportion of like-near-unlike boundary pairs across the diagonals, and $w_2$ reflects the like-near-unlike boundary pairs along the horizontal and vertical connections. Thus, in Figure~\ref{fig:CVM-2D_rich-club_cleaned-up_v2_2017-12-12} (with a large landmass of \textbf{A} units, wrapped around from right-to-left, so it appears as two distinct halves in the figure), we have relatively small values for both  $y_2$ and  $w_2$. 

In contrast, in the earlier Figure~\ref{fig:CVM-2D_Scale-free_128-nodes_pg0_2016-10-26}, we have lots of small `islands'' of \textbf{A} units scattered throughout the sea of \textbf{B}; both $y_2$ and $w_2$ are substantively increased. 

Similar to $y_1$, $z_1$ indicates the extent of large, contiguous ``landmasses`` of \textbf{A} units. In particular, it indicates the \textit{compactness} of these landmasses. 

The $z_3$ value reflects less the presence of a boundary between masses of \textbf{A} and  \textbf{B} units, and more the complexity of the boundary. The $z_3$ value also increases when we have long, spider-leg-like connections between masses or islands of \textbf{A} units. (Similarly, the value for $z_4$ increases when we have ``rivers`` of \textbf{B} appearing within the \textbf{A} landmasses.)

%
\subsection{Single starting point, two resulting topographies}
\label{subsec:two-resulting-topographies}
%

Although we will have a more detailed description later, we will get a better understanding of the relationships among the various configuration variables by introducing two comparative examples here. This work was done as Trial 2 during the experimental process.

We begin with the scale-free-like initial grid, introduced earlier in Figure~\ref{fig:CVM-2D_Scale-free_128-nodes_pg0_2016-10-26}. We will bring this initial grid to a free energy minimum point, for each of two different \textit{h-values}: 

\begin{itemize}
\item Trial 2.a: $h = 1.65$, which should push the system towards increased like-near-like clustering, and
\item Trial 2.b: $h = 1.16$, which should push the system towards like-near-unlike pairs.
\end{itemize}

\textbf{Trial 2.a}

\begin{figure}[h!]
  \centering
  \fbox{
  \rule[-.5cm]{0cm}{4cm}\rule[-.5cm]{0cm}{0cm}	
  \includegraphics [trim=0.0cm 0cm 0.0cm 0cm, clip=true,   width=0.8\linewidth]{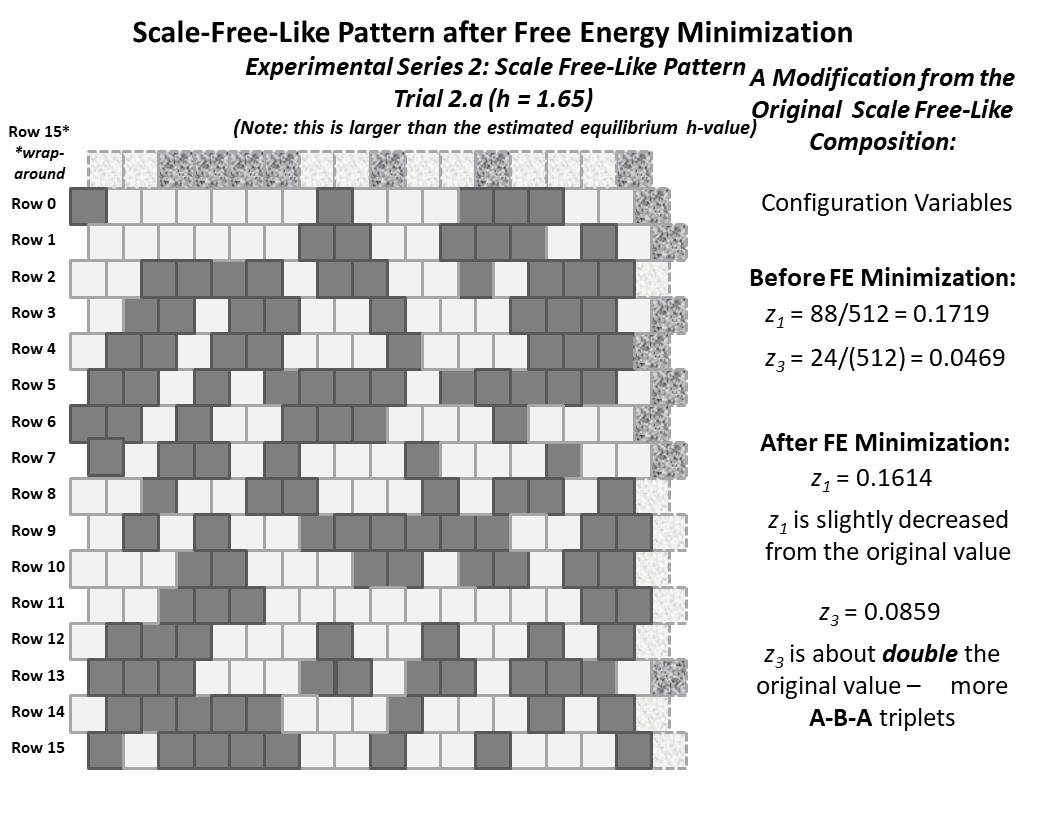}} 
  \vspace{3mm} 
  \caption{A 2-D CVM ``scale-free-like'' system, originally presented in Figure~\ref{fig:CVM-2D_Scale-free_128-nodes_pg0_2016-10-26}, brought to a free energy minimum with $h =  1.65$.}   
\label{fig:2D-CVM_Expts-scale-free-h-eq-1pt65_2019-06-17v4}
\end{figure}

When we bring the system from  Figure~\ref{fig:CVM-2D_Scale-free_128-nodes_pg0_2016-10-26} to free energy equilibrium with $h=1.65$, we get the result shown in Figure~\ref{fig:2D-CVM_Expts-scale-free-h-eq-1pt65_2019-06-17v4}. 

\vspace{3mm} 

We see in Figure~\ref{fig:2D-CVM_Expts-scale-free-h-eq-1pt65_2019-06-17v4} that, as expected, the clusters are coalesced as compared with the original pattern. 
\vspace{6 mm}

\textbf{Trial 2.b}

\begin{figure}[h!]
  \centering
  \fbox{
  \rule[-.5cm]{0cm}{4cm}\rule[-.5cm]{0cm}{0cm}	
  \includegraphics [trim=0.0cm 0cm 0.0cm 0cm, clip=true,   width=0.8\linewidth]{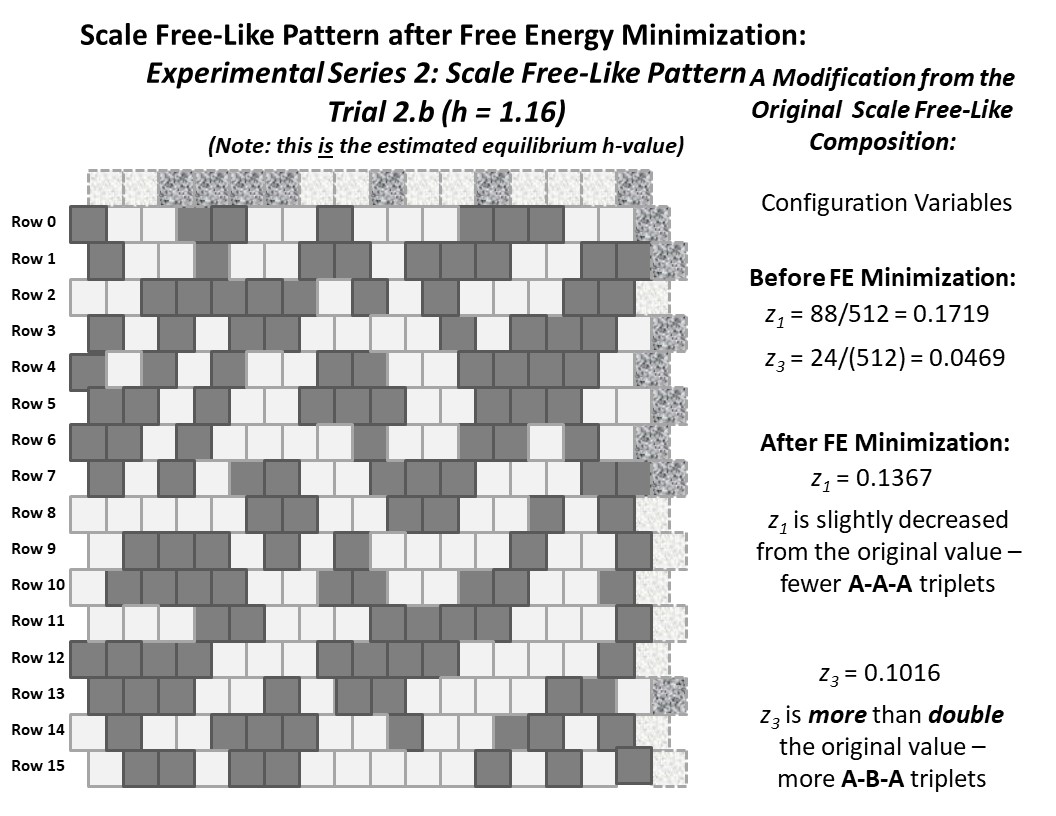}} 
  \vspace{3mm} 
  \caption{A 2-D CVM ``scale-free-like'' system, originally presented in Figure~\ref{fig:CVM-2D_Scale-free_128-nodes_pg0_2016-10-26}, brought to a free energy minimum with $h =  1.16$.}   
\label{fig:2D-CVM_Expts-scale-free-h-eq-1pt16_2019-06-17v4}
\end{figure}

When we bring the system from  Figure~\ref{fig:CVM-2D_Scale-free_128-nodes_pg0_2016-10-26} to free energy equilibrium with $h=1.16$, we get the result shown in Figure~\ref{fig:2D-CVM_Expts-scale-free-h-eq-1pt16_2019-06-17v4}. 

\vspace{3mm} 

With a lower \textit{h-value} than was previously used, we expect less clustering of like-with-like, and more spread-out connections. What we see is interesting - not surprising - but it is worth noting that we get a number of ``spider-leg'' connections between what remains from the original clusters. 

These ``spider-legs'' give us \textit{both} like-near-like (extending the length of the spider-leg) as well as like-near-unlike (on either side of the spider-leg) connections for each unit in the leg. Thus, each unit in a spider-leg gives us approximately the ratio  of $y_1$ and $y_2$ pairwise connections (two $y_2$ connections per $y_1$, accounting for the factor of two degeneracy in the $y_2$ pairs) that we'd expect near $h = 1.0$.

%
\section{The 2-D CVM Free Energy }
\label{sec:2D-CVM-free-energy}
%

The essential notion of the CVM is that we work with a more complex expression for the free energy in a system. 

As a point of comparison, the basic Ising equation is

\begin{equation}
\label{Bar-F-basic-Ising-basic-eqn}
\bar{F} = F/(Nk_{\beta}T) =  \bar{H}  - \bar{S},
\end{equation}

\noindent
where $F$ is the free energy, $H$ is the enthalpy and $S$ is the entropy for the system, and where $N$ is the total number of units in the system, $k_\beta$ is Boltzmann's constant, and $T$ is the temperature. 

For working with abstract systems, the total $Nk_{\beta}T$ can be absorbed into a \textit{reduced energy formalism}, as these values are constants during system operations. This leads to the \textit{reduced representations} of $\bar{F}$, $\bar{H}$, and $\bar{S}$. We will work consistently with reduced representations throughout this work. 

In a simple Ising model, both the enthalpy $H$ and the entropy $S$ can be computed based on only the relative fraction of active units in a bistate system. That is, there are only two kinds of computational units; \textit{active} ones in state \textbf{A}, where the fraction of these units is denoted $x_1$, and \textit{inactive} ones in state \textbf{B}, where the fraction of these units is denoted $x_2$. (Of course, $x_1 + x_2 = 1.0$.)

In contrast to the simple entropy used in the basic Ising model in statistical mechanics, in the CVM approach, we expand the entropy term. The CVM entropy term considers not only the relative fractions of units in states \textbf{A} and \textbf{B}, but also that associated with the \textit{configuration variables}, as described earlier in Section~\ref{sec:config-variables}.)

We can write the 2-D CVM free energy, using the formalism first introduced by Kikuchi in 1951 \cite{Kikuchi_1951_Theory-coop-phenomena}, and then further advanced by Kikuchi and Brush (1967) \cite{Kikuchi-Brush_1967_Improv-CVM} (explicitly for the 2-D CVM), as

\begin{equation}
\label{eqn:Bar-F-2-D-basic-eqn}
  \begin{aligned}
\bar{F}_{2-D} = F_{2-D}/N = \\
  & \bar{H}_{2-D}- \bar{S}_{2-D} \\
+ & \mu (1-\sum\limits_{i=1}^6 \gamma_i  z_i )+4 
\lambda (z_3+z_5-z_2-z_4),
  \end{aligned}
\end{equation}

\noindent
where $\mu$ and $\lambda$ are Lagrange multipliers, and we have set $k_{\beta}T = 1$.

%
\subsection{The 2-D CVM Enthalpy}
\label{subsec:2D-CVM-enthalpy}
%

The enthalpy in a simple Ising system is traditionally given as 

\begin{equation}
\label{eqn:Bar-H-2-D-basic-eqn}
  \begin{aligned}
	\bar{H}_{2-D} = H_{2-D}/N \\
  & = \bar{H}_0 + \bar{H}_1 \\
  & = \varepsilon_0 x_1 + c x_1^2,  \\
  \end{aligned}
\end{equation}

where $H_0$ and $c$ are constants.

The first term on the RHS (Right-Hand-Side) corresponds to the \textit{activation enthalpy}, or enthalpy associated with each active unit. The second term on the RHS corresponds to the \textit{interaction enthalpy}, or energy associated with pairwise interactions between active units. 

In contrast, the enthalpy for the 2-D CVM is given as 

\begin{equation}
\label{eqn:Bar-H-2-D-basic-eqn}
  \begin{aligned}
	\bar{H}_{2-D} = H_{2-D}/N \\
  & = \bar{H}_0 + \bar{H}_1 \\
  & = \varepsilon_0 x_1 + \varepsilon_1(-z_1+z_3+z_4-z_6) 
  \end{aligned}
\end{equation}

Note that in the original work by Kikuchi and Brush, Eqn.~\ref{eqn:Bar-H-2-D-basic-eqn} is simplified (K\&B Eqns. I.16 and I.17) to 

\begin{equation}
\label{eqn:Bar-H-2-D-basic-eqn-Kikuchi-and-Brush}
\bar{H}_{2-D} = \varepsilon_1(-z_1+z_3+z_4-z_6), 
\end{equation}

\noindent
that is, they omit the term linear in $x_1$; the activation enthalpy. 

We can infer that Kikuchi and Brush omit the activation enthalpy from their enthalpy term because they move directly to the analytic solution for the 2-D CVM free energy, which is solvable only in the equiprobable distribution case of $x_1 = x_2 = 0.5$. The equiprobable distribution is achieved only when the activation enthalpy is zero; that is, when $\varepsilon_0 = 0$. 

When the activation enthalpy $\varepsilon_0 > 0$, then the units in state \textbf{A} have an energy associated with them that is greater than that of the units in state \textbf{B}. Thus, an equilibrium solution will favor having fewer units in state \textbf{A}. There is no analytic solution for this case, other than that in which the interaction enthalpy is zero ($\varepsilon_1 = 0$). This latter case is trivial to solve, and is not particularly interesting, as the distribution of different kinds of nearest-neighbor pairs and triplets will be probabilistically random, excepting only that the proportions of units in states \textbf{A} and \textbf{B}, respectively, will be skewed by the activation enthalpy parameter $\varepsilon_0$.

As just noted, the typical expression for the interaction enthalpy is a quadratic term in $x_1$, that is, $H_1 = c x_1^2$. The parameter $c$ encompasses both the actual interaction energy for each pairwise interaction, and a constant that expresses the distribution of pairwise interactions as a simple linear function of the fraction of active units ($x_1$) surrounding a given active unit. This is then multiplied by the total fraction of active units, giving the quadratic expression. 

In the expression for the 2-D CVM interaction enthalpy, we have terms that expressly identifies the total fraction of nearest-neighbor ``unlike'' pairs ($y_2$) and ``like'' pairs ($y_1$ and $y_3$). Thus, we can replace $c x_1^2$ with the fraction of ``unlike`` pairs (counted twice, to account for the degeneracy in how these pairs can be counted), and the fractions of ``like'' nearest neighbor pairs. 

We recognize that this is a simplification; we are not counting interaction energies due to next-nearest neighbor pairs, the $w_i$, nor from the triplets $z_i$. We are, effectively, subsuming these into the pairwise interactions that are being modeled with the $y_i$. 

We take the interaction enthalpy parameter $\varepsilon_1$ to be a positive constant. 

We envision a system in which the free energy is reduced by creating nearest-neighbors of like units, that is, \textbf{A}-\textbf{A} or \textbf{B}-\textbf{B} pairs. (Decreasing the interaction enthalpy leads to decreasing the free energy, which is desired as we go to a free energy minimum, or equilibrium state.) Similarly, the interaction enthalpy should increase with unlike pairs, or \textbf{A}-\textbf{B} pairs (or vice versa). 

Thus, we write the interaction enthalpy equation, following the formalism introduced by Kikuchi and Brush, as 

\begin{equation}
\label{eqn:enthalpy-Kikuchi-Brush}
  \begin{aligned}
\bar{H}_{2-D} = H_{2-D}/N = 
\varepsilon_1(2y_2 - y_1 - y_3). 
  \end{aligned}
\end{equation}

We interpret this equation by noting that as we increase the fraction of unlike unit pairings (\textbf{A}-\textbf{B} or \textbf{B}-\textbf{A} pairs), we raise the interaction enthalpy. At the same time, if we're increasing \textit{unlike} unit pairings, we are also decreasing \textit{like} unit pairings (\textbf{A}-\textbf{A} and \textbf{B}-\textbf{B}), so that we are again increasing the overall interaction energy. (Note that the \textit{like} unit pairings show up in Eqn.~\ref{eqn:enthalpy-Kikuchi-Brush} with a negative sign in front of them.)

Thus, if we create a 2-D CVM that is like a checkerboard grid (alternating unit types), then we are moving to a higher interaction enthalpy and higher free energy, and away from a free energy minimum. If we create a grid such as that shown in the RHS of Figure~\ref{fig:CVM-2-D_rich-club_and_scale-free_256-nodes}, with all of the units in each state grouped together, respectively, then we lower the interaction enthalpy and correspondingly lower the free energy. 

Clearly, if the interaction enthalpy were all that mattered, we would have a 2-D CVM grid like the one on the right in Figure~\ref{fig:CVM-2-D_rich-club_and_scale-free_256-nodes}. What keeps this from happening is the entropy term, which demands some distribution over different possible configurations. 

Before moving on to the entropy, we note that we can express the interaction enthalpy using the triplet configuration variables $z_i$, instead of the nearest-neighbor pair variables $y_i$. We can do this by drawing on \textit{equivalence relations} between the $y_i$ and $z_i$ variables. Those for $y_2$ are given as

\begin{eqnarray}
\label{eqn:equivalence-relationships-y2-and-z}
  y_2 = z_2+z_4 = z_3+z_5 \\
  2 y_2 = z_2+z_4 + z_3+z_5.   
\end{eqnarray}

Notice that we have two ways of expressing $y_2$ in terms of the $z_i$, as shown in Eqn.~\ref{eqn:equivalence-relationships-y-and-z}. Since we want to work with the total $2y_2$, it is easy to express that as the sum of the two different equivalence expressions. This will prove useful later, when we analytically solve for the free energy minimum, or equilibrium point. 

We also have equivalence relations for $y_1$ and $y_3$, given as

\begin{equation}
\label{eqn:equivalence-relationships-y2-and-z}
 y_1 = z_1 + z_2
\end{equation}

\noindent
and

\begin{equation}
\label{eqn:equivalence-relationships-y2-and-z}
 y_3 = z_5 + z_6.
\end{equation}

This lets us write 

\begin{equation}
\label{Eqn:z3-analyt1-current-approach}
  \begin{aligned}
\bar{H}_{2-D} =
\varepsilon_1(2y_2 - y_1 - y_3) =
\varepsilon_1(z_4+z_3-z_1-z_6). 
  \end{aligned}
\end{equation}

As a minor note, the enthalpy used in previous related work by Maren \cite{AJMaren-TR2014-003, Maren_2016_CVM-primer-neurosci}, was

\begin{equation}
\label{Eqn:z3-analyt2-previous-approach}
  \begin{aligned}
\bar{H}_{2-D} = H_{2-D}/N =
\varepsilon_1(2y_2) = \varepsilon_1(z_2+z_3+z_4+z_5). 
  \end{aligned}
\end{equation}

The results given here are similar in form to the results presented in the two previous works by Maren; they differ in the scaling of the interaction enthalpy term.

%
\subsection{The 2-D CVM Entropy}
\label{subsec:2D-CVM-entropy}
%

The entropy for the 2-D CVM is given as 

\begin{equation}
\label{eqn:Bar-S-2-D-basic-eqn}
  \begin{aligned}
\bar{S}_{2-D} = S_{2-D}/N = \\
 & 2 \sum\limits_{i=1}^3 \beta_i Lf(y_i))
          + \sum\limits_{i=1}^3 \beta_i Lf(w_i) \\
 &      - \sum\limits_{i=1}^2 \beta_i Lf(x_i)
          - 2 \sum\limits_{i=1}^6 \gamma_i Lf(z_i), \\
  \end{aligned}
\end{equation}

 \noindent
where $Lf(v)=vln(v)-v$.

There were two primary means for obtaining validation that the code computing the thermodynamic variables was correct: 

\begin{enumerate} \itemsep0pt 
\item \textbf{Comparison with analytic for the equiprobable case} -- for the case equiprobable distribution among the $x_i$ variables ($x_1 = x_2 = 0.5$), I have developed an analytic solution, which gives a means for comparing the code-generated results against the expected (analytic) results, and 
\item \textbf{Comparison with analytic for the case where the interaction enthalpy is zero} -- the second means to check the code-generated results is for the case where the distributuion of $x$ values is not equiprobable, however the interaction enthalpy is set to zero ($h = 1$), and thus the exact distribution of other configuration values can be precisely computed, allowing further for exact analytic computation of thermodynamic variables. 
\end{enumerate}

%
\subsection{Free Energy Analytic Solution}
\label{subsec:free-energy-analytic-solution}
%

Kikuchi and Brush (1967) \cite{Kikuchi-Brush_1967_Improv-CVM} provided the results of an analytic solution for the 2-D CVM free energy, for the specific case where $x_1 = x_2 = 0.5$. Specifically, making certain assumptions about the Lagrange multipliers shown in Eqn.~\ref{eqn:Bar-F-2-D-basic-eqn}, we can then express each of the configuration variables in terms of $\varepsilon_1$. 

More usefully, since the expression actually involves the term $exp(2\varepsilon_1)$, and not $\varepsilon_1$ itself, it is much easier to use the substitution variable $h = exp(2\varepsilon_1)$. We refer to $h$ (or sometimes, the \textit{h-value}), as the \textit{interaction enthalpy parameter} throughout. 

The full derivation of the set of equations giving the configuration variable values at equilibrium (i.e., at $x_1 = x_2 = 0.5$) is given in Appendix A.

The full set of these analytic expressions for the various configuration variables is couched in terms of a denominator involving \textit{h}, specifically 

\begin{equation}
\Delta =-h^2 +6h - 1,
\label{Kikuchi-and-Brush-Delta-def-Eqn-Ipt24-main-text}
\end{equation}

\noindent
which Kikuchi and Brush present as their Eqn. (I.24) \cite{Kikuchi-Brush_1967_Improv-CVM}. 

We recall that at the equiprobable distribution point, where $x_1 = x_2$, we have a number of other equivalence relations, e.g. $z_1 = z_6$, etc. 

We then (following Kikuchi and Brush, in their Eqn.~(I.25)) identify each of the remaining configuration variables as

\begin{equation}
\begin{array}{lll}
  y_1 = y_3 &=&  \frac {( 3h  - 1)} { 2 \Delta}\\
  y_2  &=&  \frac {h( -h  +  3) } { 2 \Delta}\\
  w_1 =w_3 &=& \frac {( h  +  1)^2 } { 4 \Delta}\\ 
  w_2  &=&  \frac {(3h-1)( -h  +  3) } { 4 \Delta}\\
  z_1 = z_6 &=& \frac {(3h-1)( h  +  1) } { 8 \Delta}\\
  z_2 = z_5 &=& \frac {(3h-1)( -h  +  3) } { 8 \Delta}\\ 
  z_3 = z_4 &=&  \frac {( -h  +  3) ( h  +1)} { 8 \Delta}\\ 
 \end{array}
 \label{config-var-analytic-eqns-K-and-B-main-text}
\end{equation}

%
\subsection{Divergence in the analytic solution }
\label{subsec:divergence-analytic}
%

As is obvious from Eqn.~\ref{Kikuchi-and-Brush-Delta-def-Eqn-Ipt24-main-text}, there will be a divergence in the analytic solution for the configuration variables at the free energy minimum, because the analytic solution contains a denominator term that is quadratic in \textit{h}. Specifically, the term diverges for $h = 0.172$ or $h = 5.828$. 

We are interested in the latter case, where the value of $h>1$ indicates that $\varepsilon_1 >0$, which is the case where the interaction enthalpy favors like-near-like interactions, or some degree of gathering of similar units into clusters. This means that we expect that the computational results would differ from the analytic as $h \rightarrow 5.828$. (Actually, the computational results diverge from the analytic substantially before this point.)

This would possibly have some impact, but we find that the real point of interest comes from examining the impact of higher \textit{h-values} on the free energy itself. Increasing $h$ increases the overall value of the enthalpy term, and makes it overwhelmingly dominant with regard to the entropy term. 

%
\subsection{Significance of \textit{h} in the free energy}
\label{subsec:significance-of-h}
%

Because it is the \textit{minimum in the negative entropy} that creates the possibility for minimizing the free energy, we need to have the entropy term play a role that is at least on par with the enthalpy. If the enthalpy term is overall too large, then there will be an adverse effect on finding a useful free energy minimum. 

\begin{figure}[ht]
  \centering
  \fbox{
  \rule[-.5cm]{0cm}{4cm}\rule[-.5cm]{0cm}{0cm}	
  \includegraphics [trim=0.0cm 0cm 0.0cm 0cm, clip=true,   width=0.95\linewidth]{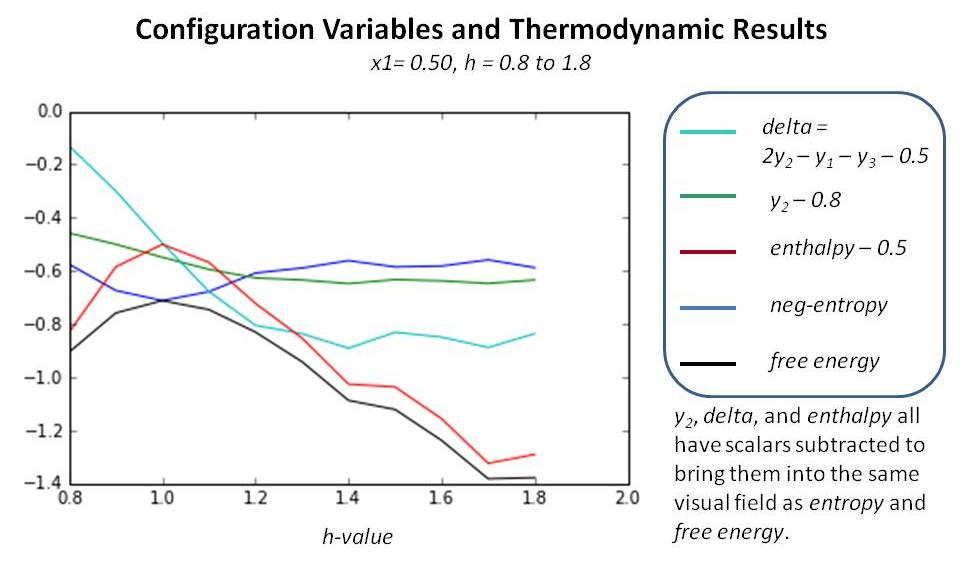}} 
  \vspace{3mm} 
  \caption{Configuration variable and thermodynamic values for the case where $x_1 = x_2 = 0.5$, and where the interaction enthalpy parameter $h$ ranges as $h = 0.8 .. 1.8$. See detailed explanation of results in the following Section~\ref{sec:dependence-config-vars}, as their nature is similar to these results. Appendix~\ref{sec:Appendix-D-experimental-results-varying-x1} also contains detailed material relating to this figure.}   
\label{fig:Perturbation-Results_x1=0pt50_lbld_crppd_2018-01-22}
\end{figure}

The impact of the \textit{interaction enthalpy parameter h} on the overall free energy is shown in Figure~\ref{fig:Perturbation-Results_x1=0pt50_lbld_crppd_2018-01-22}, which presents the results when $x_1 = 0.5$, which is the case where all of the results should conform with the analytic solution.

\vspace{3mm} 

We will discuss the results shown in Figure~\ref{fig:Perturbation-Results_x1=0pt50_lbld_crppd_2018-01-22} in more detail in Section~\ref{sec:intepretation}. First, we will put our attention on how, in general, we expect the \textit{h-values} to influence the configuration variable results. We will do this in Section~\ref{sec:dependence-config-vars}.

Before moving on, we briefly address the process for obtaining configuration variable values when we move away from the equiprobable case, which has been our mainstay up until now.

%
\subsection{Moving away from the equiprobable case }
\label{subsec:moving-away}
%

As we just saw in Subsection~\ref{subsec:divergence-analytic}, the analytic solution does not always hold. In fact, it has a divergence for a large value of the interaction enthalpy parameter. However, especially for low values of the interaction enthalpy, the analytic solution is a trustworthy starting place. 

When we move away from the equiprobable case, we need a mechanism by which the pattern of node activations on the grid can adjust in order to reach a free energy minimum. 

The code for accomplishing this works in two stages: 

\begin{enumerate} \itemsep0pt 
\item \textbf{\textit{Bring $x_1$ close to desired value}}, and
\item \textbf{\textit{Adjust configuration variables to achieve free energy minimum}}.
\end{enumerate} 

Early work on obtaining a more complete set of configuration variable and thermodynamic values was done as part of a set of \textit{perturbation experiments}. That is, in addition to bringing a system to an initial free energy minimum point, the system was then ``perturbed''. A fraction of the nodes were randomly selected for state-switching (from \textbf{A} to \textbf{B} and vice versa, keeping the relative proportions of nodes in each state the same). Then, the system was brought to a free energy minimum a second time. Visual inspection of the free energy graphs over time confirmed that the system tended to converge to approximately the same free energy minimum values. 

The V\&V aspects of this work are summarized in the V\&V Technical Report; Maren (2018) \cite{AJMaren-TR2018-001v2-V-and-V}. Appendix~\ref{sec:Appendix-D-experimental-results-varying-x1} presents detailed material results.

The actual protocol by which an initial grid is brought to a free energy minimum (equilibrium state) will be discussed in the Technical Report that will immediately follow this one; it will similarly be published to arXiv. This \textit{Protocol} document will provide detailed instructions and walk-throughs for working with a system designed to be at the equiprobable distribution point; that is, $\varepsilon_0 = 0$, but with interaction enthalpies conducive to forming like-near-like clusters; that is, $\varepsilon_1 = 0$.

%
\section{Dependence of the configuration variables on the enthalpy parameters}
\label{sec:dependence-config-vars}
%

From the analytic solution for the 2-D CVM, we can identify the configuration variables for a given \textit{h-value}. The dependence of $y_2$, $z_1$, and $z_3$ on \textit{h} is shown in the following Figure~\ref{fig:2D-CVM-config-vars-vs-h_redone_basic_crppd_2019-08-21}.

\begin{figure}[ht]
  \centering
  \fbox{
  \rule[-.5cm]{0cm}{4cm}\rule[-.5cm]{0cm}{0cm}	
  \includegraphics [trim=0.0cm 0cm 0.0cm 0cm, clip=true,   width=0.95\linewidth]{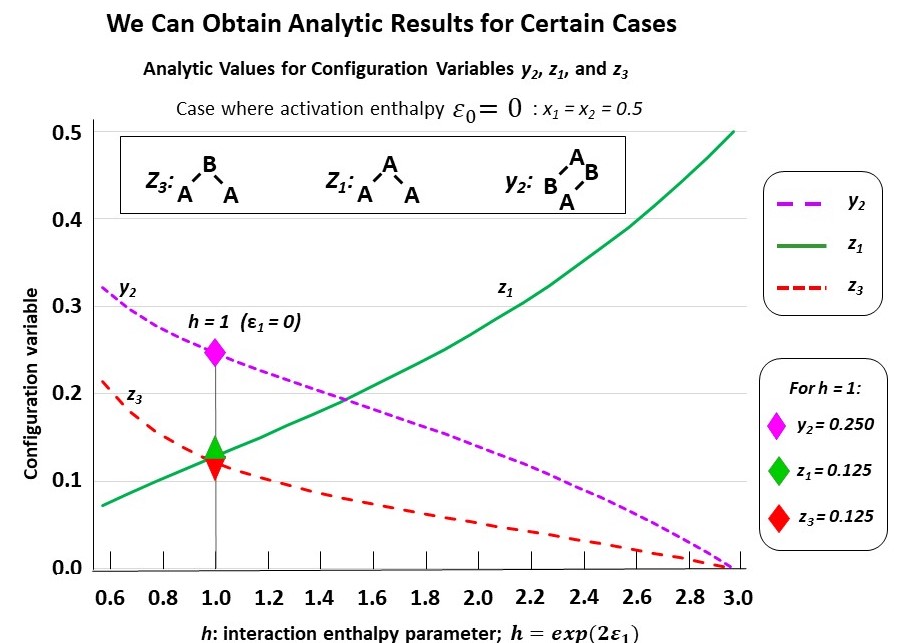}} 
  \vspace{3mm} 
  \caption{Dependence of the configuration variables $y_2$, $z_1$, and $z_3$ on the enthalpy parameter $h$, where $h = exp(2\varepsilon_1)$.}   
\label{fig:2D-CVM-config-vars-vs-h_redone_basic_crppd_2019-08-21}
\end{figure}
\vspace{3mm} 

The following three subsections assist in interpreting the results shown in Figure~\ref{fig:2D-CVM-config-vars-vs-h_redone_basic_crppd_2019-08-21}; these are the cases where the interaction enthalpy is defined as:

\begin{enumerate} 
\item	$\varepsilon_1 =0$ so that $h=e^{2\varepsilon_1} =1$,
\item $\varepsilon_1 <0$ so that $h=e^{2\varepsilon_1} <1$, and
\item  $\varepsilon_1 >0$ so that $h=e^{2\varepsilon_1} >1$.
\end{enumerate}

The material in these next three subsections was first in Maren (2016) \cite{Maren_2016_CVM-primer-neurosci}, describing the dependence of configuration variables on $h$ in a 1-D CVM system. The same essential arguments hold equally well for the 2-D CVM system, and are presented again here (with minor revisions) for completeness.

%
\subsection{Configuration variables when the interaction enthalpy $\varepsilon_1 = 0 $ ($h=1$)}
\label{subsec:config-var-interaction-enthalpy-zero}
%

We begin by noting (in the slightly left-of-center portion of Figure~\ref{fig:2D-CVM-config-vars-vs-h_redone_basic_crppd_2019-08-21}, where $h=1$) that the observed configuration variable values are exactly as we would expect. Specifically, $z_1 = z_3 = 0.125$, and $y_2 = 0.250$. This fits precisely with our expectation for the configuration variables at equilibrium when $\varepsilon_1 =0$ ($h=e^{2\varepsilon_1} =1$). 

To understand intuitively that this would be the expected result, we reflect on the total number of triplet states available, together with their degeneracies. We use various normalization equations from  Appendix~A as well, and particularly note that 

\begin{equation}
  1=x_1+x_2 =\displaystyle\sum\limits_{i=1}^6 \gamma_i z_i \nonumber
\end{equation}\\ 

\vspace{-12pt}

Since we also have $x_1 = x_2 = 0.5$, as well as the symmetry results from creating the equiprobability condition, we have 

\begin{equation}
  0.5 = y_1+y_2 = y_2+y_3 =\displaystyle\sum\limits_{i=1}^3 \gamma_i z_i \nonumber
\end{equation}\\
\vspace{-10pt}

We also have that the degeneracy factors are $\beta_2 = \gamma_2 = \gamma_5 = 2$, with all other $\beta_i$ and $\gamma_i$ set to 1. At equilibrium, with $\varepsilon_1 =0$, we expect equiprobable distribution of the $z_i$ among all possible states, so that we expect $z_1 = z_3 = 0.125$, as observed in Figure~\ref{fig:2D-CVM-config-vars-vs-h_redone_basic_crppd_2019-08-21}. Additionally (not shown in the figure), we would expect that $2z_2  = 0.250$, so that $z_1 +2 z_2 + z_3  = 0.5$.

%
\subsection{Configuration variables: $\varepsilon_1 <0$ ($h<1$)}
\label{subsec:config-var-interaction-enthalpy-less-than-zero}
%

At the left-hand-side of the preceding Figure~\ref{fig:2D-CVM-config-vars-vs-h_redone_basic_crppd_2019-08-21}, we have the case where $h=e^{2\varepsilon_1}<1$. These small values for $h$ means that $\varepsilon_1$, the interaction enthalpy between two unlike units (\textbf{A}-\textbf{B}), is negative. This further means that we stabilize the system by providing a structure that emphasizes alternate units; that is, by creating grid structures that look like \textbf{A}-\textbf{B}-\textbf{A}-\textbf{B}.

Note that in this realm, the pairwise combination $y_2$ (\textbf{A}-\textbf{B}) increases beyond the nominal expectation (when there is no interaction energy), so that $y_2 \to 0.5$, notably when $h < 0.4$. 

As a natural consequence of the increase in $y_2$ when $h \to 0$, we also have $z_3 \to 0.5$ (maximizing \textbf{A}-\textbf{B}-\textbf{A} triplets) and also $z_1 \to 0$ (minimizing \textbf{A}-\textbf{A}-\textbf{A} triplets). (Note that a divergence in the analytic solution means that we do not have analytic values for the various configuration variables for small values of $h$.)

%
\subsection{Configuration variables: $\varepsilon_1 >0$ ($h>1$)}
\label{subsec:config-var-interaction-enthalpy-greater-than-zero}
%

Consider the case of a positive interaction energy between unlike units (the \textbf{A}-\textbf{B} pairwise combination). This is the case on the right-hand side of Figure~\ref{fig:2D-CVM-config-vars-vs-h_redone_basic_crppd_2019-08-21}, where $\varepsilon_1 >0$ then yields $h=e^{2\varepsilon_1} >1$. 

The positive interaction energy ($\varepsilon_1 >0 $) suggests that a preponderance of \textbf{A}-\textbf{B} pairs ($y_2$) would destabilize the system, as each \textbf{A}-\textbf{B} pair would introduce additional enthalpy to the overall free energy. We want to bring the system to equilibrium, which is the free energy minimum. We would thus expect that as $\varepsilon_1$ increases as a positive value, we would see a decrease in $y_2$, and also see smaller values for those triplets that involve non-similar pair combinations. That is, we would expect that the \textbf{A}-\textbf{B}-\textbf{A} triplet, or $z_3$, would approach zero. 

This is exactly what we observe this on the right-hand side of the Figure~\ref{fig:2D-CVM-config-vars-vs-h_redone_basic_crppd_2019-08-21}. 

Thus, when ${h >> 1}$ (or ${\varepsilon_1 >> 0}$), we see that $z_3$ falls towards zero, and $y_2$ decreases as well. Note that $y_2$ can never go towards zero, because there will always be some \textbf{A}-\textbf{B} pairs in a system that contains units in a mixture of states \textbf{A} and \textbf{B}.

Correspondingly, this is also the situation in which $z_1 = z_6$ becomes large; for example, we see that $z_1 > 0.4$ when $h > 3.0$. (In fact, the analytic solution suggests that  $z_1 = z_6 = 0.5$, which we know is not feasible; as mentioned earlier, the analytic solution is not accurate for extrema values of $h$.)

Notice that $z_1$ cannot truly approach $0.5$, as $z_3$ did for the case where ${h << 1}$. The reason is that $z_1$ represents \textbf{A}-\textbf{A}-\textbf{A} triplets, and we will always have some \textbf{A}-\textbf{A}-\textbf{B} and \textbf{B}-\textbf{A}-\textbf{A} triplets, because we have an equiprobable distribution of units into states \textbf{A} and \textbf{B}. Thus, while we could conceivably have constructed a system composed exclusively of \textbf{A}-\textbf{B}-\textbf{A} and \textbf{B}-\textbf{A}-\textbf{B} triplets, we cannot do so exclusively with \textbf{A}-\textbf{A}-\textbf{A} and \textbf{B}-\textbf{B}-\textbf{B} triplets.

Despite this limitation, the realm of $h >> 1$ is one in which we observe a highly structured system where large ``domains'' of like units mass together. We would expect these large domains (each comprised of either overlapping \textbf{A}-\textbf{A}-\textbf{A} or \textbf{B}-\textbf{B}-\textbf{B} triplets) to stagger against each other, with relatively few ``islands'' of unlike units (e.g.,~the \textbf{A}-\textbf{B}-\textbf{A} and \textbf{B}-\textbf{A}-\textbf{B} triplets). 

In actuality, the patterns that emerge for relatively high \textit{h-values} still contain ``spiderlegs'' composed of triplets of unlike units. The actual observed topographies are discussed later in this work.

%
\section{Interpreting the computational results}
\label{sec:intepretation}
%

One important question, since we are taking a simple computational ``find-and-flip'' for finding the free energy minimum in a 2-D CVM grid, is the relative stability and depth of the free energy solution. One way to approach this is with a \textit{perturbation study}, so that we take the following steps: 

\begin{enumerate} \itemsep0pt 
\item \textbf{\textit{First free energy minimization}}, 
\item \textbf{\textit{System perturbation}}, and
\item \textbf{\textit{Second free energy minimization}}.
\end{enumerate} 

The following subsections discuss the results of this study. We pay particular attention to how the various configuration variable values change as a response to \textit{h}. 

Some of the following was originally presented in the V \& V document \cite{AJMaren-TR2018-001v2-V-and-V}. They are given again here (with both some contraction as well as expansion on certain points), to assist understanding of how the various configuration variable and thermodynamic values depend on the \textit{h-value}.

%
\subsection{Configuration variables and thermodynamic values: exemplar code run}
\label{subsec:results-code-run}
%

\begin{figure}[ht]
  \centering
  \fbox{
  \rule[-.5cm]{0cm}{4cm}\rule[-.5cm]{0cm}{0cm}	
  \includegraphics [trim=0.0cm 0cm 0.0cm 0cm, clip=true,   width=0.95\linewidth]{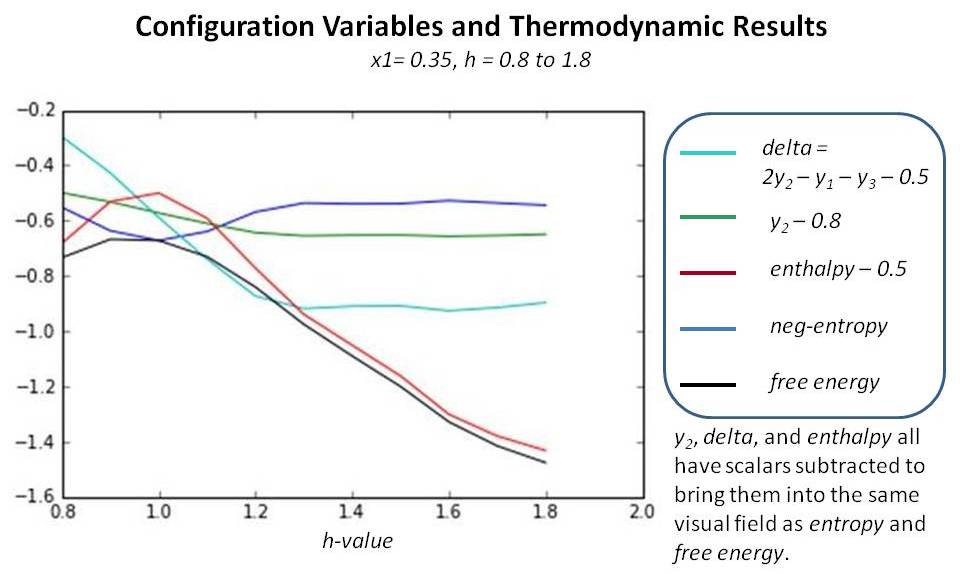}} 
  \vspace{3mm} 
  \caption{Configuration variable and thermodynamic values for the case where $x_1 = 0.35$ and $x_2 = 0.65$, and where the interaction enthalpy parameter $h$ ranges as $h = 0.8 .. 1.8$.}   
\label{fig:Perturbation-Results_x1=0pt35_lbld_crppd_2018-01-22}
\end{figure}

Figure~\ref{fig:Perturbation-Results_x1=0pt35_lbld_crppd_2018-01-22} presents an example result from the \textit{perturbation studies} experiments. These results were initially presented in the V\&V Report \cite{AJMaren-TR2018-001v2-V-and-V}.

This data is actually from a perturbation run, where the 2-D grid is established as described via random pattern generation (see the V\&V document for details), and then perturbed by a given amount (in this case, a fraction of 0.1 of the existing nodes are flipped), and then taken to free energy minimum a second time. 

What is interesting about this example, in comparison to previous analytic and computational results (see, e.g., Figure~\ref{fig:Perturbation-Results_x1=0pt50_lbld_crppd_2018-01-22}), is that it presents data for the case where $x_1 = 0.35$, and correspondingly, $x_2 = 0.65$. (By contrast, almost all previous discussions have dealt with the equiprobable case, where $x_1 = x_2 = 0.5$.)

When nodes are ``flipped'', and as the resultant system is taken back to a free energy minimum, the process ensures that the total number of nodes in each of the two states, \textbf{A} and \textbf{B}. In this particular example, this means that the value of $x_1 = 0.35$ is maintained throughout.  

\vspace{3mm} 

These results were obtained from the program 2D-CVM-perturb-expt-1-2b-2018-01-12.py, run on Friday, Jan. 12, 2018.

The parameter settings were for $x_1 = 0.35$ and $h = 0.8 .. 1.8$, with a total of twenty trials ($numTrials = 20$) for each \textit{h}-value.  The data table from this run is presented in the V\&V Report \cite{AJMaren-TR2018-001v2-V-and-V}. All reported results for configuration variable and thermodynamic values are averages over $numTrials$ runs, where $numTrials = 20$.

%
\subsection{Interpreting the $y_2$ results}
\label{subsubsec:interpreting-y2}
%

The values observed for $y_2$ conform to expectations. In Figure~\ref{fig:Perturbation-Results_x1=0pt35_lbld_crppd_2018-01-22}, \textbf{$y_2$} is shown in green, as $y_2 - 0.8$ (in order to bring the $y_2$ values within the same visual range as other results). (More details are given in the previously-mentioned V\&V document \cite{AJMaren-TR2018-001v2-V-and-V}.) 

When $h=1.0$, $y_2 = 0.2278$, which is the expected result. (The true expected result is $y_2 = 0.35 * 0.65 = 0.2275$; the observed value of $0.2278$ is an average over twenty trials. The deviance from the theoretical expectation is acceptable. )

When $h<1.0$, the $y_2$ values are greater, and when $h>1.0$, the $y_2$ values are smaller. In fact, $y_2$ ranges from $y_2 = 0.301$ (when $h = 0.8$) down to $y_2 = 0.151$ (when $h = 1.8$). These again are expected results. In brief, when $h<1.0$, then $\varepsilon_1 <0.0$, meaning that the interaction enthalpy parameter $\varepsilon_1$ is negative. When $\varepsilon_1$ is negative, the enthalpy is decreased by increasing $y_2$, as the interaction enthalpy $\varepsilon_1$ multiplies the term $(2 y_2 - y_1 - y_3)$. Thus, maximizing $y_2$ is expected when $h<1.0$.

There is a limit as to how far $y_2$ can be increased; presumably it can approach $0.5$, however, that would mean that the units were arranged in a strict checkerboard manner; that there were no instances of like-near-like at all. This is rather difficult to achieve; both in creation of highly-ordered systems, and in this particular code, which uses a simplistic \textit{find-and-flip} strategy. 

As previously noted, when $h>1.0$ ($\varepsilon_1 >0.0$), the $y_2$ values are smaller. This is because the system enthalpy now is decreased when $y_2$ is made smaller. Thus, the system moves more towards a like-with-like configuration (increasing $y_1$ and $y_3$); maximizing the size of the various ``islands,'' and decreasing the size of their borders (minimizing $y_2$). 

There is a practical limit as to how far $y_2$ can be decreased; there will always be a border area between the state \textbf{A} islands (or even a single, massive state \textbf{A} continent) and the surrounding sea of state \textbf{B} units. This means that $y_2$ will not get close to zero. The actual practical limit for $y_2$ will actually depend on the total system size (total number of nodes), because the border area will progressively decrease (although not disappear) as more and more islands join to become continents. Thus, the value of $y_2 < 0.157$, which occurs when $h \geq 1.2$, is not surprising. 

Once $y_2$ is pushed to a suitably small value, it becomes increasingly difficult for the simple \textit{find-and-flip} strategy to (randomly) find nodes where the flip will accomplish a free energy reduction. This is likely why there is general stability in the $y_2$ values beyond $h \geq 1.2$; there are simply not that many nodes where the flip will do much good, keeping in mind that \textit{two} nodes (each in a different state) have to be flipped in order to maintain the $x_1$ value.    

Thus, the preliminary conclusion is that free energy minimization is being accomplished, and that the $y_2$ values are behaving as expected.

%
\subsection{Interpreting the $delta$ results}
\label{subsubsec:interpreting-delta}
%

Again referencing Figure~\ref{fig:Perturbation-Results_x1=0pt35_lbld_crppd_2018-01-22}, we examine the curve for \textit{delta} (shown in cyan), defined as $(2 y_2 - y_1 - y_3)$, which is the actual term that is multiplied by $\varepsilon_1$ to achieve the interaction enthalpy term. (The contributing term for $y_2$ is shown in dark green in this figure.) This curve behaves as expected. 

In particular, we note that there is a nearly linear behavior in the region between $h = 0.8$ and $h = 1.3$. When $h = 0.8$, \textit{delta} = 0.2035 (according to the data table shown in V\&V document \cite{AJMaren-TR2018-001v2-V-and-V}). When $h = 1.2$, \textit{delta} = -0.3730. When $h = 1.0$, we would expect that there would be purely probabilistic distribution of units into their configurations, and thus expect that $y_1 = 0.35*0.35 = 0.1225$, $y_3 = 0.65*0.65 = 0.4225$, and $y_2 =.2275$ (as mentioned earlier). We would have then that $2 y_2 - y_1 - y_3 = 2*.2275 - 0.1225 - 0.4225 = -0.090$. The actual value is \textit{delta} = -0.0887, which is acceptably close. 

Similar arguments hold for the expected and observed values of \textit{delta} as did for $y_2$ in the preceding discussion. 

We again note that the values for \textit{delta} level out as \textit{h} increases beyond 1.2; this is because there are not that many units that the simple \textit{find-and-flip} strategy can easily find. In particular, we note that the $z_3$ value at $h \geq 1.3$ is typically around $z_3 = 0.04$, which is very small. 

In particular, we observe that this $z_3$ value indicates that we have pushed the system to its limit for minimizing $z_3$, which is the \textbf{A}-\textbf{A}-\textbf{B} configuration. This $z_3$ value indicates a border of a rather large island of state \textbf{A} units in a sea of \textbf{B} units. Specifically, for the 256-unit system that is the subject for this investigation, when $z_3 = 0.04$, then $N = 0.4*256/2 = 102.4/2 = 51$ triplets involve border units around islands / continents of state \textbf{A}. This is approximately \textit{1/5th} of the total number of units available. This suggests that we have pushed the system about as far as it can go. Of course, a visual inspection of the resulting grid would be enormously useful in confirming these assessments. This will be included in a subsequent document.  

%
\subsection{Interpreting the thermodynamic results}
\label{subsubsec:interpreting-thermodynamic}
%

The enthalpy curve is shown in red in Figure~\ref{fig:Perturbation-Results_x1=0pt35_lbld_crppd_2018-01-22}. (Actually, the curve shown is \textit{enthalpy-0.5}, so that it can be seen in close juxtaposition to the other data presented.) 

The enthalpy is maximum when $h=0$, which is to be expected. As we minimize free energy, we minimize enthalpy. As soon as we introduce some non-zero interaction enthalpy, we have an opportunity to adjust the configuration values (specifically the $y_i$, as just discussed) to lower the enthalpy. 

The entropy is similarly at a maximum (neg-entropy is at a minimum) when $h=0$. The negative entropy increases for non-zero values of \textit{h}, as expected. 

We particularly note that in the vicinity of $h=0$, or more generally, in the range of $0.8 \leq h \leq 1.3$, the variances in the entropy and enthalpy are approximately on the same scale; one does not appreciably dwarf the other. 

When we move beyond $h \geq 1.3$, we find that the enthalpy term strongly dominates the entropy, and thus dominates the free energy. It does this because we are increasing the value of the interaction enthalpy coefficient, $\varepsilon_1$, and not because we are gaining any appreciable difference in the configuration values. As noted in the previous discussions, these values have more-or-less stabilized in this range. 

Thus, \textbf{increasing $h$ beyond $h = 1.3$ does not serve any useful value}, suggesting a practical bound on \textit{h}-values for this kind of system. 

Our actual and practical choices for the \textit{h}-values should be based on the kind of behavior that we want to see in the configuration values. For modeling brain-like systems, we will most likely want $h \geq 0$, as that induces \textit{like-with-like} clustering, which seems to characterize certain neural collectives. 

In summary, the most useful result of the \textit{perturbation studies} has been to not only confirm that the various configuration variables and thermodynamic values behave as expected (which is why these results were included in the earlier V\&V study \cite{AJMaren-TR2018-001v2-V-and-V}), but also a \textbf{practical bound on the upper range for the \textit{h-value}}, i.e., $h < 1.3$.

%
\section{The phase space: boundary values}
\label{sec:phase space}
%

The previous sections introduced the configuration variables and the free energy for the 2-D CVM. What we would now like to obtain is a mapping of a set of configuration variable values to a corresponding \textit{h-value}. 

The phase space that we wish to describe has two boundaries: 

\begin{itemize}
\item Activation enthalpy is zero ($\varepsilon_0 = 0$), and 
\item Interaction enthalpy is zero ($\varepsilon_1 = 0$). 
\end{itemize}

We have already given substantial attention to the first case, where $\varepsilon_0 = 0$. This is the case where we have equiprobable units in each state, \textbf{A} and \textbf{B} ($x_1 = x_2 = 0.5$). We found the analytic solution giving the values for each of the configuration variables in this case, and also identified that the analytic solution has a divergence. Thus, we have identified a \textit{protocol} for obtaining the actual configuration variables for a given \textit{h-value}, when $\varepsilon_0 = 0$. 

What this gives us is a means of computationally obtaining the approximate configuration variables corresponding to a given \textit{h-value}; limited by the constraint that $\varepsilon_0 = 0$. We also have an approximate upper bound on a useful value for \textit{h-value}; this is not a hard limit, but it suggests that going much further for the \textit{h-values} will not give us substantially further value. 

We have not yet discussed the second case, in which the interaction enthalpy is zero, except for a brief example in Subsections \ref{subsubsec:interpreting-y2} and \ref{subsubsec:interpreting-delta}. As a brief overview, when the interaction enthalpy $\varepsilon_1 = 0$, then the distribution of the configuration variables can be found on a strictly probabilistic basis. 

This in itself is straightforward and expected. 

What is not obvious, from this correlation, is the relationship between $x_1$ and the activation enthalpy parameter $\varepsilon_0$. 

That is, we know that as $\varepsilon_0$ increases, the corresponding free energy-minimized value for $x_1$ should decrease. This is because the overall free energy is minimized when we reduce the number of active units (state \textbf{A} units) as we associate more and more energy (enthalpy) with those active units. 

Fortunately, we can obtain this correlation. 

This is addressed in a subsection immediately following the next, which reviews the impact of the interaction enthalpy parameter. First, we address the task of finding a reasonable \textit{target h-value}.

%
\subsection{Phase space boundary: activation enthalpy is zero}
\label{subsec:phase_space-activation-enthalpy-zero}
%

When the activation enthalpy $\varepsilon_0 = 0$, then we have the situation described in the previous sections. 

Given a value for $\varepsilon_1$, we can analytically obtain the various configuration variable values. However, as we've previously discussed, these analytically-obtained values are in error, especially as we increase $\varepsilon_1 > 0$ (or $h>1$). 
 
More often, we are likely to start with a 2-D CVM grid that is not at equilibrium, and the various configuration variables do not align neatly with a single corresponding \textit{h-value}. 

This means that we need a \textit{Protocol} for taking a system to a free energy equilibrium point. (Such a \textit{Protocol} was briefly identified in the preceding section.) Once we've taken a system to a free energy minimum, we can then identify the configuration variables associated with that free energy-minimized grid. 

However, before we can do this needed free energy minmization, we need some starting point - we need an estimated or \textit{target h-value}, which will guide the free energy minimization process that starts with the initial 2-D CVM grid configuration.

Once the free energy minimization is complete, the \textit{target h-value} has become the \textit{actual h-value}, and the configuration variable values have adjusted so that the free energy minimization is satisfied for that particular \textit{h-value}. That is, the configuration variable values that are obtained as a result of free energy minimization will be appoximately those that are associated (in a real sense) with the corresponding \textit{h-value}. These configuration variable values may differ from those of the analytic solution. 

Thus, given an initial grid configuration, our first step is to find a good \textit{target h-value}. 

We can identify our \textit{target h-value} by any of a variety of means. One very simple approach is to identify the approximate \textit{h-value} corresponding to different measured (counted) configuration variable values, e.g., $z_1$, $z_3$, and $y_2$. (Figure~\ref{fig:CVM-2D_scale-free_2017-12-12} illustrated select counted ($Z_1$ and $Z_3$) and fractional ($z_1$ and $z_3$) computational variable values for an initial grid.) We can then take either an average of these different corresponding \textit{h-values} or even select a visual approximation. The way in which we do this is not very important at this stage, as we're going to change the configuration variables substantially as we bring this initial starting grid to a free energy minimum. 

Later, when we have a good set of correspondences between (at-equilibrium) configuration variable values and their corresponding \textit{h-values}, we can use a table look-up to find a good \textit{target h-value}. However, we've not yet established this full set of correspondences. 

If we want a more mathematical approach, we can obtain an approximate \textit{h-value} by using an analytic method, described in Appendix B. 

Regardless of the method that we use, the \textit{h-value} that we've selected becomes a \textbf{\textit{target}} \textit{h-value}; that is, we will drive the system to a grid configuration that is at an equilibrium point for \textit{that specific h-value}. 

Once we've identified an initial 2-D CVM grid configuration and a \textit{target h-value}, we carry out a computational process of free energy minimization, which requires selecting and flipping various nodes (from \textbf{A} to \textbf{B}, and vice versa). Throughout this process, we keep the relative ratio of nodes in states \textbf{A} and \textbf{B} constant. 

For the equiprobable case where we have $x_1 = x_2 = 0.5$, we could have hoped that the analytic values for the configuration variables would be approximately correct, especially when $\varepsilon_1$ is close to 0 (\textit{h} is close to 1.0). However, as detailed results in Appendix~\ref{sec:Appendix-D-experimental-results-varying-x1} show, the at-equilibrium configuration variable values are substantially different from those predicted by the analytic solution, even for \textit{h-values} reasonably close to $h=1$. 

\begin{figure}[ht]
  \centering
  \fbox{
  \rule[-.5cm]{0cm}{4cm}\rule[-.5cm]{0cm}{0cm}	
  \includegraphics [trim=0.0cm 0cm 0.0cm 0cm, clip=true,   width=0.95\linewidth]{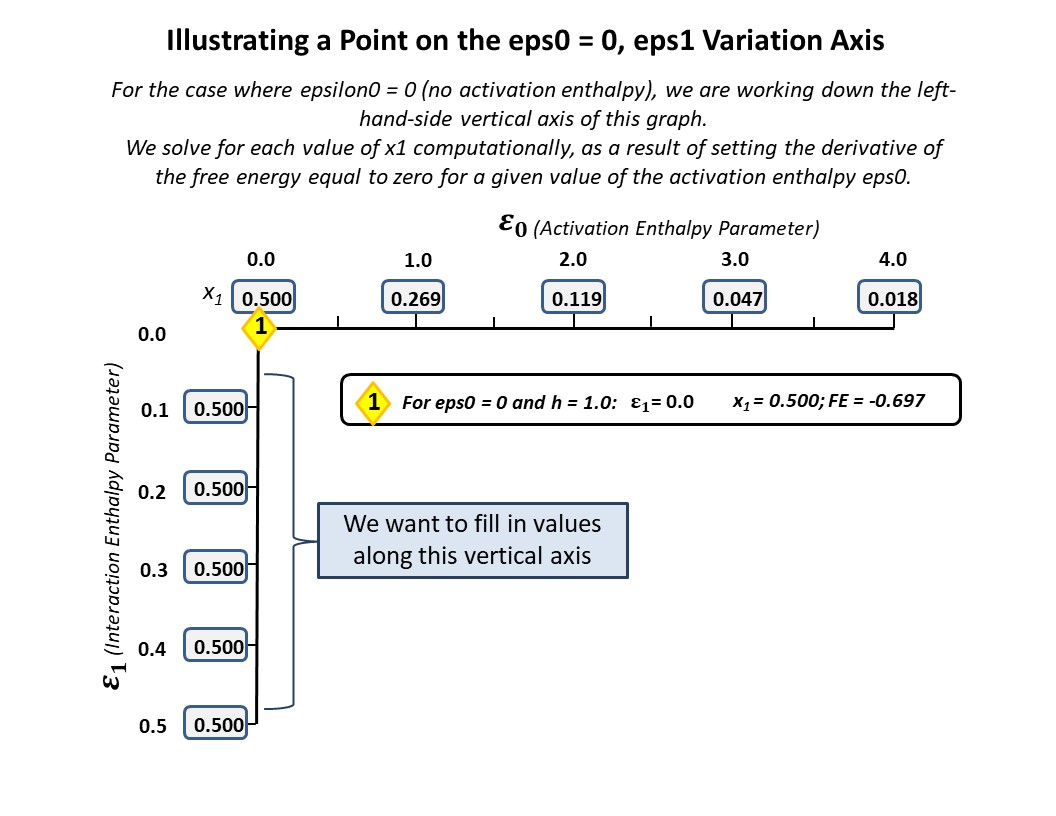}} 
  \vspace{3mm} 
  \caption{The first step is to identify the configuration variable and thermodynamic values along one axis; in this case, where we keep $\varepsilon_0 = 0$ and vary $\varepsilon_1$ (or vary \textit{h}).}   
\label{fig:2D-CVM_Expts-vary-eps1-axis_2019-08-29}
\end{figure}

Thus, our first goal is to build out a set of configuration variable values associated with corresponding \textit{h-values}, which is shown as the vertical (LHS) axis in Figure~\ref{fig:2D-CVM_Expts-vary-eps1-axis_2019-08-29}.

\vspace{3mm} 

The best way to figure out the configuration variable value sets associated with different \textit{h-values} (that is, figure out values along the LHS axis of Figure~\ref{fig:2D-CVM_Expts-vary-eps1-axis_2019-08-29}) is to start with some initial grid configuration, pick a \textit{target h-value}, and then conduct free energy minimization. Then, the resulting values for the different configuration values (as averaged over multiple runs) should be useful. 

Naturally, the trick is to pick a good \textit{target h-value} for a given initial grid. 

Appendix~\ref{sec:Appendix-B-computing-enthalpy-interaction-parameter} walks through this process in some detail, and further discusses the topographies associated with different \textit{h-values}.

%
\subsection{Phase space boundary: interaction enthalpy is zero}
\label{subsec:phase_space-interaction-enthalpy-zero}
%

When the interaction enthalpy $\varepsilon_1 = 0$, then we have the phase space boundary depicted by the upper axis of Figure~\ref{fig:2D-CVM_Expts-vary-eps1-axis_2019-08-29}.

Along this axis, the values of the configuration variables $y_i$, $w_i$, and $z_i$ are probabilistically determined by the value of $x_1$. This is because the values for the configuration variable values other than that for $x_1$ vary from their respective randomly-defined values \textit{only} because the interaction enthalpy brings like terms together, or pushes them apart (depending on the interaction enthalpy parameter value for $\varepsilon_1$, or correspondingly, when $h \neq 1.0$). 

When $h = 1.0$, then there is no reason for the distribution of unit states to differ from random. (Unless, of course, the activation enthalpy $\varepsilon_0 > 0$, in which case we'll have a smaller number of units in state \textbf{A} ($x_1 < 0.5$). Even then, the values for all the other configuration variables will be defined by the value for $x_1$.)

We can define a reasonable extent for the upper (horizontal) phase space boundary axis, where $x_1 \approx 0.05$, because when $x_1$ has that value, we have the following values for certain other configuration variables: 

\begin{itemize}
\setlength{\itemsep}{1pt}
\item \textbf{$x_1 = 0.5$} - total number of units in state \textbf{A} = 0.5*N, where $N$ is the total number of units in the grid. When $N=256$ (a 16x16 unit grid), $X_1 = 0.5*256 = 128$, which is readily achieved. 
\end{itemize}

Appendix~\ref{sec:Appendix-C-computing-activation-enthalpy-iparameter-given-x1} presents details for computing the configuration values along the upper horizontal ($\varepsilon_1 = 0$ axis), including the mechanism for identifying the $x_1$ value that correlates with specific values of  $\varepsilon_0$.

%
\subsection{Phase space: useful extent}
\label{subsec:phase_space-useful-extent}
%

\begin{figure}[ht]
  \centering
  \fbox{
  \rule[-.5cm]{0cm}{4cm}\rule[-.5cm]{0cm}{0cm}	
  \includegraphics [trim=0.0cm 0cm 0.0cm 0cm, clip=true,   width=0.95\linewidth]{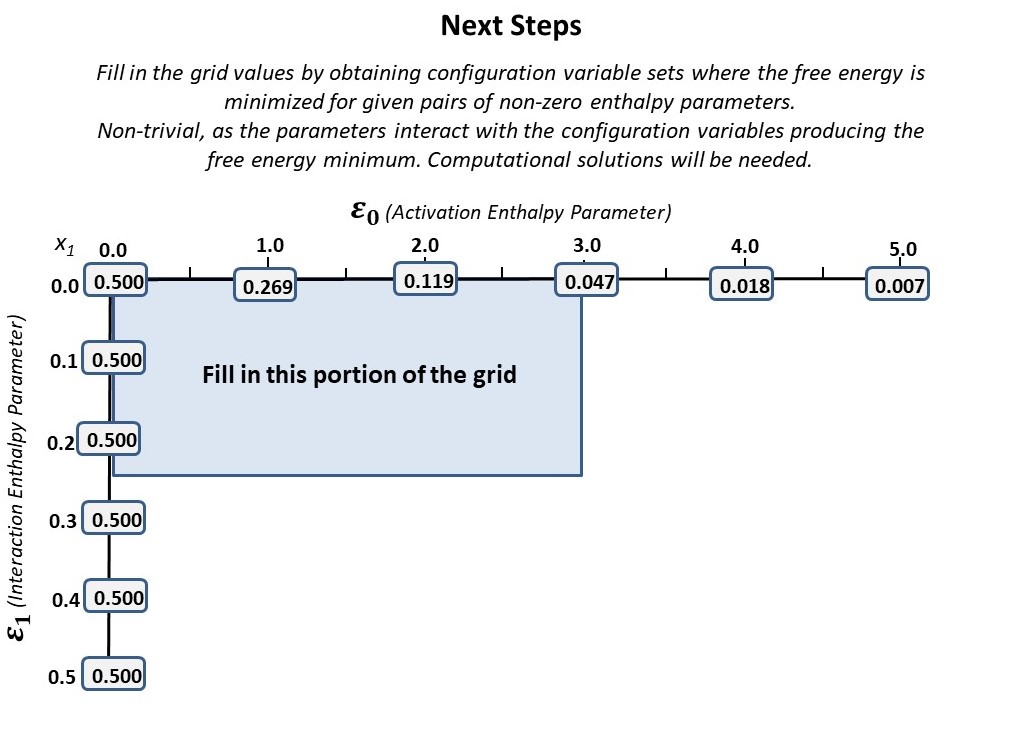}} 
  \vspace{3mm} 
  \caption{The useful extent for the phase space boundaries. The useful range for $\varepsilon_1$ (when $\varepsilon_0 = 0$, down the LHS axis) is $\varepsilon_1 = 0.0 \ .. \ 0.5$ (or $h = 1.0 \ .. \ 1.5$). Along the upper horizontal axis, we hold  $\varepsilon_1 = 0$ and vary $\varepsilon_0$ along the range  $\varepsilon_0 = 0.0 \ .. \ 3.0$.}   
\label{fig:2D-CVM_Expts-vary-eps0-and-eps1_next-steps_crppd_2019-08-29}
\end{figure}

Based on the results from the previous two sections, we can now identify the \textit{useful extent of the phase space boundaries} for the 2-D CVM. These are summarized in Figure~\ref{fig:2D-CVM_Expts-vary-eps0-and-eps1_next-steps_crppd_2019-08-29}. 

\vspace{3mm} 

%
\subsection{Phase space interior: need for incremental approach}
\label{subsec:phase_space-interior}
%

In the previous two sections, we identified an overall approach for obtaining both configuration variable and thermodynamic values across the two phase space boundary edges; i.e., where either $\varepsilon_0 = 0$ or $\varepsilon_1 = 0$. 

\begin{figure}[ht]
  \centering
  \fbox{
  \rule[-.5cm]{0cm}{4cm}\rule[-.5cm]{0cm}{0cm}	
  \includegraphics [trim=0.0cm 0cm 0.0cm 0cm, clip=true,   width=0.95\linewidth]{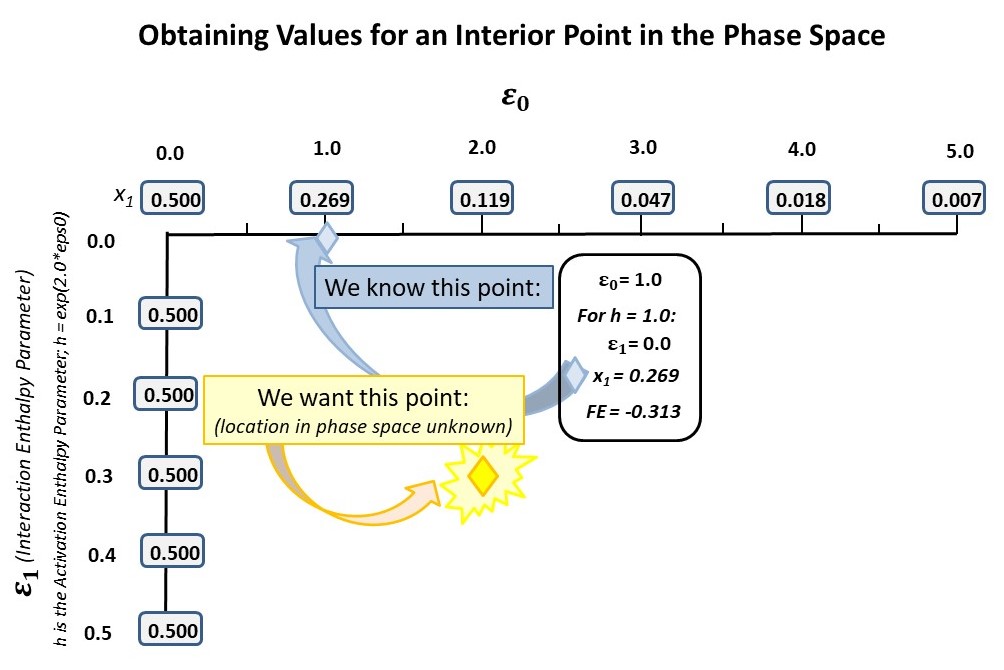}} 
  \vspace{3mm} 
  \caption  {The next step is to identify the configuration variable and thermodynamic values for the phase space interior; i.e., when both the interaction enthalpy and the activation enthalpy are greater than zero. This figure shows one known point along the upper horizontal axis, where $\varepsilon_1 = 0.0$, and $\varepsilon_0 = 1.0$. }
\label{fig:2D-CVM_Expts-vary-eps0-and-eps1_comput_phase-space-interior_crppd_2019-08-29}
\end{figure}

Now, we're interested in the interior of the phase space; in points such as that shown in Figure~\ref{fig:2D-CVM_Expts-vary-eps0-and-eps1_comput_phase-space-interior_crppd_2019-08-29}.

\vspace{3mm} 

The next step is to identify the configuration variable and thermodynamic values for the phase space interior; i.e., when both the interaction enthalpy and the activation enthalpy are greater than zero. This figure shows one known point along the upper horizontal axis, where $\varepsilon_1 = 0.0$, and $\varepsilon_0 = 1.0$. For this pair of enthalpy parameters, $x_1 = 0.269$. 

In general, along this upper horizontal asix, the non-zero values for $\varepsilon_0$ mean that we have $x_1 < 0.5$. For the specific point illustrated in Figure~\ref{fig:2D-CVM_Expts-vary-eps0-and-eps1_comput_phase-space-interior_crppd_2019-08-29}, we obtained the exact value for $x_1$ following a protocol identified in the previous section. 

\begin{figure}[ht]
  \centering
  \fbox{
  \rule[-.5cm]{0cm}{4cm}\rule[-.5cm]{0cm}{0cm}	
  \includegraphics [trim=0.0cm 0cm 0.0cm 0cm, clip=true,   width=0.95\linewidth]{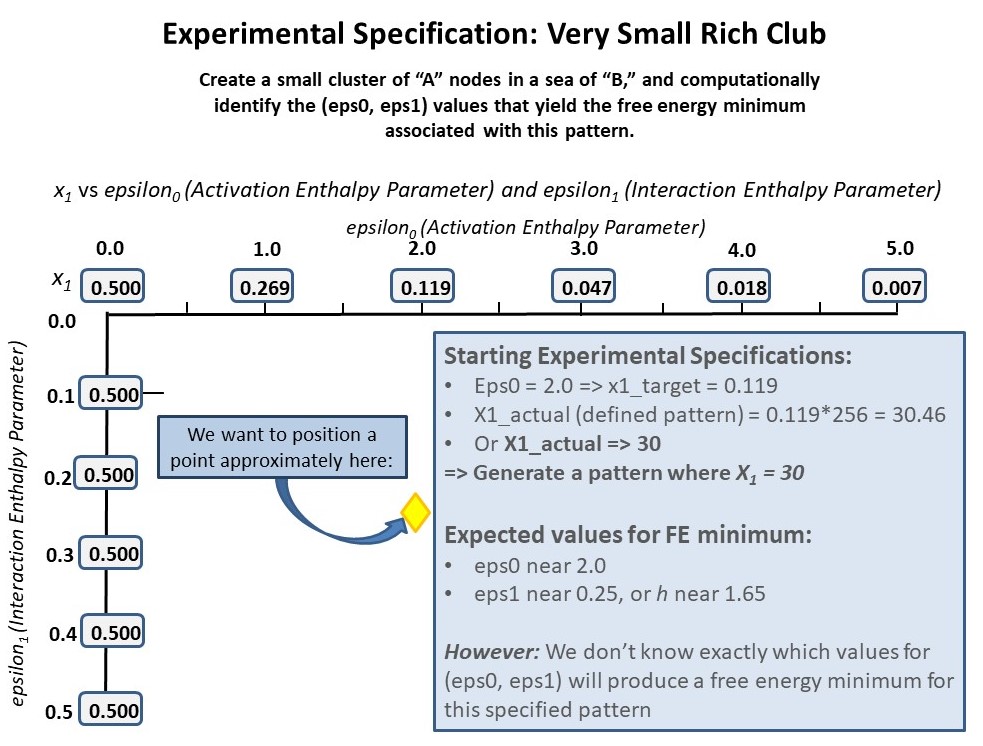}} 
  \vspace{3mm} 
  \caption  {We want to find the configuration variable values for points where the enthalpy parameters are in the phase space interior, so that both $\varepsilon_0 > 0.0$ and $\varepsilon_1 > 0.0$. }
\label{fig:2D-CVM_Expts-vary-eps0-and-eps1_very-small-rich-club_crppd_2019-08-29}
\end{figure}

It is not as straightforward to identify the configuration variable values when we have non-zero values for both $\varepsilon_0$ and $\varepsilon_1$. This is the kind of situation illustrated in Figure~\ref{fig:2D-CVM_Expts-vary-eps0-and-eps1_very-small-rich-club_crppd_2019-08-29}. 

\vspace{3mm} 

\begin{figure}[ht]
  \centering
  \fbox{
  \rule[-.5cm]{0cm}{4cm}\rule[-.5cm]{0cm}{0cm}	
  \includegraphics [trim=0.0cm 0cm 0.0cm 0cm, clip=true,   width=0.95\linewidth]{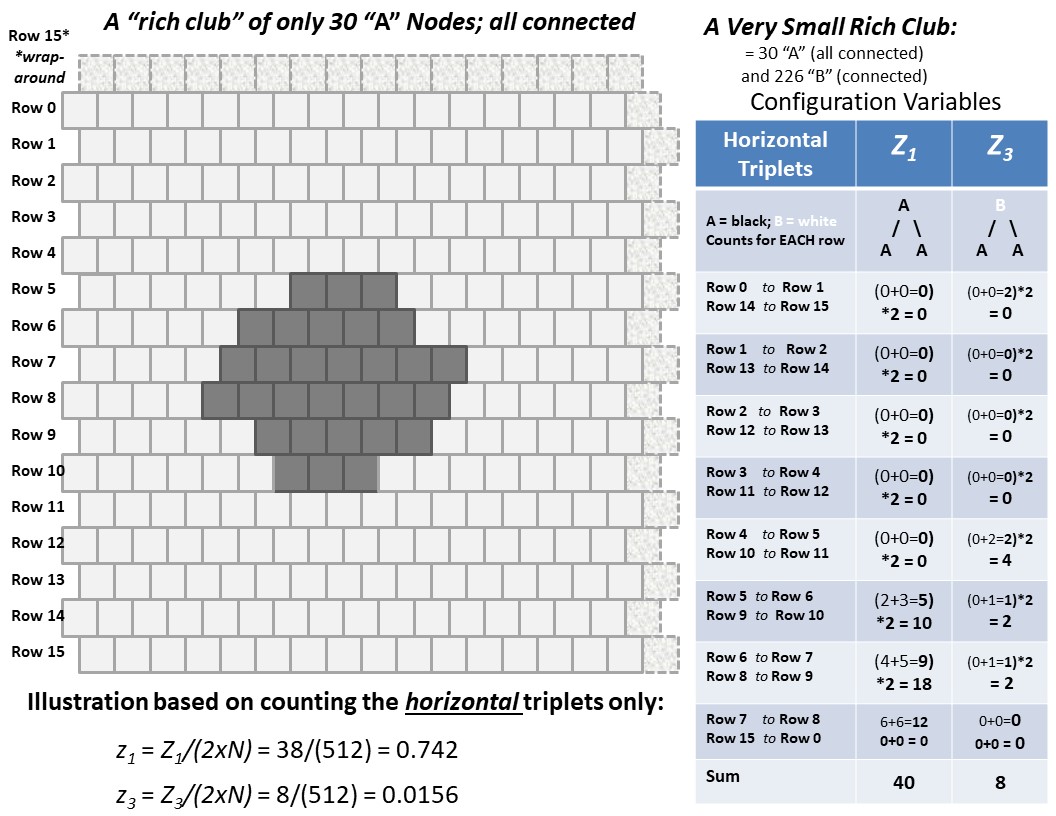}} 
  \vspace{3mm} 
  \caption  {An initial grid consisting of a very small rich club, or single small mass of 30 \textbf{A} units. We would anticipate that the enthalpy parameters would be somewhere in the neighborhood of $\varepsilon_0 = 2.0$, and $\varepsilon_1 > 1.5$. }
\label{fig:2D-CVM_Expts-vary-eps0-and-eps1_v-small-rich-club-grid_2019-08-29}
\end{figure}

This kind of system is illustrated in Figure~\ref{fig:2D-CVM_Expts-vary-eps0-and-eps1_v-small-rich-club-grid_2019-08-29}, which shows an initial grid structured as a very small rich club. 

\vspace{3mm} 

Previous experiments with ``rich-club-like'' topographies showed us that even with high values for $\varepsilon_1$, we still found that the initial grid design was disrupted as we brought the system to equilibrium. Thus, we anticipate that the system shown in Figure~\ref{fig:2D-CVM_Expts-vary-eps0-and-eps1_v-small-rich-club-grid_2019-08-29} is not at equilibrium. 

The challenge in bringing such a system to equilibrium is that we need to specify, in advance, the values for both $\varepsilon_0$ and $\varepsilon_1$. 

In this case, we don't know either of these enthalpy parameters. We know that both of them will be greater than zero. We suspect that the value for $\varepsilon_0$ will be in the neighborhood of $2.0$, because for this manually-designed system, $x_1 = 30/256 = 0.117$, and we know that if $\varepsilon_1 = 0.0$, then $x_1 = 0.119$ when $\varepsilon_0 = 2.0$. However, we don't know as yet how the configuration values change as we increase \textit{both} $\varepsilon_0$ \textit{and} $\varepsilon_1$.

To accomplish our goal, we would need to establish an initial grid, such as is shown in Figure~\ref{fig:2D-CVM_Expts-vary-eps0-and-eps1_v-small-rich-club-grid_2019-08-29}, and iteratively solve (varying not only the two enthalpy parameters but also the $x_1$ values) for a free energy minimized system. This will be the subject of a future study.

%
\section{Summary and conclusions}
\label{sec:summary}
%

There are several things that we can conclude from this preliminary investigation:

\begin{itemize}
\setlength{\itemsep}{1pt}
\item \textbf{Analytic vs. computational results} - while there is an analytic solution (when $x_1 = x_2 = 0.5$), this analytic solution has two divergence points. We would expect that the computational values for the configuration variables would differ as the \textit{h-value} approaches the divergence points. What is surprising is that the computational results differ from the analytically-predicted ones much earlier than would be expected.  
\item \textbf{Useful range of the $\varepsilon_0$ enthalpy parameter} - this study did not go so far as to computationally confirm the configuration variables given certain values of $\varepsilon_0$. Rather, we have identified how (when $\varepsilon_1 = 0$) $x_1$ and other configuration variables correlate with $\varepsilon_0$. We still need to take these values and confirm them computationally. However, this correlation between $\varepsilon_0$ and $x_1$ gives us a useful range for $\varepsilon_0$. While this range depends (somewhat) on the grid size, it is reasonable to keep $\varepsilon_0 < 3$, and possibly smaller. 
\item \textbf{Useful range of the $\varepsilon_1$ enthalpy parameter} - we can similarly identify that a useful range for $\varepsilon_1 <2$, and again, possibly smaller. A study reported in Subsection~\ref{subsubsec:interpreting-y2} suggests that even values up to $h = 1.3$ may be more than necessary, although larger \textit{h-values} (e.g., $h = 1.65$) were used in the course of this work and did indeed produce notably different results compared with smaller \textit{h-values} (e.g., $h = 1.16$).
\end{itemize}

In addition to these basic findings, we note interesting properties in the 2-D topographic grids that are generated during the free energy minimization process - topographic features such as ``spider-legs,'' ``channels,'' and ``rivers.'' The whole notion of free-energy-based topographies is new and worth substantial future study. 

%
\section{Future directions}
\label{sec:future-directions}
%

There are numerous steps to be taken: 

\begin{itemize}
\setlength{\itemsep}{1pt}
\item \textbf{Map the phase space} - we need the full set of configuration variable values associated with parameter sets $(\varepsilon_0, \varepsilon_1)$. 
\item \textbf{Smart strategies} - once we know the target configuration variables associated with a given parameter set, we can evolve smart strategies to move in the direction of a known soluiton. 
\item \textbf{Topography characterization} - as we understand the topographies induced by different parameter sets, we can begin correlating free energy-based topographies with various kinds of observable 2-D patterns, ranging from physical landscapes to brain activation patterns. 
\end{itemize}

The most immediate step, which will enable the tasks just identified, will be \textbf{object-oriented code development} - a necessary and obvious next-step. This will allow us to associate various properties with each unit (node) in the grid, and that will make possible the evolution of smart strategies for free energy minimization processes, instead of the simple find-and-replace method currently used.  

All of these are necessary preliminaries. Once completed, we can investigate use of a 2-D CVM grid and the associated free energy minimization process within a new form of computational engine.

%
\section{Code referenced in this document}
\label{sec:code-referenced}
%

\begin{itemize}
\setlength{\itemsep}{1pt}
\item \textbf{2D-CVM-perturb-expt-1-2b-2018-01-12.py} - free energy minimization including perturbation after reaching the first free energy minimum; this allows confirming that the free energy minimum is stable; allows user-specifiable (in **main**) values for $x_1$, $h$, $numTrials$, and many other parameters. 
\item \textbf{2D-CVM-def-pttrns-anlytc-FE-vary-e0-and-e1-2018-12-12.py, 2D-CVM-def-pttrns-anlytc-FE-vary-e0-and-e1-v1pt1-2018-12-29.py, and 2D-CVM-def-pttrns-anlytc-FE-vary-e0-and-e1-v1pt3-2018-12-29.py} (Note: the actual program files have underscores in place of some of the hyphens used in the filename just given; the filenames MAY be updated when the code is released) - free energy minimization given a starting user-defined 2-D grid, together with user-specified values for $x_1$ and $\varepsilon_1$ (the \textit{h-value}). Different programs perform slightly different tasks, with the same overall structure.
\end{itemize}

\textbf{Code Availability:} Codes referenced here may be made available in the future The code will be supported by extensive documentation, which will also be placed in the GitHub repository. Inquiries should be directed to: alianna@aliannajmaren.com. The verification and validation for the codes used to produce results presented here is given in Maren (2019) \cite{AJMaren-TR2018-001v2-V-and-V}. The best way to be informed of future developments is to ``opt-in'' on the author's website, at www.aliannajmaren.com.

\textbf{Copyright:} All codes referenced here have been independently develeoped by A.J. Maren. A.J. Maren holds the copyright to both the codes and this document itself. arXiv is granted a non-exclusive and irrevocable license to distribute this article.  

 \vspace{10 pt}


\begin{thebibliography}{1}


\bibitem{Kikuchi_1951_Theory-coop-phenomena}
R.~Kikuchi, ``A theory of cooperative phenomena,'' {\em Phys. Rev.}, vol.~988, no.~81, p.~127–138, 1951.  

\bibitem{Kikuchi-Brush_1967_Improv-CVM}    
R.~Kikuchi and S.G.~Brush, ``Improvement of the cluster variation method,'' {\em J. Chem. Phys.},
  vol.~47, p.~195, 1967.
 
\bibitem{Sanchez-et-al_1984_Gen-Cluster-Descrp-Multicomp-Systems}    
J. M.~Sanchez, F.~Ducastelle and D.~Gratias, ``Generalized cluster description of multicomponent systems,'' {\em Physica}, vol.~128A, pp.~334-350, November, 1984. 
\newblock doi:10.1016/0378-4371(84)90096-7.  

\bibitem{Maren-et-al_1984_Theoretical-model-hysteresis-solid-state-phase-trans}    
A.J.~Maren, S.H.~Lin, R.H.~Langley and L.~Eyring, ``A theoretical model for hysteresis in solid state phase transitions,'' {\em J. Solid State Chem.}, vol.~53, pp.~329-343, 1984. 
\newblock doi:10.1016/0022-4596(84)90110-5.   


\bibitem{Sanchez-and-Becker_1994_First-principles-CVM-phase-diagrams}
J.M.~Sanchez and J.D.~Becker, ``The role of the {C}luster {V}ariation {M}ethod in the first principles calculation of phase diagrams,''
{\em Progress of Theoretical Physics Supplement}, no.~115, pp.~131-145, 1994.
  
\bibitem{Pelizzola_1994_CVM-Pade-approx-crit-behavior}
A. Pelizzola, ``Cluster variation method, Pad\'{e} approximants and critical behaviour,''
{\em Phys Rev E Stat Phys Plasmas Fluids Relat Interdiscip Topics}, vol.~49, no.~4, pp.~R2503-R2506, Apr., 1994.

\bibitem{Finel_1994_CVM-applications}
A.~Finel, ``The Cluster Variation Method and Some Applications,'' in P.E.A.~Turchi P.E.A. and A.~Gonis (eds), {\em Statics and Dynamics of Alloy Phase Transformations, NATO ASI Series (Series B: Physics)}, vol.~319, Springer, Boston, MA., 1994. 

\bibitem{Cirillo_1996_phase-diagram-gonihedric}
E.N.M.~Cirillo, G.~Gonnella, D.A.~Johnston and A.~Pelizzola, ``The phase diagram of the gonihedric 3d {I}sing model via {CVM},'' {\em arXiv:hep-lat/9607078v1 30 Jul 1996}.

\bibitem{Kikuchi-Masuda-Jindoi_2002_CVM}
R.~Kikuchi and K.~Masuda-Jindo, ``Cluster variation method in the computational materials science,'' {\em Calphad},
vol.~26, no.~1, pp.~33-54, 2002.
\newblock doi:10.1016/S0364-5916(02)00023-8.  

\bibitem{Mohri_2013_CVM}
T.~Mohri, ``{C}luster variation method,''
{\em JOM: the Journal of the Minerals, Metals and Materials Society}, vol.~65, no.~11, pp.~1510-1522, 2013.
\newblock doi:10.1007/s11837-013-0738-5.  

\bibitem{Pelizzola_2005_CVM-stat-phys-prob-graph-models}
A. Pelizzola, ``{C}luster {v}ariation method in statistical physics and probabilistic graphical models,''
{\em J. Phys. A: Math. Gen.}, vol.~38, p.~R309, 2005.
\newblock doi:10.1088/0305-4470/38/33/R01.

\bibitem{Yedidia-Freeman-Weiss_2002_Understanding-belief-prop}
J.S.~Yedidia, W.T.~Freeman and Y.~Weiss, ``Understanding belief propagation and its generalizationss,''
{\em MERL TR-2001-22; Mitsubishi Electric Research Laboratories}, vol.~38, January, 2002.
\newblock www.merl.com.

\bibitem{Wainwright-and-Jordan_2008_Graph-models-exp-fam-var-inf}
M.J.~Wainwright and M.I.~Jordan, ``Graphical models, exponential families, and variational inference,''
{\em Foundations and Trends in Machine Learning}, vol.~1, no.~1-2, pp.~1-305, 2008.
\newblock doi:10.1561/2200000001.

\bibitem{Albers-et-al_2006_CVM-efficient-linkage-analysis}
C.A.~Albers and M.A. R.~Leisink and H. J.~Kappen, ``The cluster variation method for efficient linkage analysis on extended pedigrees,''
{in \em BMC Bioinformatics 2006, 7(Suppl 1):S1}, vol.~38, 20 March, 2006.
\newblock doi:10.1186/1471-2105-7-S1-S1.

\bibitem{Barton-Cocco_2013_Ising-models-neural-activity}
J.~Barton and S.~Cocco, ``Ising models for neural activity inferred via selective cluster expansion: structural and coding properties,''
{\em J. Stat. Mech.}, vol.~03, no.~P03002, pp.~1510-1522, 2013.
\newblock doi:10.1088/1742-5468/2013/03/P03002.


\bibitem{Pribram_1991_Brain-and-Perception}
K.~Pribram, {\em Brain and perception: holonomy and structure in figural processing}  (Hillsdale, N. J.: Lawrence Erlbaum Associates), 1991.

\bibitem{Fukushima_1980_Neocognitron}
K.~Fukushima, ``Neocognitron: a self-organizing neural network model for a mechanism of pattern recognition unaffected by shift in position,'' {\em Biol. Cybernetics}, vol.~36, pp.~193-202, 1980.

\bibitem{Grossberg-and-Mingolla_1985a-Neural-dynamics-perceptual-grouping}
S.~Grossberg and E.~Mingolla, ``Neural dynamics of perceptual grouping: Textures, boundaries, and emergent segmentations,'' {\em Perception and Psychophysics}, vol.~38, pp.~141-171, 1985.

\bibitem{Grossberg-and-Mingolla_1985b-Neural-dynamics-form-perception}
S.~Grossberg and E.~Mingolla, ``Neural dynamics of form perception: Boundary completion, illusory figures, and neon color spreading,'' {\em J. Psychological Review}, vol.~92, pp.~173-211, 1985.




\bibitem{Maren-Schwartz-Seyfried_1992_Config-entropy-stabilizes}
A.J.~ Maren,  E.~Schwartz, and J.~Seyfried, ``Configurational entropy
 stabilizes pattern formation in a hetero-associative neural network,''
 in {\em Proc. IEEE Int’l Conf. SMC} (Chicago, IL), pp.~89–93, October 1992.
\newblock doi:10.1109/ICSMC.1992.271796.

\bibitem{Maren_1993_Free-energy-as-driving-function}
A.J.~Maren, ``Free energy as driving function in neural networks,'' in {\em Symposium
 on Nonlinear Theory and Its Applications} (Hawaii), December 5-10 1993. 
\newblock doi:10.13140/2.1.1621.1529.

\bibitem{Schwartz-Maren_1994_Domains-interacting-neurons}
E.~Schwartz and A.J.~ Maren, ``Domains of interacting neurons: a statistical mechanical model,''
 in {\em Proc. World Congress on Neural Networks (WCNN)}, 1994.
\newblock doi:10.1109/ICSMC.1992.271796.

\bibitem{Maren_2016_CVM-primer-neurosci}
A.J.~Maren, ``The cluster variation method: a primer for neuroscientists,'' {\em
  Brain Sciences}, vol.~6, no.~4, p.~44, 2016.

\bibitem{AJMaren-TR2014-003}
A.~ Maren, ``The Cluster Variation Method II: 2-D grid of zigzag chains:
Basic theory, analytic solution and free energy variable distributions at
midpoint (x1 = x2 = 0.5),'' Tech. Rep. THM TR2014-003 (ajm), Themasis,
2014. 
\newblock doi:10.13140/2.1.4112.5446.


\bibitem{AJMaren-TR2018-001v2-V-and-V}
A.J.~Maren, ``Free Energy Minimization Using the 2-D Cluster Variation Method: Initial Code Verification and Validation,'' {\em Themasis Technical Report 2018-001v2 (ajm)}.
\newblock v1: 2018; v2: 2019.
\newblock arXiv:1801.08113v2 [cs.NE] 25 Jun 2019.


\bibitem{AJMaren_2019_Expt-Results_Two-epsilon-params}
A.J.~Maren, ``2-{D} cluster variation method: Comparison of Computational and Analytic Results,'' June, 2019.
\newblock {\em GitHub: github.com/ajmaren/2{D}-{C}luster-{V}ariation-{M}ethod; 
  MS (TM) PPT Slidedeck: 2D-CVM-Expts-vary-eps0-and-eps1-computational-2019-06-17v4.pptx}.



\bibitem{Friston_2010_Free-energy-principle-unified-brain-theory}
K.~Friston, ``The free-energy principle: a unified brain theory?,'' {\em Nat.
  Rev. Neurosci.}, vol.~11, pp.~127–138, 2010.
\newblock doi:10.1038/nrn2787.

\bibitem{Friston_2013_Life-as-we-know-it}
K.~Friston, ``Life as we know it,'' {\em Journal of The Royal Society
  Interface}, vol.~10, no.~86, 2013.
    
\bibitem{Friston-et-al_2015_Knowing-ones-place-free-energy-pattern-recognition}
K.~Friston, M.~Levin, B.~Sengupta, and G.~Pezzulo, ``Knowing one’s place: a
  free-energy approach to pattern regulation,'' {\em J. R. Soc. Interface},
  vol.~12, p.~20141383, 2015.
\newblock doi:10.1098/rsif.2014.1383; available online at:
  http://dx.doi.org/10.1098/rsif.2014.1383.
  
\bibitem{Beal_2003_Variational-algorithm-approx-Bayes-inference}
M.~J. Beal, {\em Variational algorithms for approximate Bayesian inference}.
\newblock PhD thesis, University College London, 2003.
\newblock PDF: http://www.cse.buffalo.edu/faculty/mbeal/papers/beal03.pdf.

\bibitem{AJMaren-TR2019-001v4-Deriv-Var-Bayes} 
A.J.~Maren, ``Derivation of the variational Bayes equations,''
Tech. Rep. THM TR2019-001v4 (ajm), Themasis, {\em arXiv:1906.08804v3 [cs.NE] 26 Jun 2019},
2019.









\end{thebibliography}

%
\appendix
\section{Appendix A: The 2-D CVM Equation}
\label{sec:Appendix-A-CVM-2-D-Eqn}
%

\renewcommand{\theequation}{A-\arabic{equation}}
\setcounter{equation}{0}  

This appendix recapitulates the free energy formalism for the 2-D CVM grid  originally made by Kikuchi in 1951 \cite{Kikuchi_1951_Theory-coop-phenomena}, and further refined by Kikuchi and Brush in 1967 \cite{Kikuchi-Brush_1967_Improv-CVM} (Eqn. I.16). Following their approach, we treat the 2-D CVM grid as being constructed from a series of zigzag chains. 

We begin with the free energy equation, previously introduced in Section \ref{subsec:2D-CVM-entropy} as Eqn.~\ref{eqn:Bar-F-2-D-basic-eqn}, and repeated here for convenience as

\begin{equation}
\label{Bar-G-2-D-basic-eqn-appendix}
  \begin{aligned}
\bar{G}_{2-D} = G_{2-D}/N = \\
  & \varepsilon(-z_1+z_3+z_4-z_6) - \bar{S}_{2-D}\\
+ & \mu (1-\sum\limits_{i=1}^6 \gamma_i  z_i )+4 
\lambda (z_3+z_5-z_2-z_4)
  \end{aligned}
\end{equation}

\noindent
where the entropy, previously expressed as Eqn.~\ref{eqn:Bar-S-2-D-basic-eqn}, is also repeated here as

\begin{equation}
\label{Bar-S-2-D-basic-eqn-appendix}
\bar{S}_{2-D} = 
  2 \sum\limits_{i=1}^3 \beta_i Lf(y_i)
          + \sum\limits_{i=1}^3 \beta_i Lf(w_i)
          - \sum\limits_{i=1}^2 Lf(x_i) 
          - 2 \sum\limits_{i=1}^6 \gamma_i Lf(z_i),           
\end{equation}

\noindent
and where $Lf(v)=vln(v)-v$. The Lagrange multipliers are $\mu$ and $\lambda$, and we have set $k_{\beta}T = 1$. 

As noted previously in Section \ref{subsec:2D-CVM-enthalpy}, we are working with two terms in the enthalpy expression; the  activation enthalpy and the interaction enthalpy. 

We base the interaction enthalpy term in this equation on the expression introduced first by Kikuchi and Brush \cite{Kikuchi-Brush_1967_Improv-CVM} (Eqn. I.16), who express the enthalpy for the 2-D CVM as

\begin{equation}
\label{enthalpy-Kikuch-Brush-appendix}  
\bar{H}_{2-D} = 2 \varepsilon_1(2 y_2 - y_1 - y_3) = 2 \varepsilon_1(-z_1 +z_3 + z_4 - z_6),
\end{equation}

\noindent
using the equivalence relations  

\begin{subequations} \label{sub:y-def-set-appendix}
\begin{gather} 
  y_1 = z_1 + z_2  \label{sub:y1-z1-z2-reln-appendixn} \\
  y_2 = z_2 + z_4 = z_3+z_5 \label{sub:y2-z2-z4-and-z3-z5-reln-appendix}\\
  y_3 = z_5 + z_6.  \label{sub:y3-z5-z6-reln-appendix}
\end{gather}
\end{subequations}

This expresses the notion that the interaction enthalpy is identified as twice the value of each nearest-neighbor interaction ($y_i$). (The multiplier in front of the $y_2$ term is due to the double degeneracy of $y_2$.) 

If the interaction enthalpy parameter ${\varepsilon_1}$ is positive, then we reduce the overall free energy by increasing the relative proportion of nearest-neighbor interactions between those that are like each other ($y_1$ and $y_3$, or \textbf{A} - \textbf{A} and \textbf{B} - \textbf{B}, respectively), and decreasing the relative proportion of interactions between unlike nodes ($y_2$, or \textbf{A} - \textbf{B} interactions). Conversely, if the interaction enthalpy ${\varepsilon_1}$ is negative, we minimize the free energy by increasing the proportion of unlike nearest neighbor pairs (increasing $y_2$). When  ${\varepsilon_1} = 0$, the configuration variables should all be at what would be expected from random distribution. Specifically, for the case where $x_1 = x_2 = 0.5$, we would expect that $2y_2 = 0.5$, and $y_1 = y_3 = 0.25$.

%
\subsection{Equivalence Relations Among the Configuration Variables}
\label{subsec:App-A-Equiv-Rel_Config-Vars}
%

For completeness, we present the entire set of equivalence relations among the configuration variables. This means that we express the sets of $x_i$, $y_i$, and $w_i$ in terms of the $z_i$ (Eqns. I.1 - I.4 \cite{Kikuchi-Brush_1967_Improv-CVM}):

For the $y_i$:

\begin{subequations} \label{sub:y-def-set-appendix}
\begin{gather} 
  y_1 = z_1 + z_2  \label{sub:y1-z1-z2-reln-appendixn} \\
  y_2 = z_2 + z_4 = z_3+z_5 \label{sub:y2-z2-z4-and-z3-z5-reln-appendix}\\
  y_3 = z_5 + z_6.  \label{sub:y3-z5-z6-reln-appendix}
\end{gather}
\end{subequations}

For the $w_i$:

\begin{subequations} \label{sub:w-def-set}
\begin{gather} 
  w_1 = z_1+z_3 \\
  w_2 = z_2+z_5 \\
  w_3 = z_4+z_6
\end{gather}
\end{subequations}

For the $x_i$:

\begin{subequations} \label{sub:x-y-z-relations-def-set}
\begin{gather} 
  x_1 = y_1+y_2=w_1+w_2=z_1+z_2+z_3+z_5 \label{sub:x1-y-z-relations-def}  \\
  x_2 = y_2+y_3=w_2+w_3=z_2+z_4+z_5+z_6  \label{sub:x2-y-z-relations-def} 
\end{gather}
\end{subequations}

The normalization is:
\begin{equation}
  1=x_1+x_2 =\displaystyle\sum\limits_{i=1}^6 \gamma_i z_i.
\end{equation}\\ 

\vspace{-10pt}

%
\subsection{Microstates for the 1-D CVM}
\label{subsec:App-A-Microstates-1-D-CVM}
%

\begin{figure}[ht]
  \centering
  \fbox{
  \rule[-.5cm]{0cm}{4cm}\rule[-.5cm]{0cm}{0cm}	
  \includegraphics [trim=0.0cm 0cm 0.0cm 0cm, clip=true,   width=0.8\linewidth]{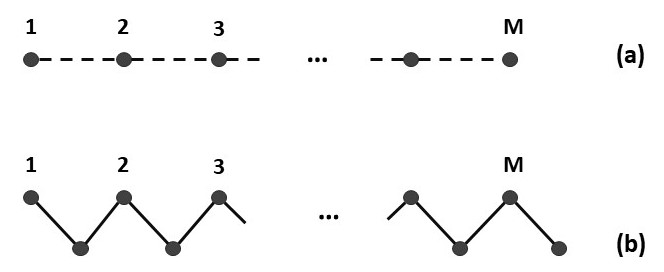}} 
  \vspace{3mm} 
  \caption{A horizontal single row (a) and a double row (b), from a square lattice CVM grid, turned to the diagonal; originally presented as Figure 2 in Kikuchi and Brush \cite{Kikuchi-Brush_1967_Improv-CVM}. }   
\label{fig:Single-zigzag-chain_Kikuchi-Brush-1967_crppd_2019-07-04}
\end{figure}

We begin with describing the 1-D CVM, constructed as a single zigzag chain, illustrated previously in Figure~\ref{fig:single_zigzag_chain_lbld}. We consider how we can construct a 1-D CVM system, composed as a single zigzag chain, originally presented as Figure 2 in Kikuchi and Brush \cite{Kikuchi-Brush_1967_Improv-CVM}, and shown here as Figure~\ref{fig:Single-zigzag-chain_Kikuchi-Brush-1967_crppd_2019-07-04}. Following the notation introduced by Kikuchi and Brush, we allow (in this section) $N_2$ to be the number of nodes per single row. 

\vspace{3mm} 

Viewing the zigzag chain as being composed of two horizontal rows, the number of ways of constructing a \textit{single row} of this chain are given as (Eqn. I.7 - 1.8 \cite{Kikuchi-Brush_1967_Improv-CVM})

\begin{equation} 
  \Omega_{single} = \frac{ {\{point\}_{M}} } { {\{diagonal pair\}_{M}} } ,
\label{omega-single-first-eqn}    
\end{equation}

\noindent
where 

\begin{equation} 
  {\{point\}_{M}} =\prod\limits_{i=1}^3 (M {x_i})!
\end{equation}

\noindent
and

\begin{equation} 
 {\{diagonal pair\}_{M}} =\prod\limits_{i=1}^3 (M {w_i})!^{\beta_i}.
 \label{diagonal-pair-eqn}
\end{equation}

%
\subsection{Microstates for the 2-D CVM}
\label{subsec:App-A-Microstates-2-D-CVM}
%

Referring now to Figure~\ref{fig:Single-zigzag-chain_Kikuchi-Brush-1967_crppd_2019-07-04}(b), we note that it is a zigzag form connecting two adjacent rows. This essentially constitutes a 1-D CVM system, and the number of ways of constructing this system can be derived similar to the previous equations, and is given as 

\begin{equation} 
  \Omega_{double} = \frac{ {\{pair\}_{2M}} } { {\{angle\}_{2M}} } ,
  \label{omega-double-first-eqn}  
\end{equation}

\noindent
where 

\begin{equation} 
  {\{pair\}_{2M}} =\prod\limits_{i=1}^3 (2M {y_i})!^{\beta_i}
\end{equation}

\noindent
and

\begin{equation} 
 {\{angle\}_{2M}} =\prod\limits_{i=1}^6 (2M {z_i})!^{\gamma_i}.
 \label{angle-eqn}
\end{equation}

As noted by Kikuchi and Brush (Eqn. I.11), when $M$ is large, we can use Stirling's approximation to write Eqn.~(\ref{omega-double-first-eqn}) as

\begin{equation} 
  \Omega_{double} = \left[\frac{ {\{pair\}_{M}} } { {\{angle\}_{M}} } \right]^2 .
  \label{omega-double-initial-eqn}
\end{equation}

\noindent
where $M$ is the number of lattice points in a row, and $\Omega_{double}$ refers to the juxtaposition of two rows. The degeneracy factors $\beta_i$ and $\gamma_i$ were described in Section ~\ref{subsec:Introducing-config-vars}, and illustrated in Figure~\ref{fig:Config-var-weights_v3_crppd_2017-05-17}. (\textit{Note:} This equation is identical with the original presentation given by Eqns. I.9 - I.10 from Brush and Kikuchi (1967) \cite{Kikuchi-Brush_1967_Improv-CVM}.)

When $M$ is large, Stirling{'}s approximation can be used to express the previous equation as (Eqn. I.11 from Brush and Kikuchi)

\begin{equation} 
  \Omega_{double} = \left[\frac{\prod\limits_{i=1}^3 (M_{y_i})!^{\beta_i} }
  {\prod\limits_{i=1}^6((M_{z_i})!^{\gamma_i}}\right]^2,
\label{omega-double-first-eqn}   
\end{equation}
\vspace{6pt}

\noindent
where Stirling{'}s approximation is given as: $N! = N ln(N) - N$.

We now wish to construct a 2-D system by adding a series of rows on top of the initial 1-D system, composed as the initial zigzag chain. 

\begin{figure}[ht]
  \centering
  \fbox{
  \rule[-.5cm]{0cm}{4cm}\rule[-.5cm]{0cm}{0cm}	
  \includegraphics [trim=0.0cm 0cm 0.0cm 0cm, clip=true,   width=0.8\linewidth]{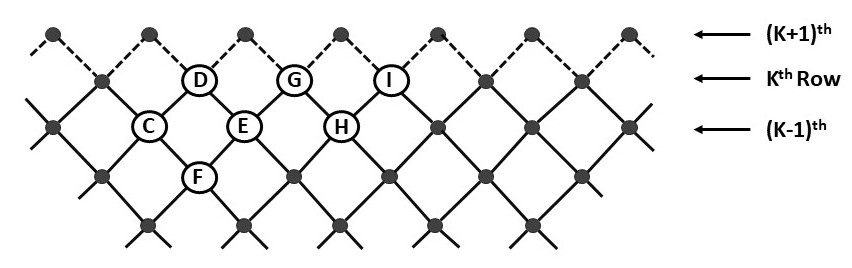}} 
  \vspace{3mm} 
  \caption{A 2-D CVM system, completed up to the $K^{th}$ row, originally presented as Figure 1 in Kikuchi and Brush \cite{Kikuchi-Brush_1967_Improv-CVM}. }   
\label{fig:Full-2D-CVM-grid_Kikuchi-Brush-1967_lettered_v2_crppd_2018-10-16}
\end{figure}

Suppose that the system has been completed up to the $k^{th}$ row, as shown in Figure~\ref{fig:Full-2D-CVM-grid_Kikuchi-Brush-1967_lettered_v2_crppd_2018-10-16}.

\vspace{3mm} 

The number of ways in which we can add the ${K+1}^{th}$ row, when the system has been completed to the $K^{th}$ row, is the ratio $\Omega_{double}/\Omega_{single}$. Thus, the number of ways of adding a total of $K$ rows and completing the entire system is (Eqn. I.12 from Brush and Kikuchi)

\begin{equation} 
  \Omega = \left[\frac{ \Omega_{double} } { \Omega_{single}} \right]^K.
\label{omega-first-eqn}  
\end{equation}
\vspace{6pt}

We can now substitute from Eqns.~\ref{omega-single-first-eqn} and \ref{omega-double-first-eqn} into Eqn.~(\ref{omega-first-eqn}), and use Stirling's approximation to obtain

\begin{equation} 
  \Omega = \frac{{\{pair\}_{N^2}} {\{diagonal pair\}_{N}}  } { {\{angle\}_{N^2}}  {\{point\}_{N}} } .
\label{omega-last-eqn}  
\end{equation}
\vspace{6pt} 

Kikuchi and Brush note that ``The source of approximation in this expression lies in identifying $\Omega_{double}$ and $\Omega_{single}$ in Eqn. I.12 [Eqn.~\ref{omega-first-eqn}] with those in (I.11) and (I.7) [Eqns.~\ref{omega-double-initial-eqn} and \ref{omega-single-first-eqn}, respectively]. In order to be rigorous, the $\Omega$'s in Eqn. I.12 [Eqn.~\ref{omega-first-eqn}] should be constructed using an infinitely long zigzag form, as in Figure 2(b) [Figure~\ref{fig:Single-zigzag-chain_Kikuchi-Brush-1967_crppd_2019-07-04}(b)], for the basic cluster.''   

We now wish to find an expression for the system entropy, which is defined as 

\begin{equation} 
  S = k_{\beta} \ln{\Omega} .
\label{basic-entropy-eqn}  
\end{equation}
\vspace{6pt}

We substitute from Eqn.~\ref{omega-last-eqn}, Eqn.~\ref{diagonal-pair-eqn}, and Eqn.~\ref{angle-eqn} into Eqn.~\ref{basic-entropy-eqn}, and once again use Stirling{'}s approximation to obtain the entropy for the system as

\begin{equation} 
  S_{2-D} = k_{\beta} ln \Omega = 
 k_{\beta} N \left[  2 \sum\limits_{i=1}^3 \beta_i Lf(y_i)
          + \sum\limits_{i=1}^3 \beta_i Lf(w_i)
          - \sum\limits_{i=1}^2 Lf(x_i) 
          - 2 \sum\limits_{i=1}^6 \gamma_i Lf(z_i)    \right].  
\end{equation}
\vspace{3pt}

\noindent
where $Lf(x)=xln(x)-x$, and $k_{\beta}$ is Boltzmann's constant.

When we divide through by $k_{\beta}N$, we recover the reduced system entropy as given in Eqn.~\ref{Bar-S-2-D-basic-eqn-appendix}. This is the entropy associated with a a full 2-D CVM grid.

%
\subsection{Analytic Solution for the 2-D CVM: Establishing the Equilibrium Equations }
\label{subsec:App-A-Analytic-Solution-2-D-CVM-equilibrium-equations}
%

Achieving the analytic solution involves solving a free energy equation for the equilibrium point, where the enthalpy-per-active-unit is given as zero, and the pairwise-interaction enthalpy is given as $\varepsilon = \varepsilon_1$. Because the enthalpy-per-active-unit is zero ($\varepsilon_0 = 0$), there is no {a priori} preference for a unit to be in either the \textbf{A} or \textbf{B} states; thus, $x_1 = x_2 = 0.5$.

Because states  \textbf{A} and \textbf{B} are equiprobable, we have a symmetric distribution with the other configuration variables; specifically: 

\begin{equation}
\begin{array}{lll}
  y_1 &=& y_3   \\
  z_1 &=& z_6 \\ 
  z_2 &=& z_5 \\
  z_3 &=& z_4  \\ 
 \end{array}
\end{equation}

As a first step, we take the derivative of the free energy equation Eqn.~\ref{Bar-G-2-D-basic-eqn-appendix} with respect to each of the configuration variables $z_i$. This gives us 

\begin{equation}
\begin{array}{lll}
  z_1 q &=& e^{\beta\varepsilon} y_1  (w_1/x_1)^{1/2}  \\
  z_2 q &=& e^{\lambda} (y_1 y_2 w_2)^{1/2}  (x_1 x_2)^{-1/4}\\ 
  z_3 q &=& e^{-2 \lambda} e^{-\beta\varepsilon} y_2  (w_1/x_1)^{1/2} \\
  z_4 q &=& e^{2 \lambda} e^{-\beta\varepsilon} y_2  (w_3/x_2)^{1/2} \\ 
  z_5 q &=& e^{-\lambda}(y_2 y_3 w_2)^{1/2} (x_1 x_2)^{-1/4} \\
  z_6 q &=& e^{\beta\varepsilon} y_3  (w_3/x_2)^{1/2}\\ 
 \end{array}
 \label{deriv-G-resp-z-six-eqns}
\end{equation}

\noindent
where $q = e^{- \mu \beta / 2}$, according to Kikuchi and Brush (Eqn. I.19), and (since we are working with reduced equations) we set $\beta = 1$, so that we have $q = e^{- \mu / 2}$. 

Note that Eqn.~\ref{deriv-G-resp-z-six-eqns} is identical with Eqn. (I.18) in Kikuchi and Brush; it is different from the corresponding set of equations (A.22) in Maren \cite{AJMaren-TR2014-003}. This is because the interaction enthalpy term used by Kikuch and Brush, and in this work, is $2\varepsilon_1(2y_2 - y_1- y_3)$, and in Maren \cite{AJMaren-TR2014-003}, the term used was the simpler $\varepsilon_1(2y_2)$. The additional factor of $2$ in this work and in the original Kikuch and Brush development is used to credit each interaction between units separately; from unit \textit{X} to \textit{Y} and then again from \textit{Y} back to \textit{X}. That factor of $2$ was absorbed in the previous work by Maren. 

Our goal is to find the equilibrium values for each of these configuration variables in terms of \textit{only} the enthalpy parameters. This is a fairly complex and convoluted derivation. 

Our strategy will be to obtain the equilibrium expression for a single configuration variable, $z_3$. Once we have that, we can use it (via substitutions and equivalence relations) to obtain the remaining configuration variables very easily. 

%
\subsection{The Configuration Variable $z_3$ at Equilibrium: Precursor Steps }
\label{subsec:App-A-z-3-at-equilibrium}
%

As an illustration, we compute the term for $z_3$. 

To do this, we seek the dependence of the free energy on $z_3$, given as

\begin{multline}
0 = \frac {\partial \bar{G}_{2-D}} {\partial z_3}  =   \frac {\partial } {\partial z_3}  \Big [ \varepsilon_0 x_1 +  2 \varepsilon_1(-z_1 + z_3 + z_4 - z_6)
- \bar{S}_{2-D} \\
+ \mu (1-\sum\limits_{i=1}^6 \gamma_i  z_i )+4 \lambda (z_3+z_5-z_2-z_4) \Big]. \\
\end{multline}

We use the equivalence relation in Eqn.~{\ref{sub:x1-y-z-relations-def}} to identify an expression for $x_1$ in terms of $z_3$, and so can write

\begin{equation}
 \frac {\partial ( \varepsilon_0 x_1)} {\partial z_3} =  \varepsilon_0  \frac {\partial ( z_1+z_2+z_3+z_5) } {\partial z_3} = \varepsilon_0.
\end{equation}

Carrying through with the terms that are linear in $z_3$, we can write

\begin{equation}
0 = \frac {\partial \bar{G}_{2-D}} {\partial z_3}  =   \varepsilon_0 +  2 \varepsilon_1  - \mu +4 \lambda  - \frac {\partial \bar{S}_{2-D}} {\partial z_3}
\label{eqn:deriv-G-w-z3}
\end{equation}

To find the dependence of the entropy on a specific configuration variable, we first obtain the derivative of the $Lf(x)$ term. 

\begin{equation}
 \frac {\partial Lf(x)} {\partial x}  =  \frac {\partial \big[ x\  ln(x) - x   \big]} {\partial x} = ln(x).
\end{equation}

In order to find the specific dependence of the entropy term on $z_3$, we first need to find the dependence of each of the configuration variables on $z_3$. 

%
\vspace{8 mm}
\textbf{First Step: Find the dependence of the entropy on $z_3$}
\vspace{3 mm}
%

To complete this, we need the dependence of each of the configuration variables $x_i$, $y_i$, and $w_i$ on $z_3$. Referring again to our set of equivalence relations in Eqns.~\ref{sub:y-def-set-appendix}, \ref{sub:w-def-set}, and \ref{sub:x-y-z-relations-def-set}, we have

\begin{equation}
\begin{array}{lll}
 \displaystyle
  \qquad \sfrac {\partial x_1} {\partial z_3}  = \displaystyle \qquad  \sfrac {\partial z_1+z_2+z_3+z_5)} {\partial z_3} = 1   \\
  \qquad \sfrac {\partial x_2} {\partial z_3}  = \qquad \sfrac {\partial (z_2+z_4+z_5+z_6)} {\partial z_3} = 0   \\ 
  \qquad \sfrac {\partial y_1} {\partial z_3}  = \qquad  \sfrac {\partial (z_1 + z_2)} {\partial z_3} = 0   \\
  \qquad \sfrac {\partial y_2} {\partial z_3}  = \qquad \sfrac {\partial (z_3 + z_5 + z_2 + z_4)} {2\partial z_3} = 1/2   \\  
  \qquad \sfrac  {\partial y_3} {\partial z_3}  = \qquad \sfrac  {\partial (z_5 + z_6)} {\partial z_3} = 0   \\
  \qquad \sfrac  {\partial w_1} {\partial z_3}  = \qquad \sfrac  {\partial (z_1 + z_3)} {\partial z_3} = 1  \\
  \qquad \sfrac  {\partial w_2} {\partial z_3}  =  \qquad \sfrac  {\partial (z_2 + z_5)} {\partial z_3} = 0   \\
  \qquad \sfrac {\partial w_3} {\partial z_3}  =   \qquad \sfrac {\partial (z_4 + z_6)} {\partial z_3} = 0   \\
 \end{array}
\label{eqn:Dep-of-config-vars-on-z3-app-A}
\end{equation}


Consulting Eqn.~\ref{eqn:Dep-of-config-vars-on-z3-app-A}, we see that the only configuration variables that have a dependence on $z_3$ are $x_1$, $y_2$, and $w_1$. Thus, 

\begin{equation}
\frac {\partial\bar{S}_{2-D}} {\partial z_3}  =  
   \frac {\partial } {\partial z_3}
   \left[  2 \sum\limits_{i=1}^3 \beta_i Lf(y_i)
          + \sum\limits_{i=1}^3 \beta_i Lf(w_i)
          - \sum\limits_{i=1}^2 Lf(x_i) 
          - 2 \sum\limits_{i=1}^6 \gamma_i Lf(z_i)    \right] , 
\end{equation}

\noindent
or

\begin{equation}
\frac {\partial\bar{S}_{2-D}} {\partial z_3}  = 2*2 ln(y_2)(1/2) + ln(w_1) - ln(x_1) - 2 ln(z_3). 
\label{eqn:deriv-S-w-z3}
\end{equation}

%
\vspace{8 mm}
\textbf{Second Step: Express $z_3$ in terms of other configuration variables}
\vspace{3 mm}
%

We insert Eqn.~\ref{eqn:deriv-S-w-z3} into  Eqn.~\ref{eqn:deriv-G-w-z3} to obtain

\begin{equation}
0 = \frac {\partial \bar{G}_{2-D}} {\partial z_3}  =   \varepsilon_0 +  2 \varepsilon_1  - \mu +4 \lambda  - [2 ln(y_2) + ln(w_1) - ln(x_1) - 2 ln(z_3)].
\end{equation}

We set $\varepsilon_0 = 0$ for the equiprobable case, and rearrange to obtain

\begin{equation}
2 ln(z_3) - \mu=  -2 \varepsilon_1  - 4 \lambda + 2 ln(y_2) + ln(w_1) - ln(x_1).
\end{equation}

We divide through by 2 to obtain

\begin{equation}
ln(z_3)  - \mu/2 =   -\varepsilon_1  - 2 \lambda +  ln(y_2) + ( ln(w_1) - ln(x_1))/2.
\label{eqn:deriv-G-w-z3-updated}
\end{equation}

We take the exponent of both sides to obtain

\begin{equation}
{z_3} e^{ -\mu/2}= e^{-\varepsilon_1}   e^{-2\lambda}{y_2} (w_1/x_1)^{1/2}.
\end{equation}

We recall that we defined $q = e^{- \mu / 2}$ and make the substitution to obtain

\begin{equation}
{z_3} q = e^{-\varepsilon_1} e^{-2\lambda} {y_2}(w_1/x_1)^{1/2}.
\end{equation}

Our goal has been to obtain the results previously presented in Eqn.~\ref{deriv-G-resp-z-six-eqns} as

\begin{equation}
z_3 q  =  e^{-2 \lambda} e^{-\beta\varepsilon} y_2  (w_1/x_1)^{1/2}.
\end{equation}

These results are the same as those of Kikuchi and Brush, keeping in mind that we have let $\beta = 1$ and $\varepsilon_1 = \varepsilon$.

The remaining expressions in Eqn.~\ref{deriv-G-resp-z-six-eqns} can be similarly obtained.

%
\subsection{Obtaining the Equilibrium Solution for the 2-D CVM }
\label{subsec:App-A-Equilibrium-Solution-2-D-CVM}
%

In the previous subsection, we obtained expressions for the $z_i$ in terms of the remaining configuration variables; the $x_i$, $y_i$, and $z_i$. Now, we use this to find an analytic solution that gives us the equilibrium values for specific configuration variables in terms of the interaction enthalpy. 

We will find it convenient to use a transformation introduced by Kikuchi and Brush (Eqn. I.23), where $h = \exp(2\beta\varepsilon_1)$. As with Kikuchi and Brush, we let $\beta = 1$. Since we are solving for the equilibrium case, we set the activation enthalpy $\varepsilon_0 = 0$. We thus simplify the notation be setting  $\varepsilon_1 = \varepsilon$, so we can write $h = \exp(2\varepsilon)$.

This is a particularly long and convoluted derivation. Thus, it is labeled in terms of different steps. 

%
\vspace{8 mm}
\textbf{Step 1: Express $z_1/z_3$ as the ratio $s$}
\vspace{3 mm}
%
 
We approach the solution given by Kikuchi and Brush (Eqn. I.25) by first letting $s=z_1/z_3$. Then

\begin{equation}
s=z_1/z_3 = \frac {e^{\varepsilon} y_1  (w_1/x_1)^{1/2}} {e^{-2 \lambda} e^{-\varepsilon} y_2  (w_1/x_1)^{1/2}}
\label{s-initial-definition-z1-and-z3}
\end{equation}

\noindent
or

\begin{equation}
s=e^{2\varepsilon} e^{2 \lambda}  \frac { y_1  } { y_2 } = h  e^{2 \lambda}  \frac { y_1  } { y_2 },
\label{s-initial-definition}
\end{equation}

\noindent
where $h = exp^{2 \varepsilon}$.

\vspace{8 mm}
\textbf{The Kikuch-Brush notation for the $h-value$}
\vspace{3 mm}

Kikuchi and Brush \cite{Kikuchi-Brush_1967_Improv-CVM} restrict their attention to the case where the activation enthalpy parameter ${\varepsilon_0} = 0$, so that there is an equiprobable distribution of units into states \textbf{A} and \textbf{B} ($x_1 = x_2 = 0.5$). In this particular case,  it is possible to find an analytic solution for ${\varepsilon_1}$  in terms of the configuration variables, and vice versa. It becomes more convenient to express the interaction enthalpy ${\varepsilon_1}$ in terms of an interaction enthalpy parameter \textit{h}. 

Kikuchi and Brush used the parameter replacement $h=e^{{2 \beta \varepsilon_1}}$ (Eqn. I-23 \cite{Kikuchi-Brush_1967_Improv-CVM}; Kikuchi and Brush use $H$ instead of $h$). As we are working with the reduced equation formalism, we let ${\beta} = 1$, so that $h=e^{{2 \varepsilon_1}}$. Further, since the analytic solution is specific to the case where $\varepsilon_0 = 0$, we set $\varepsilon_1 = \varepsilon$, so that $h=e^{{2 \varepsilon}}$. Throughout this paper, and others in this series, we refer to $h$ as the \textit{h-value}, which is the key parameter governing the topography at a free energy minimum.

%
\vspace{8 mm}
\textbf{Step 2: Multiply $z_1$ and $z_3$}
\vspace{3 mm}
%

We next perform the multiplication of $z_1q$ and $z_3q$ from Eqn.~\ref{deriv-G-resp-z-six-eqns} to obtain

\begin{equation}
  z_1 z_3 q^2 =  e^{\varepsilon} y_1  (w_1/x_1)^{1/2} *e^{-2 \lambda} e^{-\varepsilon} y_2  (w_1/x_1)^{1/2}
 \label{z1q-times-z3q-step1}
\end{equation}

\noindent
or

\begin{equation}
  z_1 z_3 q^2  = e^{-2 \lambda} y_1  y_2 (w_1/x_1) .
 \label{z1q-times-z3q-step2}
\end{equation}

%
\vspace{8 mm}
\textbf{Step 3: Square $z_2$ }
\vspace{3 mm}
%

Next, we square the expression that we have for $z_2q$, so that from

\begin{equation}
  z_2 q = e^{\lambda} (y_1 y_2 w_2)^{1/2}  (x_1 x_2)^{-1/4} 
\end{equation}

\noindent
we obtain

\begin{equation}
  {z_2}^2 q^2 = e^{2\lambda} (y_1 y_2 w_2)  (x_1 x_2)^{-1/2}.
 \label{z2q-squared-step1}  
\end{equation}

%
\vspace{8 mm}
\textbf{Step 4: Reduce number of configuration variables by isolating $y_1 y_2$ }
\vspace{3 mm}
%

Our next step is to reduce the number of configuration variables used in the expressions for the $z_1$, $z_2$, and $z_3$ variables. We do this by noticing that the factor of $y_1y_2$ occurs in both Eqns.~\ref{z1q-times-z3q-step2} and \ref{z2q-squared-step1}. We isolate this term from the previous Eqn.~\ref{z2q-squared-step1}. 

\begin{equation}
 y_1 y_2 = \frac { {z_2}^2 q^2e^{-2\lambda} }{w_2} (x_1 x_2)^{1/2} .
 \label{z2q-squared-step3}  
\end{equation} 

We substitute the expression for $y_1y_2$ into the equation for $z_1z_3$ to obtain

\begin{equation}
  z_1 z_3 q^2  = e^{-2 \lambda} y_1  y_2 (w_1/x_1) =  e^{-2 \lambda}  \left[\frac { {z_2}^2 q^2e^{-2\lambda} }{w_2} (x_1 x_2)^{1/2}\right] (w_1/x_1) .
 \label{z1q-times-z3q-step3}
\end{equation}

We collect terms and divide through by $q^2$ on both sides to obtain

\begin{equation}
  z_1 z_3   =  e^{-4 \lambda}  {z_2}^2\frac {w_1 }{w_2} (x_2/ x_1)^{1/2} .
 \label{z1q-times-z3q-step4}
\end{equation}

We recall that at equilibrium for $\varepsilon_0 = 0$, and we have $x_1 = x_2$  to obtain

\begin{equation}
  z_1 z_3   =  e^{-4 \lambda}  {z_2}^2\frac {w_1 }{w_2} .
 \label{z1q-times-z3q-step5}
\end{equation}

Following the line of thought presented by Kikuchi and Brush (Eqn. I.22), we can set $\lambda=0$, and so write 

\begin{equation}
  z_1 z_3   =  {z_2}^2\frac {w_1 }{w_2} .
 \label{z1q-times-z3q-step6}
\end{equation}

We have moved from a representation for the configuration variables that requires all of the $x_i$, $y_i$, and $w_i$ to express the $z_i$ to a representation for the $z_i$ that involves only the $z_i$ and $w_i$; that is, we have removed the  $x_i$ and the  $y_i$ from the representation for the  $z_i$.

%
\vspace{8 mm}
\textbf{Step 5: Represent $z_2$ in terms of $z_1$ and $z_3$ }
\vspace{3 mm}
%

As a next step, we will write $z_2$ in terms of $z_1$ and $z_3$. 

We recall, from the \textit{equivalence relations} presented at the beginning of this appendix, that

\begin{equation}
 x_1 = y_1 + y_2 = z_1 + z_2 + z_3 + z_5.
 \label{equiv-relns-for-z2-expression-step1}
\end{equation}

Also, since at $x_1 = 0.5$, we have $z_2 = z_5$, we rewrite the previous equation as

\begin{equation}
 x_1 = 0.5 = z_1 + 2z_2 + z_3 .
\label{equiv-relns-for-z2-expression-step2}
\end{equation}

We can express $z_2$ in terms of $z_1$ and $z_3$, by writing

\begin{equation}
2z_2 = 0.5 - z_1 - z_3 .
 \label{equiv-relns-for-z2-expression-step3}
\end{equation}

We divide through by 2, and square both sides to obtain

\begin{equation}
z_2^2 = [0.5 - z_1 - z_3]^2/4 = [1 - 2z_1 - 2z_3]^2/16.
 \label{equiv-relns-for-z2-expression-step4}
\end{equation}

We substitute this into Eqn.~\ref{z1q-times-z3q-step6} to obtain

\begin{equation}
  z_1 z_3   =   [1 - 2z_1 - 2z_3]^2\frac {w_1 }{16 w_2} .
 \label{z1q-times-z3q-step7}
\end{equation}

%
\vspace{8 mm}
\textbf{Step 6: Divide through by $z_3^2$; multiply both sides by $\frac {w_2 }{w_1}$  }
\vspace{3 mm}
%

We divide through by $z_3^2$ and multiply both sides by $\frac {w_2 }{w_1}$ to obtain

\begin{equation}
\frac { z_1 }{ z_3}  \frac {w_2 }{w_1}   =   [1 - 2z_1 - 2z_3]^2/(16 z_3^2).
 \label{z1q-times-z3q-step8}
\end{equation}

Carrying through the division by $z_3^2$ on the RHS and rearranging terms, we get

\begin{equation}
\frac { z_1 }{ z_3}  \frac {w_2 }{w_1}   =   [1/z_3 - 2z_1/z_3 - 2]^2/(16).
 \label{z1q-times-z3q-step9}
\end{equation}

We previously defined $s = z_1/z_3$. We now substitute this into Eqn.~\ref{z1q-times-z3q-step9} to obtain

\begin{equation}
s  \frac {w_2 }{w_1}   =   [1/z_3 - 2s - 2]^2/(16).
 \label{z1q-times-z3q-step10}
\end{equation}

%
\vspace{8 mm}
\textbf{Step 7: Express $w_1$ and $w_2$ in terms of $z_3$  }
\vspace{3 mm}
%

We now need to express the fraction variables $w_1$ and $w_2$ in terms of $z_3$. To do this, we recall from the \textit{equivalence relations} that 

\begin{equation}
w_1 = z_1 + z_3
 \label{equiv-for-w1}
\end{equation}

\begin{equation}
w_2 = z_2 + z_5
 \label{equiv-for-w1}
\end{equation}

Further, at equilibrium, $z_2$ = $z_5$, so we have $w_2 = 2z_2$. 

We have previously established that, at equilibrium, 

\begin{equation}
2z_2 = 0.5 - z_1 - z_3 .
 \label{equiv-relns-for-z2-expression-step3-repeat}
\end{equation}

Thus, we can write

\begin{equation}
w_2 = 0.5 - z_1 - z_3 .
 \label{w2-in-terms-of-z2}
\end{equation}

Using this expression for $w_2$, as well as the expression that $w_1 = z_1 + z_3$, we make substitutions into Eqn.\ref{z1q-times-z3q-step10} to obtain an expression involving only the $z_i$ configuration variables, given as

\begin{equation}
s  \frac {[0.5 - z_1 - z_3 ]}{z_1 + z_3}   =   [1/z_3 - 2s - 2]^2/16.
 \label{z1q-times-z3q-step11}
\end{equation}

We multiply the LHS of Eqn.~\ref{z1q-times-z3q-step11} through, top and bottom, by $2/z_3$ to obtain an equation that is only in terms of $s$ and $z_3$, given as

\begin{equation}
s  \frac {[1/z_3 - 2s - 2 ]}{2(s + 1)}   =   [1/z_3 - 2s - 2]^2/16,
 \label{z1q-times-z3q-step12}
\end{equation}

\noindent
which we can immediately rewrite as

\begin{equation}
\frac {s}{(s + 1)}   =   [1/z_3 - 2s - 2]/8.
 \label{z1q-times-z3q-step13}
\end{equation}

We now solve for $s$ in terms of $z_3$. Cross-multiplication gives

\begin{equation}
\frac {8s}{(s + 1)}   =   1/z_3 - 2s - 2,
 \label{z1q-times-z3q-step14}
\end{equation}

\noindent
and we reorganize terms to rewrite as

\begin{equation}
1/z_3 = \frac {8s}{(s + 1)} +2(s+1).
 \label{z1q-times-z3q-step15}
\end{equation}

We bring the terms on the RHS over the common denominator of $s+1$ and write

\begin{equation}
1/z_3 = \frac {8s +2(s+1)^2}{(s + 1)}, 
\label{z1q-times-z3q-step16}
\end{equation}

\noindent
which we can rewrite as

\begin{equation}
1/z_3 = \frac {2(4s +(s+1)^2}{(s + 1)} = \frac {2(4s +s^2 + 2s+1}{(s + 1)} = \frac {2(s^2 + 6s+1)}{(s + 1)} 
\label{z1q-times-z3q-step16}
\end{equation}

%
\vspace{8 mm}
\textbf{Step 8: Express $y_1$ and $y_2$ in terms of $z$  }
\vspace{3 mm}
%

We recall our initial defintion of $s$ in Eqn.~\ref{s-initial-definition}, which we simplify, knowing that we have already set $\lambda=1$, so that we have

\begin{equation}
s = h  e^{2 \lambda}  \frac { y_1  } { y_2 } = h \frac { y_1  } { y_2 }.
\label{s-definition-second-time}
\end{equation}

We now wish to obtain an expression for $y_2/y_1$ in terms of $s$ and $z_3$, so we recall our \textit{equivalence expressions} for $y_1$ and $y_2$ as

\begin{subequations} \label{sub:y-def-set-appendix-second-time}
\begin{gather} 
  y_1 = z_1 + z_2  \label{sub:y1-z1-z2-reln-appendix-second-time} \\
  y_2 = z_2 + z_4 = z_3+z_5 \label{sub:y2-z2-z4-and-z3-z5-reln-appendix-second-time}\\
\end{gather}
\end{subequations}

We also recall that at equlibrium, when $x_1 = x_2$, we also have $y_1 = y_3$ and that there are a set of equalities among the $z_i$, specifically that $z_3 = z_4$. We further recall Eqn.~\ref{equiv-relns-for-z2-expression-step3-repeat}, in which we expressed $z_2$ in terms of $z_1$ and $z_3$

\begin{equation}
2z_2 = 0.5 - z_1 - z_3 .
 \label{equiv-relns-for-z2-expression-step3-repeat}
\end{equation}

This allows us to write 

\begin{equation}
  y_1 = z_1 + z_2 = z_1 + [ 0.5 - z_1 - z_3]/2
\label{y1-in-z1-and-z3}
\end{equation}

\noindent
and

\begin{equation}
  y_2 = z_2 + z_4 = z_2 + z_3 =  [ 0.5 - z_1 - z_3]/2 + z_3.
\label{y2-in-z1-and-z3}
\end{equation}

\noindent
and

\begin{equation}
  y_2 = [ 0.5 - z_1 + z_3]/2.
\label{y2-in-z1-and-z3}
\end{equation}

%
\vspace{8 mm}
\textbf{Step 9: Obtain $y_2/y_1$  }
\vspace{3 mm}
%

We combine Eqns.~\ref{y1-in-z1-and-z3} and \ref{y2-in-z1-and-z3}, and multiply both top and bottom by 2, to create an expression for $y_2/y_1$ as

\begin{equation}
  y_2/y_1 =   \frac { 0.5 - z_1 + z_3} {0.5 + z_1  - z_3} =  \frac { 1 - 2z_1 + 2z_3} {1 + 2z_1  - 2z_3}
\label{y2-div-y1}
\end{equation}

We recall the earlier Eqn.~\ref{s-initial-definition} of $s$ in terms of $y_1$ and $y_2$, and also, from Eqn.~\ref{s-initial-definition-z1-and-z3}, in terms of $z_1$ and $z_3$, where we had

\begin{equation}
s=e^{2\varepsilon} e^{2 \lambda}  \frac { y_1  } { y_2 } = h e^{2 \lambda}  \frac { y_1  } { y_2 } = h  \frac { y_1  } { y_2 }.
\label{s-initial-definition-repeat-2}
\end{equation}

We rearrange to isolate $h$ as

\begin{equation}
 h  = s \frac { y_2  } { y_1 }.
\label{h-sqrd-as-s-y1-and-y2}
\end{equation}

We substitute the expression that we have just obtained for $y_2/y_1$ into this equation to obtain

\begin{equation}
h =  s \frac { 1 - 2z_1 + 2z_3} {1 + 2z_1  - 2z_3}
\label{s-evolution-step1}
\end{equation}

We divide through, top and bottom, by $z_3$, and use the expression that $s = z_1/z_3$, to obtain

\begin{equation}
 h =  s  \frac { 1/z_3 - 2s + 2} {1/z_3 + 2s  - 2}
\label{s-evolution-step3}
\end{equation}

We recall Eqn.~\ref{z1q-times-z3q-step16}, which gave us the expression for $1/z_3$ as

\begin{equation}
1/z_3 = \frac {2(s^2 + 6s+1)}{(s + 1)} 
\label{z1q-times-z3q-step16-repeat}
\end{equation}

We substitute from Eqn.~\ref{z1q-times-z3q-step16-repeat} into Eqn.~\ref{s-evolution-step3} to obtain

\begin{equation}
 h =  s \frac { [ \frac {2(s^2 + 6s+1)}{(s + 1)} ] - 2s + 2} {[ \frac {2(s^2 + 6s+1)}{(s + 1)} ] + 2s  - 2}
\label{s-evolution-step4}
\end{equation}

We multiply Eqn.~\ref{s-evolution-step4} through, top and bottom, by $s+1$, and divide through by 2, to obtain

\begin{equation}
 h =  s \frac { (s^2 + 6s+1)  - (s - 1)(s+1)} {(s^2 + 6s+1) + (s  -1)(s+1)}
\label{s-evolution-step5}
\end{equation}

\noindent
or

\begin{equation}
 h =  s \frac { s^2 + 6s+1  - s^2 + 1} {s^2 + 6s+1 + s^2  -1}
\label{s-evolution-step6}
\end{equation}

\noindent
or

\begin{equation}
 h =  s \frac {  6s + 2 } {2s^2 + 6s } =  \frac {  3s + 1 } {s + 3 }.
\label{s-evolution-step7}
\end{equation}

We now create an expression for $s$ in terms of $h$, and the first step is 

\begin{equation}
 h (s + 3 ) =    3s + 1 .
\label{s-evolution-step8}
\end{equation}

We collect like terms and obtain

\begin{equation}
  s(h  -  3)  =    1 - 3 h, 
\label{s-evolution-step9}
\end{equation}

\noindent
or

\begin{equation}
  s  =    \frac {1 - 3 h}{h  -  3 } 
\label{s-evolution-step10}
\end{equation}

%
\vspace{8 mm}
\textbf{Step 10: Obtain $1/z_3$ in terms of $h$  }
\vspace{3 mm}
%

Once again, we recall Eqn.~\ref{z1q-times-z3q-step16}

\begin{equation}
1/z_3 = \frac {2(s^2 + 6s+1)}{(s + 1)}. 
\label{z1q-times-z3q-step16_repeat}
\end{equation}

We substitute our expression for $s$ from Eqn.~\ref{s-evolution-step10} into Eqn.~\ref{z1q-times-z3q-step16} to obtain

\begin{equation}
1/z_3 = \frac {2}{ \left(  \frac {1 - 3 h}{h  -  3 } + 1\right)}  \left[ \left(   \frac {1 - 3 h}{h  -  3 } \right)^2 + 6 \frac {1 - 3 h}{h -  3 }+1 \right]. 
\label{1-over-z3-step1}
\end{equation}

We multiply through, top and bottom, by $(h - 3)^2$ to obtain

\begin{equation}
1/z_3 = \frac {2 (h - 3)^2} {( h  -  3) \left( (1 - 3 h) + (h - 3)\right)}  \left[ \left(   \frac {1 - 3 h}{h  -  3 } \right)^2 + 6 \frac {1 - 3 h}{h  -  3 }+1 \right]
\label{1-over-z3-step2}
\end{equation}

\noindent
or

\begin{equation}
1/z_3 = \frac {2 (h - 3)^2} {( h  -  3) \left( (-2) (  h  +1)\right)}  \left[ \left(   \frac {1 - 3 h}{h  -  3 } \right)^2 + 6 \frac {1 - 3 h}{h  -  3 }+1 \right]
\label{1-over-z3-step3}
\end{equation}

\noindent
or

\begin{equation}
1/z_3 = \frac {(-1)} {( h  -  3) ( h  +1)}  \left[  (1 - 3 h)^2 + 6  (1 - 3 h) (h - 3)+ (h - 3)^2 \right]
\label{1-over-z3-step4}
\end{equation}

We simplify the terms on the right-hand-side to obtain

\begin{equation}
 (1 - 3 h)^2 = 1 - 6h + 9 h^2,
\label{RHS-term1}
\end{equation}

\begin{equation}
6  (1 - 3 h) (h - 3) = 6(h - 3h^2 - 3 +9 h) = -6 (3h^2 -10h + 3) = -18h^2 +60h -18,
\label{RHS-term2}
\end{equation}

\noindent
and

\begin{equation}
 (h - 3)^2 = h^2 - 6h + 9.
\label{RHS-term3}
\end{equation}

Combining all terms, we obtain

\begin{equation}
 (1 - 3 h)^2 + 6  (1 - 3 h) (h - 3)+ (h - 3)^2 = 1 - 6h + 9 h^2  - 18h^2 +60h -18 + h^2 - 6h + 9.
\label{RHS-combined-terms1}
\end{equation}

Identifying this simply as the RHS (RHS-term), we combine like terms

\begin{equation}
RHS = ( 9 - 18 +1)  h^2  + ( - 6 +60 - 6 )h  -18 + 1+9 = -8h^3 + 48h -8 = 8(-h^2 +6h - 1)
\label{RHS-combined-terms1}
\end{equation}

We substitute this expression for the RHS back into Eqn.~\ref{1-over-z3-step3} to obtain

\begin{equation}
1/z_3 =\frac {(-1)} {( h  -  3) ( h  +1)}  \left[  8(-h^2 +6h - 1) \right]
\label{1-over-z3-step5}
\end{equation}

We invert Eqn.~\label{1-over-z3-step5} to obtain

\begin{equation}
z_3 = ( h  -  3) ( h  +1) \frac {(-1)} { \left[  8(-h^2 +6h - 1) \right]}
\label{z3-final-1}
\end{equation}

\noindent
or

\begin{equation}
z_3 =  \frac {( -h  +  3) ( h  +1)} { \left[  8(-h^2 +6h - 1) \right]}
\label{z3-final-2}
\end{equation}

%
\vspace{8 mm}
\textbf{Success at last!  }
\vspace{3 mm}
%

Kikuchi and Brush present, as their Eqn. (I.24)

\begin{equation}
\Delta =-h^2 +6h - 1,
\label{Kikuch-and-Brush-Delta-def-Eqn-Ipt24}
\end{equation}

and then give, as one element of their Eqn. (I.25),

\begin{equation}
z_3 =  \frac {( -h  +  3) ( h  +1)} { 8 \Delta }
\label{z3-K-and-B}
\end{equation}

%
\vspace{8 mm}
\textbf{The remaining configuration variables}
\vspace{3 mm}
%

It is not necessary to repeat this entire (and tedious) sequence in order to obtain the remaining configuration variables. Instead, we can simply use the equivalence relations to obtain most of the equations. 

However, we do first need to obtain an expression for $z_1$. A useful starting place is Eqn.~\ref{s-evolution-step10}, which gives an expression for $s$ in terms of $h$ as

\begin{equation}
  s  =    \frac {1 - 3 h}{h  -  3 }. 
\label{s-evolution-step10-repeat}
\end{equation}

Since we know that $s = z_1/z_3$, and we have just obtained an expression for $z_3$, it is straightforward to do the substitutions and obtain $z_1$. Then, we can use the equivalence relations presented at the beginning of the Appendix to obtain the remaining configuration variables. 

For completeness, these are listed below as (to be filled in) ...

%
\section{Appendix B: Computing the enthalpy interaction parameter from the \textit{h}-value}
\label{sec:Appendix-B-computing-enthalpy-interaction-parameter}
%

\renewcommand{\theequation}{B-\arabic{equation}}
\setcounter{equation}{0}  

If we make the assumption that the system is at equilibrium, we can estimate \textit{h} using Eqn.~\ref{s-initial-definition-z1-and-z3} from Appendix A; that is

\begin{equation}
\label{compute-h-given-config-vars-eqn}
 z_1/z_3 = y_1*h^2/y_2 
\end{equation}

This allows us to write \textit{h} as

\begin{equation}
\label{compute-h-given-config-vars-eqn_v1}
 h = \sqrt{{\frac {z_1*y_2}{z_3*y_1}}} 
\end{equation}

As an example of using this, we pick up on the discussion begun in Section~\ref{sec:phase_space-activation-enthalpy-zero}. This discussion referenced Figure~\ref{fig:CVM-2D_scale-free_2017-12-12}, which illustrates select counted ($Z_1$ and $Z_3$) and fractional ($z_1$ and $z_3$) computational variable values for an initial scale free-like grid. To compute our \textit{target h-value}, we also need $y_1$ and $y_2$.

\begin{figure}[ht]
  \centering
  \fbox{
  \rule[-.5cm]{0cm}{4cm}\rule[-.5cm]{0cm}{0cm}	
  \includegraphics [trim=0.0cm 0cm 0.0cm 0cm, clip=true,   width=0.8\linewidth]{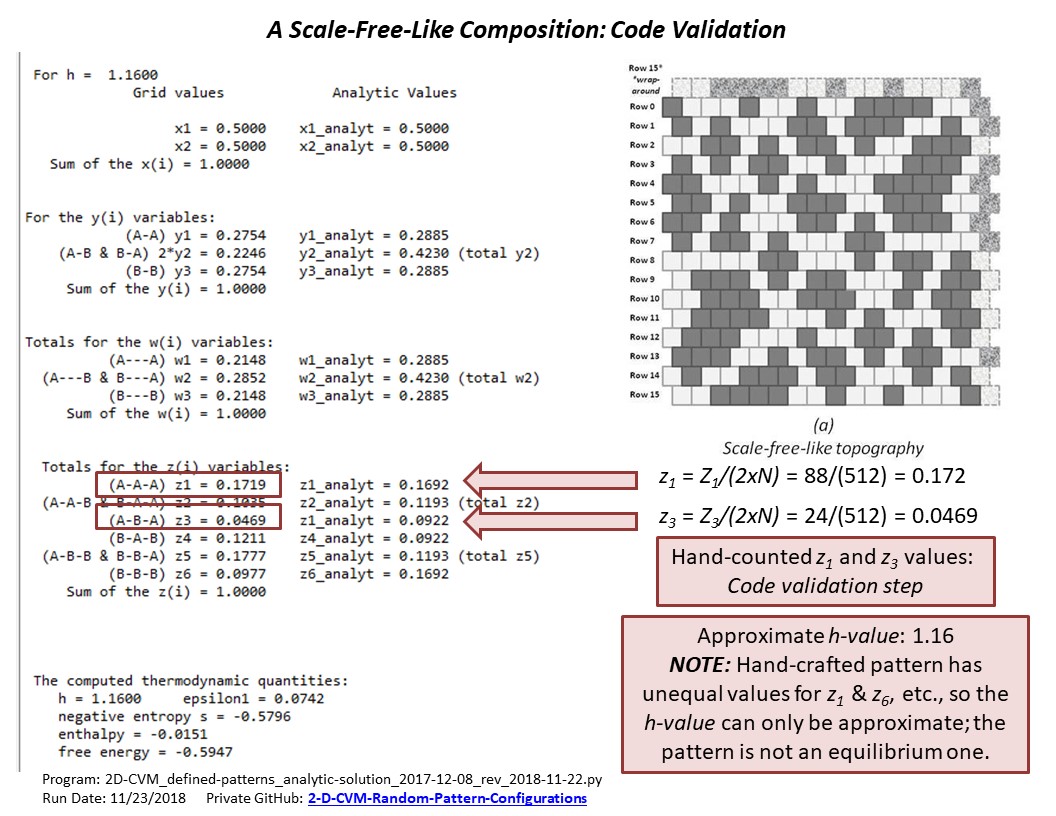}} 
  \vspace{3mm} 
  \caption{Configuration variable value counts for an initial manually-designed scale free-like 2-D CVM system. }   
\label{fig:2D-CVM_Expts-vary-eps0-1_comput_scale-free-results_2019-09-17}
\end{figure}

To obtain the additional configuration variable values $y_1$ and $y_2$, we refer to the original experimental results. The results summary is shown in Figure~\ref{fig:2D-CVM_Expts-vary-eps0-1_comput_scale-free-results_2019-09-17}. This shows the results for $y_1$ and $y_2$ as well as $z_1$ and $z_3$. The grid here is the same manually-designed scale free-like system as shown in Figure~\ref{fig:CVM-2D_scale-free_2017-12-12}.

\vspace{3mm} 

These results (given in a slidedeck summarizing select experimental trials \cite{AJMaren_2019_Expt-Results_Two-epsilon-params}) are summarized in the following Table~\ref{tbl:config-variables-table-initial-scale-free-grid}. 

\begin{table}[t]
    \caption{Configuration Variable Values for Different Values of $\varepsilon_0$ }
    \label{tbl:config-variables-table-initial-scale-free-grid}
    \centering
    \vspace{3mm}
    \begin{tabular}{|p{2cm}|p{2cm}|p{2cm}|p{2cm}|p{2cm}|}
    \hline
	 \multicolumn{1}{|>{\centering\arraybackslash}m{2cm}}	{\textbf{$y_1$}} 
    & 	 \multicolumn{1}{|>{\centering\arraybackslash}m{2cm}}	{\textbf{$y_2$}}       
    & 	 \multicolumn{1}{|>{\centering\arraybackslash}m{2cm}}	{\textbf{$z_1$}}               
    &   \multicolumn{1}{|>{\centering\arraybackslash}m{2cm}|}   {\textbf{$z_3$}} \T\B \\ 
    \hline    	    

	 \multicolumn{1}{|>{\centering\arraybackslash}m{2cm}}	{$0.2754$} 
     &   \multicolumn{1}{|>{\centering\arraybackslash}m{2cm}}	{$0.2246$} 	    	        
    &   \multicolumn{1}{|>{\centering\arraybackslash}m{2cm}}	{$0.1719$} 			  
    &   \multicolumn{1}{|>{\centering\arraybackslash}m{2cm}|}	{$0.0469$} \\ [3pt] 	

    \hline
  \end{tabular}
\end{table}

\begin{figure}[ht]
  \centering
  \fbox{
  \rule[-.5cm]{0cm}{4cm}\rule[-.5cm]{0cm}{0cm}	
  \includegraphics [trim=0.0cm 0cm 0.0cm 0cm, clip=true,   width=0.8\linewidth]{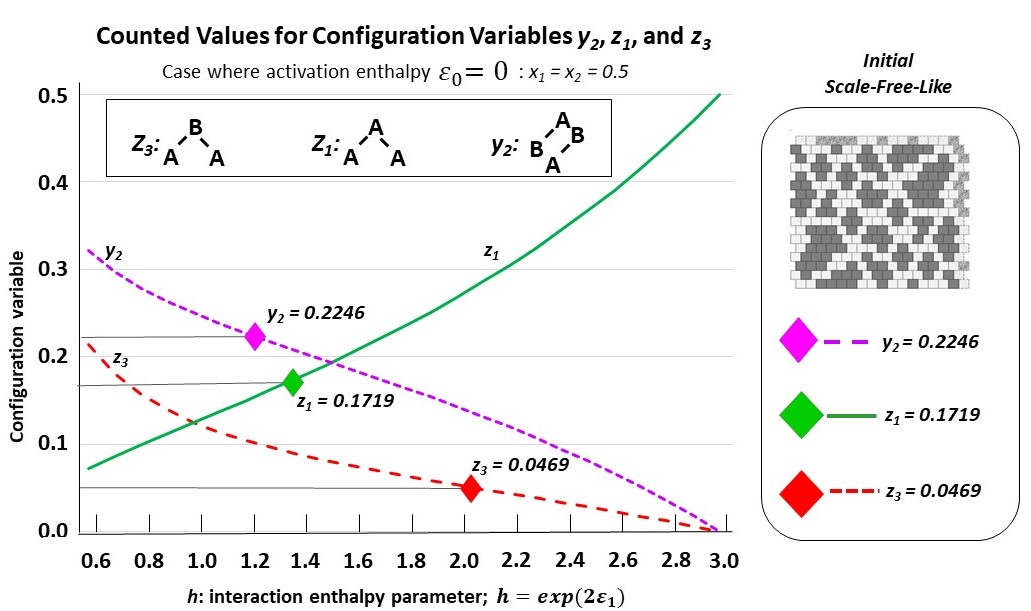}} 
  \vspace{3mm} 
  \caption{Configuration variable value counts for an initial manually-designed scale free-like 2-D CVM system. }   
\label{fig:2D-CVM-config-vars-vs-h_redone_scale-free_crppd_2019-09-17}
\end{figure}

We can visually see what the corresponding \textit{h-values} are for each of the configuration variables $y_2$, $z_1$, and $z_3$ are in Figure~\ref{fig:2D-CVM-config-vars-vs-h_redone_scale-free_crppd_2019-09-17}.

\vspace{3mm} 

From this, we can see some approximate \textit{h-values}. There is clearly a great deal of difference in the \textit{h-values} corresponding to each of the different configuration variables. This is because the system is not at equilibrium. 

We can take any one of several approaches to obtaining a \textit{target h-value}. One of these methods, which we discuss here, will be to use the results from the same analytic approach as discussed previously in Appendix A, to give a computed \textit{h-value}. 

If we apply this approach here, we obtain 

\begin{equation}
\label{compute-h-given-config-vars-eqn_v2}
 h = \sqrt{{\frac {0.1719*0.2246}{0.0469*0.2754}}} =  \sqrt{2.989} = 1.729
\end{equation}

This is a relatively high \textit{h-value} for this system. It has obviously been skewed upwards by the very low value for $z_3 = 0.0459$. 

This illustrative walk-through tells us two things. First, relying on the analytic solution to help us obtain a useful \textit{target h-value} is not necessarily the best approach, especially since we know that the analytic solution diverges for higher values of \textit{h}. 

Second, and perhaps more important, we should not blindly assume that all configuration variable values are created equal. In this illustration, the value for $z_3$ is associated with a much higher \textit{h-value} than the \textit{h-values} associated with $y_1$ and $z_3$, respectively. (We're not including $y_1$ in this discussion; we can get enough insight by looking at $z_3$.) 

When we look at Figure~\ref{fig:2D-CVM-config-vars-vs-h_redone_scale-free_crppd_2019-09-17}, we see that a reasonable \textit{target h-value} (one that corresponds to the \textit{h-values} for $y_2$ and $z_1$) would be in the neighborhood of $h = 1.2$. This is a relatively small \textit{h-value}. 

If we were to bring the system into equilibrium around $h= 1.2$, we would expect that the the $z_3$ value would shift and become a bit larger. Specifically, we would expect that $z_3$ would increase from 0.0469 (its current value) to about 0.1. 

The $z_3$ configuration value expresses the relative fraction of \textbf{A}-\textbf{B}-\textbf{A} triplets. One of the most obvious ways to increase $z_3$ is create many small ``islands''``landmasses'' of \textbf{A} units that are relatively close to each other. Specifically, they should separated by narrow channels that are only one unit of \text{B} in width.  

We can imagine what would happen if we brought this grid into a free energy-minimized state around $h=1.2$, so that the $z_3$ value would increase. This would mean that we'd have more ``narrow channels'' between the various islands of \textbf{A} units. We might expand the boundaries of certain landmasses to reduce the distance between them. 

Since $z_3$ is too small, compared with what we'd expect if we had $h=1.2$, it's interesting to take a look at the value for $z_4$ is also too small. Consulting the experimental results (see Figure~\ref{fig:2D-CVM_Expts-vary-eps0-1_comput_scale-free-results_2019-09-17}, we find that $z_4 = 0.1211$. This is actually very close to what we'd expect for $h = 1.2$. In fact, it is a bit higher than what we'd expect. 

The $z_4$ value reflects the fraction of \textbf{B}-\textbf{A}-\textbf{B} triplets. When we examine Figure~\ref{fig:2D-CVM-config-vars-vs-h_redone_scale-free_crppd_2019-09-17} (or go back to the original Figure~\ref{fig:CVM-2D_scale-free_2017-12-12}), we see that there are indeed many narrow ``channels'' of \textbf{B} units separating the islands of \textbf{A} units. 

\begin{figure}[ht]
  \centering
  \fbox{
  \rule[-.5cm]{0cm}{4cm}\rule[-.5cm]{0cm}{0cm}	
  \includegraphics [trim=0.0cm 0.0cm 0.0cm 0cm, clip=true, width=0.7\linewidth]{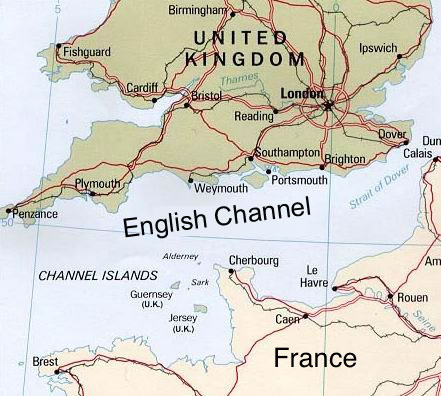}} 
  \vspace{3mm} 
  \caption{Illustration of a narrow channel (the English Channel) in geographic topology. Image is in the public domain, see Wikimedia for the EnglishChannel: ``This image is a work of a Central Intelligence Agency employee, taken or made as part of that person's official duties. As a Work of the United States Government, this image or media is in the public domain in the United States.'' }   
\label{fig:English_Channel-wikimedia_2019-09-20}
\end{figure}

In this sense, we can see that there is some similarity between these narrow channels and the kind of naturally-occurring channels that occur in a physical topography. A good example of this would be how the English Channel separates Great Britain from France (see Figure~\ref{fig:English_Channel-wikimedia_2019-09-20}). 

\vspace{3mm} 

As a contrasting discussion, we can think through what we'd expect in terms of how the topography would change if $h$ were increased, e.g., $h = 1.65$. (This is pretty close to the \textit{h-value} that we obtained via analytic solution at the beginning of this Appendix; where we found $h = 1.729$.)

When we consider natural topographies that have a relatively high ``agglutinative'' factor, meaning that the land coalesces largely into larger islands and landmasses, we see that there is not that much in the way of very closely-spaced little islands. Lots of little islands would be more characteristic of a lower \textit{h-value}, and when we push the \textit{h-value} to be less than zero, we wind up moving towards a checkerboard-like pattern; the islands get smaller and the space between them is narrow as well. 

If we increase the \textit{h-value}, then we'd expect to be joining some of these islands together. That is, we'd increase $z_1$ (the relative fraction of \textbf{A}-\textbf{A}-\textbf{A} triplets), and decrease $y_2$ (the relative fraction of boundaries between unlike masses, that is, the  \textbf{A}-\textbf{B} and  \textbf{B}-\textbf{A} pairs). 

\begin{figure}[ht]
  \centering
  \fbox{
  \rule[-.5cm]{0cm}{4cm}\rule[-.5cm]{0cm}{0cm}	
  \includegraphics [trim=0.0cm 0.0cm 0.0cm 0cm, clip=true, width=0.7\linewidth]{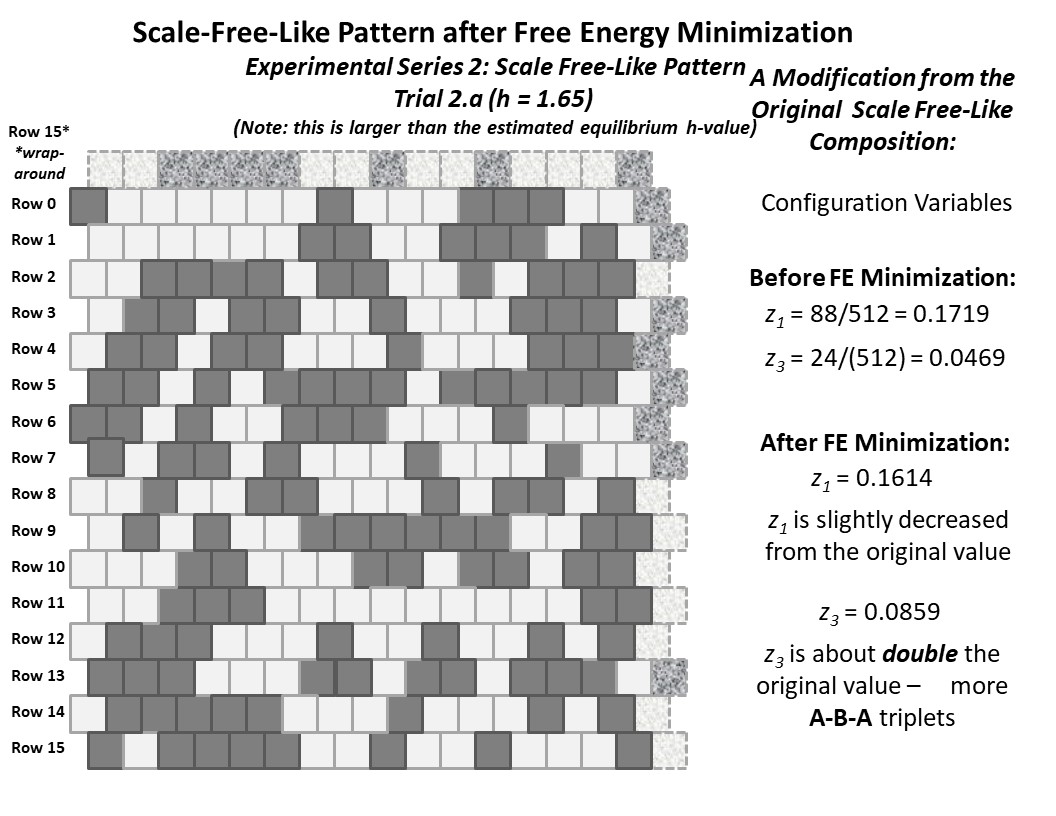}} 
  \vspace{3mm} 
  \caption{The result of free energy minimization applied to the initial scale-free-like grid, for \textit{h = 1.65}. }   
\label{fig:2D-CVM_Expts-vary-eps0-1_comput_scale-free_FEM_h-eq-1pt65_2019-09-17}
\end{figure}

Figure~\ref{fig:2D-CVM_Expts-vary-eps0-1_comput_scale-free_FEM_h-eq-1pt65_2019-09-17} shows the results of a free energy minimization trial applied to this initial grid, when we set $h = 1.65$. 

\vspace{3mm} 

We can see, in Figure~\ref{fig:2D-CVM_Expts-vary-eps0-1_comput_scale-free_FEM_h-eq-1pt65_2019-09-17}, that we actually have combined many of the isolated islands into contiguous landmasses. What is particularly interesting here is that we can see several new instances of long \textbf{A}-\textbf{A}-\textbf{A} channels, increasing the $z_3$ value. (A good instance is in the left-hand-side of figure.)

Although $z_3$ is actually \textit{increased}, we can refer back to Figure~\ref{fig:2D-CVM-config-vars-vs-h_redone_scale-free_crppd_2019-09-17}. Considering what has happened, we can see that the increase in $z_3$ is a natural and expected result; the original value was too small, even for a relatively high \textit{h-value}. 

In this resulting configuration, the value for $z_1$ is actually decreased (a little), and we would have expected a slight increase. This is most likely due to the simplistic nature of the free energy minimization algorithm; it is currently a simple random pick-and-switch process. 

If we were working with a more sophisticated, topographically-sensitive free energy minimization algorithm, we would likely see that more of the islands would coagulate into landmasses. We would also likely see more ``rivers'' of \textbf{B} units flowing out from these landmasses of \textbf{A} units. 

This suggests, of course, that it would be fascinating to characterize natural topographies using this 2-D CVM minimization process.

%
\section{Appendix C: Computing $\varepsilon_0$ for a given $x_1$ when $\varepsilon_1 = 0$}
\label{sec:Appendix-C-computing-activation-enthalpy-iparameter-given-x1}
%

\renewcommand{\theequation}{C-\arabic{equation}}
\setcounter{equation}{0}  

Our goal in this Appendix is to find the correlation between $x_1$ values and the $\varepsilon_0$ activation enthalpy parameter, for the case where $\varepsilon_1 = 0$.

We begin with the free energy equation, previously introduced in Section \ref{subsec:2D-CVM-entropy} as Eqn.~\ref{eqn:Bar-F-2-D-basic-eqn}, and repeated here for convenience as

\begin{equation}
\label{eqn:Bar-F-2-D-basic-eqn-appendix-C}
  \begin{aligned}
\bar{F}_{2-D} = F_{2-D}/N = \\
  & \varepsilon_1(-z_1+z_3+z_4-z_6) - \bar{S}_{2-D}\\
+ & \mu (1-\sum\limits_{i=1}^6 \gamma_i  z_i )+4 
\lambda (z_3+z_5-z_2-z_4)
  \end{aligned}
\end{equation}

We will change this equation by setting $\varepsilon_1 = 0$ and introducing the term containing the activation enthalpy $\varepsilon_0$. (See Eqn.~\ref{eqn:Bar-H-2-D-basic-eqn}.) Also, since we will be taking the derivative of this equation with respect to $x_1$, and we know that we will set the Lagrangian parameters equal to zero (as we did in Appendix A), we will remove them from this next iteration of the equation, giving us

\begin{equation}
\label{eqn:Bar-F-2-D-basic-eqn-appendix-C}
\bar{F}_{2-D} =  \varepsilon_0 x_1 - \bar{S}_{2-D}
\end{equation}

We take the derivative with respect to $x_1$

\begin{equation}
\label{eqn:Bar-F-2-D-basic-eqn-derivative-appendix-C}
0 = \frac {\partial \bar{F}_{2-D}} {\partial x_1}  =  \varepsilon_0  - \frac {\partial } {\partial x_1}  \Big [  \bar{S}_{2-D} \Big]. 
\end{equation}

We recall, from Eqn.~\ref{eqn:Bar-S-2-D-basic-eqn} Subsection~\ref{subsec:2D-CVM-entropy}, our definition for $\bar{S}_{2-D}$ as

\begin{equation}
\label{eqn:Bar-S-2-D-basic-eqn-App-C}
  \begin{aligned}
\bar{S}_{2-D} = S_{2-D}/N = \\
 & 2 \sum\limits_{i=1}^3 \beta_i Lf(y_i))
          + \sum\limits_{i=1}^3 \beta_i Lf(w_i) \\
 &      - \sum\limits_{i=1}^2 Lf(x_i)
          - 2 \sum\limits_{i=1}^6 \gamma_i Lf(z_i), \\
  \end{aligned}
\end{equation}

 \noindent
where $Lf(v)=vln(v)-v$.

We recall, from Appendix A, the derivative of $Lf(v)$ as

\begin{equation}
 \frac {\partial Lf(v)} {\partial v}  =  \frac {\partial \big[ v\  ln(v) - v   \big]} {\partial v} = ln(v).
\end{equation}

We recall, from earlier in this work, that when we had no interaction enthalpy and thus a random distribution of units into their various configurations, that the 2-D CVM entropy was the same as the basic entropy when the extra CVM ($y_i$, $w_i$, and $z_i$) were identical - that is, at this point, the extra configuration variables did not contribute a difference to the total 2-D CVM entropy. 

Thus, for simplicity, we'll look just at the entropy involving the $x_i$ terms. 

\begin{equation}
\label{eqn:simple-entropy-eqn-App-C}
\bar{S}_{2-D} = S_{2-D}/N =   - \sum\limits_{i=1}^2  Lf(x_i),
\end{equation}

\noindent
and thus

\begin{equation}
\label{eqn:basic-entropy-derivative-appendix-C}
  \begin{aligned}
  \frac {\partial } {\partial x_1}  \Big [  \bar{S}_{2-D} \Big]  \\
  & = - \frac {\partial } {\partial x_1}  \Big [  ln(x_1) +  ln(1-x_1)  \Big] \\
  & = - \Big [ ln(x_1) +  ln(1-x_1) \frac {\partial } {\partial x_1} (1-x_1)  \Big] \\
  & = - \Big [ ln(x_1) -  ln(1-x_1) \Big] = - ln(x_1) + ln(1-x_1).
  \end{aligned}
\end{equation}

We substitute this back into Eqn.~\ref{eqn:Bar-F-2-D-basic-eqn-derivative-appendix-C} to obtain

\begin{equation}
\label{eqn:Bar-F-2-D-basic-eqn-derivative-simple-appendix-C}
 \varepsilon_0  = \frac {\partial } {\partial x_1}  \Big [  \bar{S}_{2-D} \Big] =  ln(x_1) - ln(1-x_1). 
\end{equation}

We can obtain a very simple set of values for $\varepsilon_0$ from this. For example:

\begin{itemize}
\setlength{\itemsep}{1pt}
\item \textbf{$x_1 = 0.5$}:   $\varepsilon_0 =  -ln(0.5) + ln(1-0.5) = 0$
\item \textbf{$x_1 = 0.269$}:   $\varepsilon_0 =  -ln(0.269) + ln(0.731) = -( -1.313) +(-0.313) = 1.0$
\item \textbf{$x_1 = 0.119$}:   $\varepsilon_0 =  -ln(0.119) + ln(0.881) = -( -2.129) +(-0.127) = 2.002$
\end{itemize}

(Note: Other values for $x_1$ corresponding to nominal values for $\varepsilon_0$ were created using interpolation.)

Table~\ref{tbl:config-variables-values-table} identifies select configuration variable values for certain values of $\varepsilon_0$. Additionally, it identifies the number of corresponding units / pairs / triplets within a 256-unit 2-D CVM grid that is designed with the corresponding value for $x_1$. 

From this table, it is clear that - especially if we are dealing with a small-scale 2-D CVM grid (as we are doing here for tractability), it is in our best interests to keep $\varepsilon_0$ relatively small; e.g.  $\varepsilon_0 <1$. When we increase the value of  $\varepsilon_0$ much beyond that, then we run the risk of having too few active units to create even a single instance of each of the different kinds of triplets. 

\begin{table}[t]\footnotesize
    \caption{Configuration Variable Values for Different Values of $\varepsilon_0$ }
    \label{tbl:config-variables-values-table}
    \centering
    \vspace{3mm}
    \begin{tabular}{|p{2cm}|p{2cm}|p{2cm}|p{2cm}|p{2cm}|}
    \hline
	 \multicolumn{1}{|>{\centering\arraybackslash}m{2cm}}	{\textbf{$\varepsilon_0$}} 
    & 	 \multicolumn{1}{|>{\centering\arraybackslash}m{2cm}}	{\textbf{$x_1$}}  
    & 	 \multicolumn{1}{|>{\centering\arraybackslash}m{2cm}}	{\textbf{$y_1$}}       
    & 	 \multicolumn{1}{|>{\centering\arraybackslash}m{2cm}}	{\textbf{$z_1$}}               
    &   \multicolumn{1}{|>{\centering\arraybackslash}m{2cm}|}   {\textbf{$Z_1$}} \T\B \\ 
    \hline    	    

	 \multicolumn{1}{|>{\centering\arraybackslash}m{2cm}}	{$0$} 
    &   \multicolumn{1}{|>{\centering\arraybackslash}m{2cm}}	{$0.5$} 	
    &   \multicolumn{1}{|>{\centering\arraybackslash}m{2cm}}	{$0.25$} 	
     &   \multicolumn{1}{|>{\centering\arraybackslash}m{2cm}}	{$0.075$} 	   	        
    &   \multicolumn{1}{|>{\centering\arraybackslash}m{2cm}|}	{$19$} \\ [3pt] 			  
  
	 \multicolumn{1}{|>{\centering\arraybackslash}m{2cm}}	{$1.0$} 
    &   \multicolumn{1}{|>{\centering\arraybackslash}m{2cm}}	{$0.269$} 	
    &   \multicolumn{1}{|>{\centering\arraybackslash}m{2cm}}	{$0.0724$} 	
     &   \multicolumn{1}{|>{\centering\arraybackslash}m{2cm}}	{$0.0194$} 	   	        
    &   \multicolumn{1}{|>{\centering\arraybackslash}m{2cm}|}	{$5$} \\ [3pt]

	 \multicolumn{1}{|>{\centering\arraybackslash}m{2cm}}	{$2.0$} 
    &   \multicolumn{1}{|>{\centering\arraybackslash}m{2cm}}	{$0.119$} 	
    &   \multicolumn{1}{|>{\centering\arraybackslash}m{2cm}}	{$0.0141$} 	
     &   \multicolumn{1}{|>{\centering\arraybackslash}m{2cm}}	{$0.0017$} 	   	        
    &   \multicolumn{1}{|>{\centering\arraybackslash}m{2cm}|}	{$0.43$} \\ [3pt] 		 

	 \multicolumn{1}{|>{\centering\arraybackslash}m{2cm}}	{$3.0$} 
    &   \multicolumn{1}{|>{\centering\arraybackslash}m{2cm}}	{$0.047$} 	
    &   \multicolumn{1}{|>{\centering\arraybackslash}m{2cm}}	{$0.0022$} 	
     &   \multicolumn{1}{|>{\centering\arraybackslash}m{2cm}}	{$0.0001$} 	   	        
    &   \multicolumn{1}{|>{\centering\arraybackslash}m{2cm}|}	{$0.03$} \\ [3pt] 	

    \hline
  \end{tabular}
\end{table}

%
\section{Appendix D: Experimental results: configuration variables for different $x_1$ when varying $\varepsilon_1$}
\label{sec:Appendix-D-experimental-results-varying-x1}
%

\renewcommand{\theequation}{D-\arabic{equation}}
\setcounter{equation}{0}  

We are able to find the computational values for $y_2$, $z_1$, and $z_3$ (as well as others, if needed) for values of $x_1 \neq 0.5$ by following the \textit{Protocol} briefly identified in previous sections. 

Table \ref{tbl:config-variables-values-y2-analyt-and-computed-table} presents the values for $y_2$, for the cases where $x_1 = 0.5$ and $x_1 = 0.35$, as well as the analytic values for $y_2$. 

\begin{table}[t]\footnotesize
    \caption{Configuration Variable \textbf{$y_2$} Values for Different Values of $\varepsilon_1$ and $x_1$ }
    \label{tbl:config-variables-values-y2-analyt-and-computed-table}
    \centering
    \vspace{3mm}
    \begin{tabular}{|p{2cm}|p{2cm}|p{2cm}|p{2cm}|p{2cm}|}
    \hline
	 \multicolumn{1}{|>{\centering\arraybackslash}m{2cm}}	{\textbf{$\varepsilon_1$}} 
    & 	 \multicolumn{1}{|>{\centering\arraybackslash}m{2cm}}	{\textbf{$x_1 = 0.5$}}  
    & 	 \multicolumn{1}{|>{\centering\arraybackslash}m{2cm}|}	{\textbf{$x_1 = 0.35$}}    \T\B \\ 
    \hline    	    

	 \multicolumn{1}{|>{\centering\arraybackslash}m{2cm}}	{$0.8$} 
    &   \multicolumn{1}{|>{\centering\arraybackslash}m{2cm}}	{$0.342$} 		
    &   \multicolumn{1}{|>{\centering\arraybackslash}m{2cm}|}	{$0.301$} 	    \\ [3pt] 			  
  
	 \multicolumn{1}{|>{\centering\arraybackslash}m{2cm}}	{$0.9$} 
    &   \multicolumn{1}{|>{\centering\arraybackslash}m{2cm}}	{$0.300$} 	
    &   \multicolumn{1}{|>{\centering\arraybackslash}m{2cm}|}	{$0.268$} 	  \\ [3pt]  	        		   
	 
	 \multicolumn{1}{|>{\centering\arraybackslash}m{2cm}}	{$1.0$} 
    &   \multicolumn{1}{|>{\centering\arraybackslash}m{2cm}}	{$0.251$} 	
    &   \multicolumn{1}{|>{\centering\arraybackslash}m{2cm}|}	{$0.228$} 	   \\ [3pt] 		  
 
	 \multicolumn{1}{|>{\centering\arraybackslash}m{2cm}}	{$1.1$} 
    &   \multicolumn{1}{|>{\centering\arraybackslash}m{2cm}}	{$0.206$} 		
    &   \multicolumn{1}{|>{\centering\arraybackslash}m{2cm}|}	{$0.190$} 	    \\ [3pt] 	 

	 \multicolumn{1}{|>{\centering\arraybackslash}m{2cm}}	{$1.2$} 
    &   \multicolumn{1}{|>{\centering\arraybackslash}m{2cm}}	{$0.174$} 	
    &   \multicolumn{1}{|>{\centering\arraybackslash}m{2cm}|}	{$0.157$} 	   \\ [3pt] 	 

	 \multicolumn{1}{|>{\centering\arraybackslash}m{2cm}}	{$1.3$} 
    &   \multicolumn{1}{|>{\centering\arraybackslash}m{2cm}}	{$0.166$} 		
    &   \multicolumn{1}{|>{\centering\arraybackslash}m{2cm}|}	{$0.145$} 	    \\ [3pt] 	 

	 \multicolumn{1}{|>{\centering\arraybackslash}m{2cm}}	{$1.4$} 
    &   \multicolumn{1}{|>{\centering\arraybackslash}m{2cm}}	{$0.152$} 	
    &   \multicolumn{1}{|>{\centering\arraybackslash}m{2cm}|}	{$0.148$} 	    \\ [3pt] 	 
  
	 \multicolumn{1}{|>{\centering\arraybackslash}m{2cm}}	{$1.5$} 
    &   \multicolumn{1}{|>{\centering\arraybackslash}m{2cm}}	{$0.167$} 	
    &   \multicolumn{1}{|>{\centering\arraybackslash}m{2cm}|}	{$0.148$} 	   \\ [3pt] 	 

	 \multicolumn{1}{|>{\centering\arraybackslash}m{2cm}}	{$1.6$} 
    &   \multicolumn{1}{|>{\centering\arraybackslash}m{2cm}}	{$0.163$} 	
    &   \multicolumn{1}{|>{\centering\arraybackslash}m{2cm}|}	{$0.144$} 	   \\ [3pt] 	 

	 \multicolumn{1}{|>{\centering\arraybackslash}m{2cm}}	{$1.7$} 
    &   \multicolumn{1}{|>{\centering\arraybackslash}m{2cm}}	{$0.153$} 		
    &   \multicolumn{1}{|>{\centering\arraybackslash}m{2cm}|}	{$0.146$} 	    \\ [3pt] 	 

	 \multicolumn{1}{|>{\centering\arraybackslash}m{2cm}}	{$1.8$} 
    &   \multicolumn{1}{|>{\centering\arraybackslash}m{2cm}}	{$0.166$} 	   	        
    &   \multicolumn{1}{|>{\centering\arraybackslash}m{2cm}|}	{$0.151$} \\ [3pt] 	     
    \hline
  \end{tabular}
\end{table}

\begin{figure}[h!]
  \centering
  \fbox{
  \rule[-.5cm]{0cm}{4cm}\rule[-.5cm]{0cm}{0cm}	
  \includegraphics [trim=0.0cm 0.0cm 0.0cm 0cm, clip=true, width=0.6\linewidth]{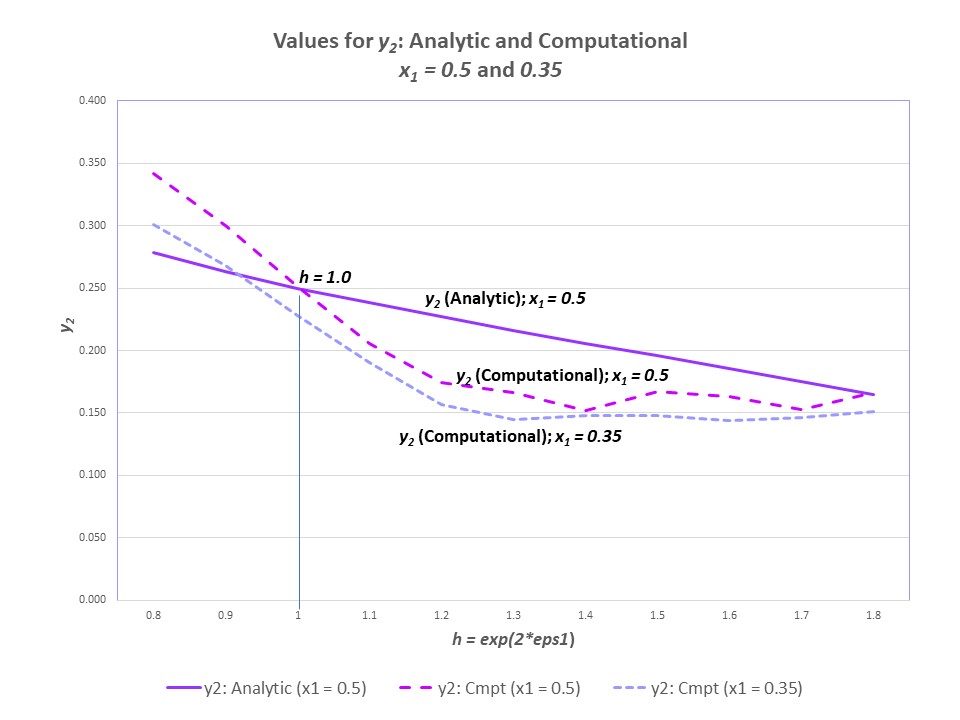}} 
  \vspace{3mm} 
  \caption{The $y_2$ values; analytic (when $x_1 = 0.5$), and computational (when $x_1 = 0.5, 0.35$ for \textit{h = 0.8 .. 1.8}. }   
\label{fig:2D-CVM-y2-analyt-and-comput_2019-09-18}
\end{figure}

Figure~\ref{fig:2D-CVM-y2-analyt-and-comput_2019-09-18} presents the same information as in Table~\ref{tbl:config-variables-values-y2-analyt-and-computed-table}.

\vspace{3mm} 

The computational results are obtained by averaging the results from twenty trials for each data point.

We can see that the computational value for $y_2$ when $x_1 = 0.5$ is $y_2 = 0.251$, which is acceptably close to the analytic value of $y_2 = 0.250$. However, the increased values of $y_2$ when $h<1.0$ (as compared with the analytic) are a bit surprising. We note that any value of $h<1.0$ pushes in the direction of the divergence that occurs in the analytic solution when $h = 0.172$.

For convenience, we reproduce the previous Figure~\ref{fig:Perturbation-Results_x1=0pt35_lbld_crppd_2018-01-22} here. 

\begin{figure}[ht]
  \centering
  \fbox{
  \rule[-.5cm]{0cm}{4cm}\rule[-.5cm]{0cm}{0cm}	
  \includegraphics [trim=0.0cm 0cm 0.0cm 0cm, clip=true,   width=0.95\linewidth]{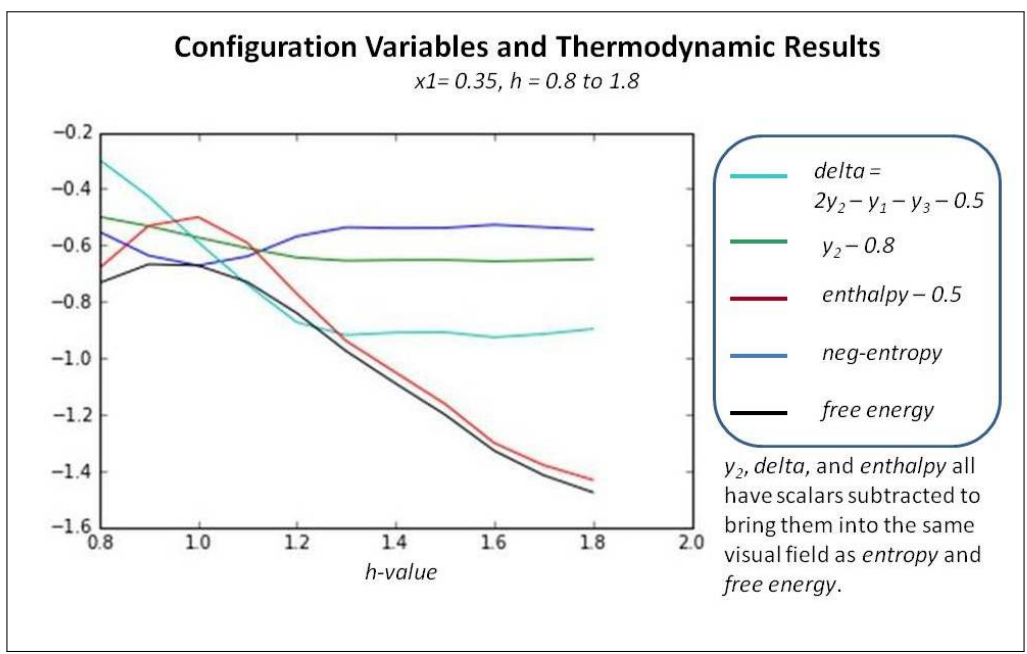}} 
  \vspace{3mm} 
  \caption{Configuration variable and thermodynamic values for the case where $x_1 = 0.35$ and $x_2 = 0.65$, and where the interaction enthalpy parameter $h$ ranges as $h = 0.8 .. 1.8$.}   
\label{fig:Perturbation-Results_x1=0pt35_lbld_crppd_2018-01-22-repeat}
\end{figure}
\vspace{3mm} 

If we refer to Figure~\ref{fig:Perturbation-Results_x1=0pt35_lbld_crppd_2018-01-22-repeat}, we see that the free energy (shown in black) drops off sharply when we move to either the left or the right of $h=1.0$. This is largely due to the influence of the enthalpy term (shown in maroon).

\begin{figure}[h!]
  \centering
  \fbox{
  \rule[-.5cm]{0cm}{4cm}\rule[-.5cm]{0cm}{0cm}	
  \includegraphics [trim=0.0cm 0.0cm 0.0cm 0cm, clip=true, width=0.6\linewidth]{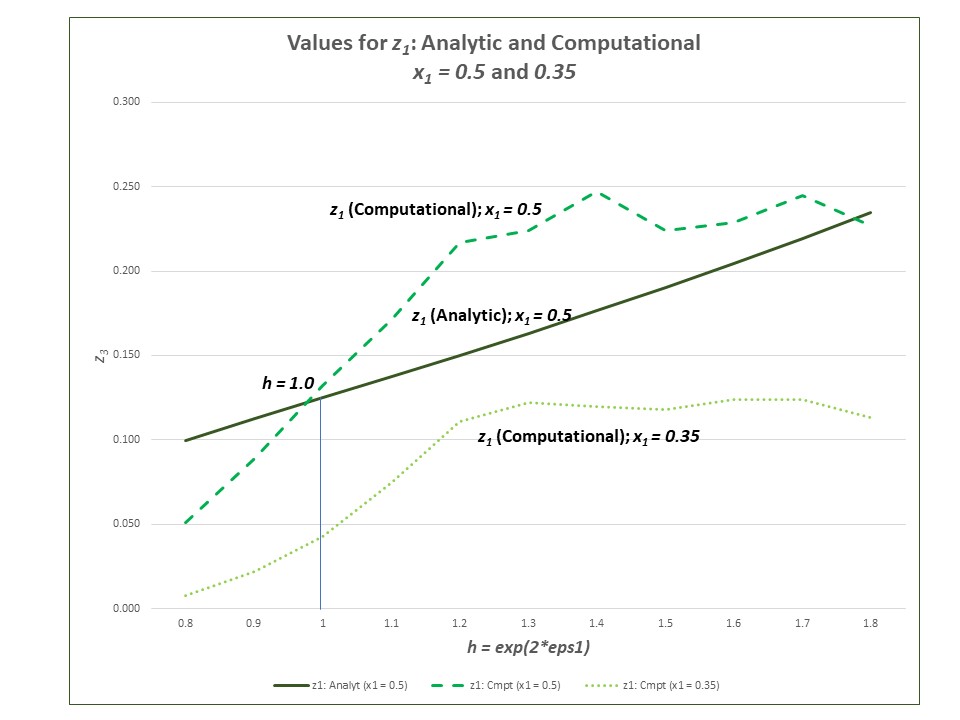}} 
  \vspace{3mm} 
  \caption{The $z_1$ values; analytic (when $x_1 = 0.5$), and computational (when $x_1 = 0.5, 0.35$ for \textit{h = 0.8 .. 1.8}. }   
\label{fig:2D-CVM-z1-analyt-and-comput_2019-09-18}
\end{figure}

We get similar results for $z_1$ and $z_3$. Table~\ref{tbl:config-variables-values-z1-analyt-and-computed-table} gives the results for $z_1$, and Figure~\ref{fig:2D-CVM-z1-analyt-and-comput_2019-09-18} gives us visual presentation of this data.

\begin{table}[h!]\footnotesize
    \caption{Configuration Variable \textbf{$z_1$} Values for Different Values of $\varepsilon_1$ and $x_1$ }
    \label{tbl:config-variables-values-z1-analyt-and-computed-table}
    \centering
    \vspace{3mm}
    \begin{tabular}{|p{2cm}|p{2cm}|p{2cm}|p{2cm}|p{2cm}|}
    \hline
	 \multicolumn{1}{|>{\centering\arraybackslash}m{2cm}}	{\textbf{$\varepsilon_1$}} 
    & 	 \multicolumn{1}{|>{\centering\arraybackslash}m{2cm}}	{\textbf{$x_1 = 0.5$}}           
    &   \multicolumn{1}{|>{\centering\arraybackslash}m{2cm}|}   {\textbf{$x_1 = 0.35$}} \T\B \\ 
    \hline    	    

	 \multicolumn{1}{|>{\centering\arraybackslash}m{2cm}}	{$0.8$} 
    &   \multicolumn{1}{|>{\centering\arraybackslash}m{2cm}}	{$0.051$} 		   	        
    &   \multicolumn{1}{|>{\centering\arraybackslash}m{2cm}|}	{$0.008$} \\ [3pt] 			  
  
	 \multicolumn{1}{|>{\centering\arraybackslash}m{2cm}}	{$0.9$} 
    &   \multicolumn{1}{|>{\centering\arraybackslash}m{2cm}}	{$0.089$} 	 	   	        
    &   \multicolumn{1}{|>{\centering\arraybackslash}m{2cm}|}	{$0.022$} \\ [3pt] 		   
	 
	 \multicolumn{1}{|>{\centering\arraybackslash}m{2cm}}	{$1.0$} 
    &   \multicolumn{1}{|>{\centering\arraybackslash}m{2cm}}	{$0.132$} 		   	        
    &   \multicolumn{1}{|>{\centering\arraybackslash}m{2cm}|}	{$0.043$} \\ [3pt] 		  
 
	 \multicolumn{1}{|>{\centering\arraybackslash}m{2cm}}	{$1.1$} 
    &   \multicolumn{1}{|>{\centering\arraybackslash}m{2cm}}	{$0.171$} 		   	        
    &   \multicolumn{1}{|>{\centering\arraybackslash}m{2cm}|}	{$0.075$} \\ [3pt] 	 

	 \multicolumn{1}{|>{\centering\arraybackslash}m{2cm}}	{$1.2$} 
    &   \multicolumn{1}{|>{\centering\arraybackslash}m{2cm}}	{$0.217$} 		   	        
    &   \multicolumn{1}{|>{\centering\arraybackslash}m{2cm}|}	{$0.111$} \\ [3pt] 	 

	 \multicolumn{1}{|>{\centering\arraybackslash}m{2cm}}	{$1.3$} 
    &   \multicolumn{1}{|>{\centering\arraybackslash}m{2cm}}	{$0.224$} 	   	        
    &   \multicolumn{1}{|>{\centering\arraybackslash}m{2cm}|}	{$0.122$} \\ [3pt] 	 

	 \multicolumn{1}{|>{\centering\arraybackslash}m{2cm}}	{$1.4$} 
    &   \multicolumn{1}{|>{\centering\arraybackslash}m{2cm}}	{$0.247$} 		   	        
    &   \multicolumn{1}{|>{\centering\arraybackslash}m{2cm}|}	{$0.120$} \\ [3pt] 	 
  
	 \multicolumn{1}{|>{\centering\arraybackslash}m{2cm}}	{$1.5$} 
    &   \multicolumn{1}{|>{\centering\arraybackslash}m{2cm}}	{$0.224$} 	   	        
    &   \multicolumn{1}{|>{\centering\arraybackslash}m{2cm}|}	{$0.118$} \\ [3pt] 	 

	 \multicolumn{1}{|>{\centering\arraybackslash}m{2cm}}	{$1.6$} 
    &   \multicolumn{1}{|>{\centering\arraybackslash}m{2cm}}	{$0.229$} 	   	        
    &   \multicolumn{1}{|>{\centering\arraybackslash}m{2cm}|}	{$0.124$} \\ [3pt] 	 

	 \multicolumn{1}{|>{\centering\arraybackslash}m{2cm}}	{$1.7$} 
    &   \multicolumn{1}{|>{\centering\arraybackslash}m{2cm}}	{$0.245$} 	   	        
    &   \multicolumn{1}{|>{\centering\arraybackslash}m{2cm}|}	{$0.124$} \\ [3pt] 	 

	 \multicolumn{1}{|>{\centering\arraybackslash}m{2cm}}	{$1.8$} 
    &   \multicolumn{1}{|>{\centering\arraybackslash}m{2cm}}	{$0.227$} 	   	        
    &   \multicolumn{1}{|>{\centering\arraybackslash}m{2cm}|}	{$0.113$} \\ [3pt] 	     
    \hline
  \end{tabular}
\end{table}

\vspace{3mm} 

\begin{figure}[h!]
  \centering
  \fbox{
  \rule[-.5cm]{0cm}{4cm}\rule[-.5cm]{0cm}{0cm}	
  \includegraphics [trim=0.0cm 0.0cm 0.0cm 0cm, clip=true, width=0.7\linewidth]{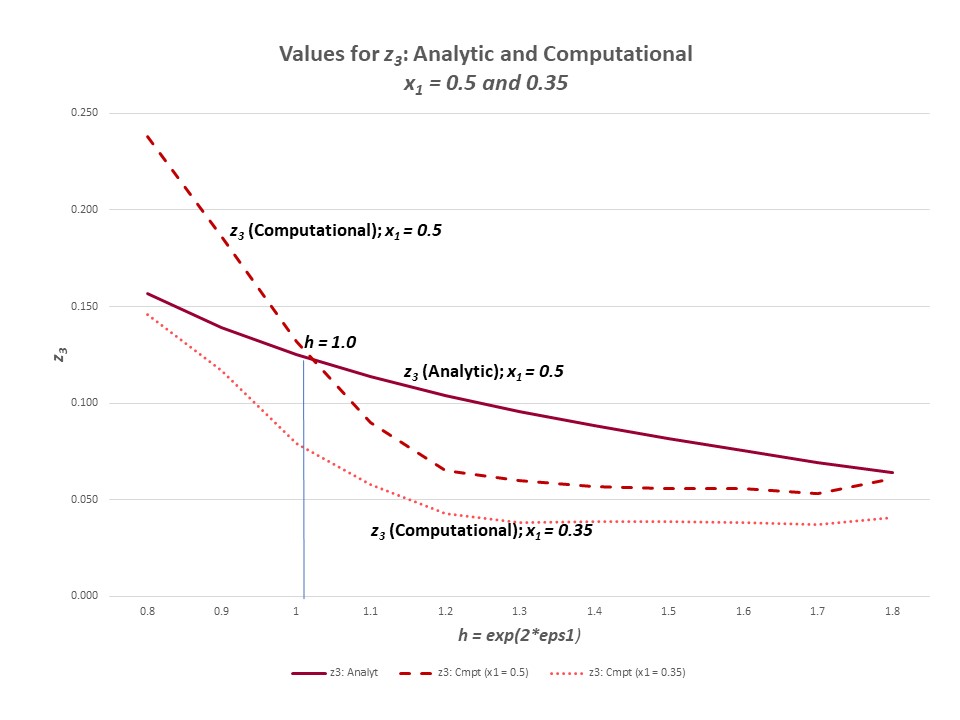}} 
  \vspace{3mm} 
  \caption{The $z_3$ values; analytic (when $x_1 = 0.5$), and computational (when $x_1 = 0.5, 0.35$ for \textit{h = 0.8 .. 1.8}. }   
\label{fig:2D-CVM-z3-analyt-and-comput_2019-09-18}
\end{figure}

Table~\ref{tbl:config-variables-values-z3-analyt-and-computed-table} gives the results for $z_3$.  for $z_1$, and Figure~\ref{fig:2D-CVM-z3-analyt-and-comput_2019-09-18} gives us a similar presentation for $z_3$.

\begin{table}[h!]
    \caption{Configuration Variable \textbf{$z_3$} Values for Different Values of $\varepsilon_1$ and $x_1$ }
    \label{tbl:config-variables-values-z3-analyt-and-computed-table}
    \centering
    \vspace{3mm}
    \begin{tabular}{|p{2cm}|p{2cm}|p{2cm}|p{2cm}|p{2cm}|}
    \hline
	 \multicolumn{1}{|>{\centering\arraybackslash}m{2cm}}	{\textbf{$\varepsilon_1$}} 
    & 	 \multicolumn{1}{|>{\centering\arraybackslash}m{2cm}}	{\textbf{$x_1 = 0.5$}}             
    &   \multicolumn{1}{|>{\centering\arraybackslash}m{2cm}|}   {\textbf{$x_1 = 0.35$}} \T\B \\ 
    \hline    	    

	 \multicolumn{1}{|>{\centering\arraybackslash}m{2cm}}	{$0.8$} 
    &   \multicolumn{1}{|>{\centering\arraybackslash}m{2cm}}	{$0.238$} 		   	        
    &   \multicolumn{1}{|>{\centering\arraybackslash}m{2cm}|}	{$0.146$} \\ [3pt] 			  
  
	 \multicolumn{1}{|>{\centering\arraybackslash}m{2cm}}	{$0.9$} 
    &   \multicolumn{1}{|>{\centering\arraybackslash}m{2cm}}	{$0.186$} 		   	        
    &   \multicolumn{1}{|>{\centering\arraybackslash}m{2cm}|}	{$0.117$} \\ [3pt] 		   
	 
	 \multicolumn{1}{|>{\centering\arraybackslash}m{2cm}}	{$1.0$} 
    &   \multicolumn{1}{|>{\centering\arraybackslash}m{2cm}}	{$0.132$} 	   	        
    &   \multicolumn{1}{|>{\centering\arraybackslash}m{2cm}|}	{$0.079$} \\ [3pt] 		  
 
	 \multicolumn{1}{|>{\centering\arraybackslash}m{2cm}}	{$1.1$} 
    &   \multicolumn{1}{|>{\centering\arraybackslash}m{2cm}}	{$0.090$} 		   	        
    &   \multicolumn{1}{|>{\centering\arraybackslash}m{2cm}|}	{$0.058$} \\ [3pt] 	 

	 \multicolumn{1}{|>{\centering\arraybackslash}m{2cm}}	{$1.2$} 
    &   \multicolumn{1}{|>{\centering\arraybackslash}m{2cm}}	{$0.065$} 	   	        
    &   \multicolumn{1}{|>{\centering\arraybackslash}m{2cm}|}	{$0.043$} \\ [3pt] 	 

	 \multicolumn{1}{|>{\centering\arraybackslash}m{2cm}}	{$1.3$} 
    &   \multicolumn{1}{|>{\centering\arraybackslash}m{2cm}}	{$0.060$} 		   	        
    &   \multicolumn{1}{|>{\centering\arraybackslash}m{2cm}|}	{$0.038$} \\ [3pt] 	 

	 \multicolumn{1}{|>{\centering\arraybackslash}m{2cm}}	{$1.4$} 
    &   \multicolumn{1}{|>{\centering\arraybackslash}m{2cm}}	{$0.057$} 	   	        
    &   \multicolumn{1}{|>{\centering\arraybackslash}m{2cm}|}	{$0.039$} \\ [3pt] 	 
  
	 \multicolumn{1}{|>{\centering\arraybackslash}m{2cm}}	{$1.5$} 
    &   \multicolumn{1}{|>{\centering\arraybackslash}m{2cm}}	{$0.056$} 	   	        
    &   \multicolumn{1}{|>{\centering\arraybackslash}m{2cm}|}	{$0.039$} \\ [3pt] 	 

	 \multicolumn{1}{|>{\centering\arraybackslash}m{2cm}}	{$1.6$} 
    &   \multicolumn{1}{|>{\centering\arraybackslash}m{2cm}}	{$0.056$} 	   	        
    &   \multicolumn{1}{|>{\centering\arraybackslash}m{2cm}|}	{$0.038$} \\ [3pt] 	 

	 \multicolumn{1}{|>{\centering\arraybackslash}m{2cm}}	{$1.7$} 
    &   \multicolumn{1}{|>{\centering\arraybackslash}m{2cm}}	{$0.053$} 		   	        
    &   \multicolumn{1}{|>{\centering\arraybackslash}m{2cm}|}	{$0.037$} \\ [3pt] 	 

	 \multicolumn{1}{|>{\centering\arraybackslash}m{2cm}}	{$1.8$} 
    &   \multicolumn{1}{|>{\centering\arraybackslash}m{2cm}}	{$0.061$} 	   	        
    &   \multicolumn{1}{|>{\centering\arraybackslash}m{2cm}|}	{$0.041$} \\ [3pt] 	     
    \hline
  \end{tabular}
\end{table}

\vspace{3mm} 

%
%

\end{document}